\documentclass[10pt,twocolumn,letterpaper]{article}

\usepackage{cvpr}              

\usepackage{graphicx}
\usepackage{amsmath}
\usepackage{amssymb}
\usepackage{booktabs}
\usepackage{multirow}
\usepackage{hhline}
\usepackage{paralist, tabularx}
\DeclareMathOperator*{\argmax}{arg\,max}
\newcommand{\blockcomment}[1]{}
\newcommand{\KO}[1]{\textcolor{blue}{[Kemal: #1]}}
\usepackage{pifont}
\newcommand{\cmark}{\ding{51}}%
\newcommand{\xmark}{\ding{55}}%
\usepackage{algpseudocode}
\usepackage{algorithm}
\usepackage{listings}
\usepackage{booktabs}
\usepackage{amsthm}
\usepackage[dvipsnames]{xcolor}

\newtheorem{theorem}{Theorem}

\usepackage[acronym,nowarn,section,nogroupskip,nonumberlist]{glossaries}

\glsdisablehyper{}
\newacronym{ID}{ID}{in-distribution}
\newacronym{BA}{BA}{Balanced Accuracy}
\newacronym{OOD}{OOD}{out-of-distribution}
\newacronym{SAOD}{SAOD}{Self-aware Object Detection}
\newacronym{SAODet}{SAODet}{Self-aware Object Detector}
\newacronym{AP}{AP}{Average Precision}
\newacronym{DAQ}{DAQ}{Detection Awareness Quality}
\newacronym{IDQ}{IDQ}{In-Distribution Quality}
\newacronym{LaECE}{LaECE}{Localisation-aware Expected Calibration Error}
\newacronym{ECE}{ECE}{Expected Calibration Error}
\newacronym{AUC}{AUC}{area-under-curve}
\newacronym{TP}{TP}{true-positive}
\newacronym{FP}{FP}{false-positive}
\newacronym{FN}{FN}{false-negative}
\newacronym{IoU}{IoU}{Intersection-over-Union}
\newacronym{LRP}{LRP}{Localisation-Recall-Precision Error}
\newacronym{AV}{AV}{Autonomous Vehicles}
\newacronym{AUROC}{AUROC}{Area-under ROC Curve}

\newcommand{\oodname}{SiNObj110K}
\newcommand{\indata}{\mathcal{D}_{\mathrm{ID}}}
\newcommand{\testdata}{\mathcal{D}_{\mathrm{Test}}}
\newcommand{\traindata}{\mathcal{D}_{\mathrm{Train}}}
\newcommand{\valdata}{\mathcal{D}_{\mathrm{Val}}}
\newcommand{\shiftdata}{\mathcal{T}(\mathcal{D}_{\mathrm{ID}})}
\newcommand{\ooddata}{\mathcal{D}_{\mathrm{OOD}}}
\newcommand{\image}{X}

\usepackage[pagebackref,breaklinks,colorlinks]{hyperref}

\usepackage[capitalize]{cleveref}
\crefname{section}{Sec.}{Secs.}
\Crefname{section}{Section}{Sections}
\Crefname{table}{Table}{Tables}
\crefname{table}{Tab.}{Tabs.}

\def\cvprPaperID{2681} 
\def\confName{CVPR}
\def\confYear{2023}

\begin{document}

\title{Towards Building Self-Aware Object Detectors via Reliable Uncertainty Quantification and Calibration}
\author{Kemal Oksuz, \; Tom Joy, \; Puneet K. Dokania \\
Five AI Ltd., United Kingdom\\
{\tt\small \{kemal.oksuz, tom.joy, puneet.dokania\}@five.ai}
}
\maketitle

\begin{abstract}
The current approach for testing the robustness of object detectors suffers from serious deficiencies such as improper methods of performing out-of-distribution detection and using calibration metrics which do not consider both localisation and classification quality.
In this work, we address these issues, and introduce the \textbf{S}elf \textbf{A}ware \textbf{O}bject \textbf{D}etection (SAOD) task, a unified testing framework which respects and adheres to the challenges that object detectors face in safety-critical environments such as autonomous driving.
Specifically, the SAOD task requires an object detector to be: robust to domain shift; obtain reliable uncertainty estimates for the entire scene; and provide calibrated confidence scores for the detections.
We extensively use our framework, which introduces novel metrics and large scale test datasets, to test numerous object detectors in two different use-cases, allowing us to highlight critical insights into their robustness performance.
Finally, we introduce a simple baseline for the SAOD task, enabling researchers to benchmark future proposed methods and move towards robust object detectors which are fit for purpose. Code is available at: \url{https://github.com/fiveai/saod}.
\end{abstract}

\section{Introduction}\label{sec:introduction}
%
%
%
%
The safe and reliable usage of object detectors in safety critical systems such as autonomous driving~\cite{Cityscapes,bdd100k,waymo}, depends not only on its accuracy, but also critically on other robustness aspects which are often only considered in addition or not all.
These aspects represent its ability to be robust to domain shift, obtain well-calibrated predictions and yield reliable uncertainty estimates at the image-level, enabling it to flag the scene for human intervention instead of making unreliable predictions.
%
Consequently, the development of object detectors for safety critical systems requires a thorough evaluation framework which also accounts for these robustness aspects, a feature lacking in current evaluation methodologies.

Whilst object detectors are able to obtain uncertainty at the \emph{detection-level}, they do not naturally produce uncertainty at the \emph{image-level}.
This has lead researchers to often evaluate uncertainty by performing \gls{OOD} detection at the detection-level~\cite{RegressionUncOD,VOS}, which cannot be clearly defined.
Thereby creating a misunderstanding between \gls{OOD} and \gls{ID} data.
This leads to an improper evaluation, as defining \gls{OOD} at the detection level is non-trivial due to the presence of known-unknowns or background objects~\cite{opensetelephant}. 
Furthermore, the test sets for \gls{OOD} in such evaluations are small, typically containing around 1-2K images\cite{RegressionUncOD,VOS}.

Moreover, as there is no direct access to the labels of the test sets and the evaluation servers only report accuracy \cite{COCO,LVIS}, researchers have no choice but to use small validation sets as testing sets to report robustness performance, such as calibration and performance under domain shift. 
As a result, either the training set~\cite{devil,Norcal}; the validation set~\cite{RegressionUncOD,VOS}; or a subset of the validation set~\cite{mvcalibrationod} is employed for cross-validation, leading to an unideal usage of the dataset splits and a poor choice of the hyper-parameters.
%
%
%
\begin{figure}[t]
        \centering
        \includegraphics[width=0.48\textwidth]{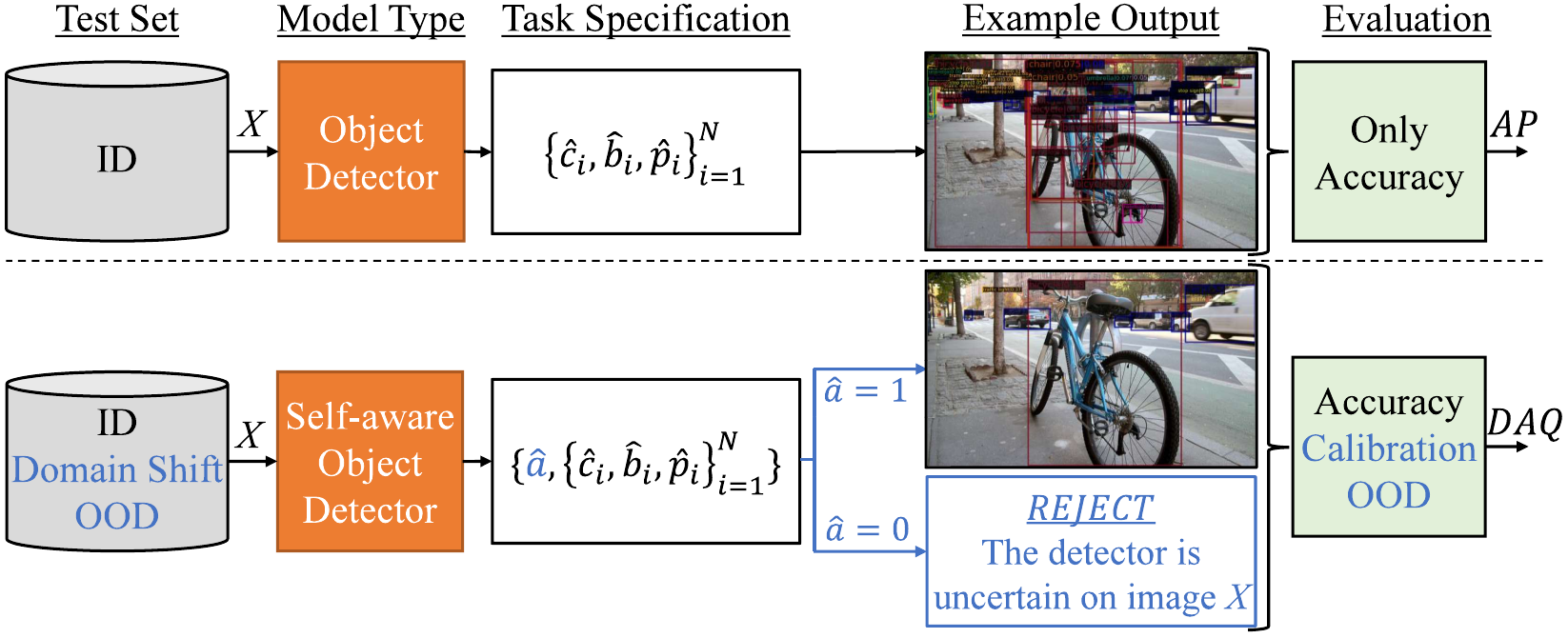}
        \vspace{-4.0ex}
        \caption{(Top) The vanilla object detection task vs. (Bottom) the self-aware object detection (SAOD) task. Different from the vanilla approach; the SAOD task requires the detector to: predict $\hat{a} \in \{0,1\}$ representing whether the image $X$ is accepted or not for further processing; yield accurate and calibrated detections; and be robust to domain shift. Accordingly, for SAOD we evaluate on ID, domain-shift and OOD data using our novel DAQ measure. Here, $\{\hat{c}_i, \hat{b}_i, \hat{p}_i\}^N$ are the predicted set of detections.
        }
        \label{fig:teaser}
        \vspace{-3.0ex}
\end{figure} 

Finally, prior work typically focuses on only one of:  calibration~\cite{CalibrationOD,mvcalibrationod}; \gls{OOD} detection~\cite{VOS}; domain-shift~\cite{RobustnessODBenchmark,oddomainshiftcycle,DGOD,clipgap}; or leveraging uncertainty to improve accuracy \cite{KLLoss,sampleweightingnet,BayesOD,narrowingthegap,ActiveLearningOD}, with no prior work taking a holistic approach by evaluating all of them.
Specifically for calibration, previous studies either consider classification calibration~\cite{CalibrationOD}, or localisation calibration~\cite{mvcalibrationod}, completely disregarding the fact that object detection is a joint problem.

In this paper, we address the critical need for a robust testing framework which evaluates object detectors thoroughly, thus alleviating the aforementioned deficiencies.
To do this, we introduce the \gls{SAOD} task, which considers not only accuracy, but also calibration
using our novel \gls{LaECE} as well as the reliability of image-level uncertainties.
Furthermore, the introduction of \gls{LaECE} addresses a critical gap in the literature as it respects both \emph{classification} and \emph{localisation} quality, a feature ignored in previous methods~\cite{CalibrationOD, mvcalibrationod}.
Moreover, the SAOD task requires an object detector to either perform reliably or reject images outside of its training domain.

%
We illustrate the \gls{SAOD} task in \cref{fig:teaser}, which not only evaluates the accuracy, but also the calibration and performance under \gls{OOD} or domain-shifted data.
We can also see the functionality to reject an image, and to only produce detections which have a high confidence; unlike for a standard detector which has to accept every image and produce detections.
%
%
To summarise, our main contributions are:
\blockcomment{
\begin{compactitem}
\item We introduce the \gls{SAOD} task, which evaluates: accuracy; robustness to domain shift; ability to accept of reject an image; and calibration in a unified manner. To facilitate this, we make further contributions: i) The introduction of \gls{LaECE}, a unified metric which considers the calibration of localisation and classification performance; and ii) the introduction of large-scale test datasets totaling 155K images, enabling robust evaluation.
\item We subsequently provide a baseline for the \gls{SAOD} task which converts and object detector into one the is self-aware. In doing so we discover that object detectors are inherently very strong \gls{OOD} detectors and provide reliable image-level uncertainties and can be calibrated using an appropriate method.
\end{compactitem}
}
\begin{compactitem}
\item We introduce the \gls{SAOD} task, which evaluates: accuracy; robustness to domain shift; ability to accept of reject an image; and calibration in a unified manner. We further construct large datasets totaling 155K images and provide a simple baseline for future researchers to benchmark against.
\item We explore how to obtain image-level uncertainties from any object detector, enabling it to reject the entire scene for the \gls{SAOD} task. Through our investigations, we discover that object detectors are inherently strong \gls{OOD} detectors and provide reliable uncertainties.
\item Finally, we define the \gls{LaECE} as a novel calibration measure for object detectors in \gls{SAOD}, which requires the confidence of a detector to represent both its \emph{classification} as well as its \emph{localisation} quality.
\end{compactitem}

\section{Notations and Preliminaries}
\label{sec:relatedwork}

\label{subsec:objectdetectors}
\textbf{Object Detection} Given that the set of $M$ objects in an image $X$ is represented by $\{b_i, c_i\}^M$ where $b_i \in \mathbb{R}^{4}$ is a bounding box and $c_i \in  \{1,\dots, K\}$ its class; the goal of an object detector is to predict the bounding boxes and the class labels for the objects in $X$, $f(X) = \{\hat{c}_i, \hat{b}_i, \hat{p}_i\}^N$, where $\hat{c}_i, \hat{b}_i, \hat{p}_i$ represent the class, bounding box and confidence score of the $i$th detection respectively and $N$ is the number of predictions.
%
%
%
Conventionally, the detections are obtained in two steps, $f(X) = (h \circ g)(X)$~\cite{FasterRCNN,FocalLoss,DETR,sparsercnn}: 
where $g(X) = \{\hat{b}^{raw}_i, \hat{p}^{raw}_i \}^{N^{raw}}$ is a deep neural network predicting raw detections with bounding boxes $\hat{b}^{raw}_i$ and predicted class distribution $\hat{p}^{raw}_i$.
Given these raw-detections $h(\cdot)$ applies post-processing to obtain the final detections\footnote{for probabilistic detectors~\cite{BayesOD,KLLoss,sampleweightingnet,PDQ,RegressionUncOD}, $\hat{b}^{raw}_i$ follows a probability distribution mostly of the form $g(X) = \{\mathcal{N}(\mu_i, \Sigma_i), \hat{p}^{raw}_i\}^{N^{raw}}$, where $\Sigma_i$ is either a diagonal~\cite{KLLoss,sampleweightingnet} or full covariance matrix~\cite{BayesOD}}. 
In general, $h(\cdot)$ comprises removing the detections predicted as background; Non-Maximum-Suppression (NMS) to discard duplicates; and keeping useful detections, normally achieved through top-$k$ survival, where in practice $k=100$ for COCO dataset~\cite{COCO}.

\textbf{Evaluating the Performance of Object Detectors}
\gls{AP}~\cite{COCO,LVIS,PASCAL}, or the area under the precision-recall (PR) curve, has been the common performance measure of object detection.
Though widely accepted, AP suffers from the following three main drawbacks~\cite{LRPPAMI}.
First, it only validates \glspl{TP} using a localisation quality threshold, completely disregarding the continuous nature of localisation.
Second, as an \gls{AUC} measure, AP is difficult to interpret, as PR curves with different characteristics can yield the same value.
Also, AP rewards a detector that produces a large number of low scoring detections than actual objects in the image, which becomes a significant issue when relying on top-$k$ survival as shown in \cref{fig:teaser}. App. \ref{app:calibration} includes details.


Alternatively, the recently proposed \gls{LRP}~\cite{LRP,LRPPAMI} 
%
%
combines the number of \gls{TP}, \gls{FP}, \gls{FN}, denoted by $\mathrm{N_{TP}}$, $\mathrm{N_{FP}}$, $\mathrm{N_{FN}}$, respectively, as well as the \gls{IoU} of \gls{TP}s with the objects that they match with:
{\small
\begin{align}\label{eq:LRPdefcompact}
     \frac{1}{\mathrm{N_{FP}} +\mathrm{N_{FN}}+\mathrm{N_{TP}}}\left(\mathrm{N_{FP}} +\mathrm{N_{FN}} + \sum \limits_{\psi(i) > 0} (1 - \mathrm{lq}(i)) \right)
\end{align}
}
where $\mathrm{lq}(i) = \frac{\mathrm{IoU}(\hat{b}_i, b_{\psi(i)})-\tau}{1-\tau}$ is the localisation quality with $\tau$ being the \gls{TP} assignment threshold, $\psi(i)$ is the index of the object that a \gls{TP} $i$ matches to; else $i$ is a \gls{FP} and $\psi(i) = -1$. 
LRP can be decomposed into components providing insights on: the localisation quality; the precision; and the recall error.
Besides, low-scoring detections are demoted by the term $\mathrm{N_{FP}}$ in Eq. \eqref{eq:LRPdefcompact}.
Thus, \gls{LRP} arguably alleviates the aforementioned drawbacks of AP.

\section{An Overview to the \gls{SAOD} Task} \label{sec:saod}
For object detectors to be deployed in safety critical systems it is imperative that they perform in a robust manner.
Specifically, we would expect the detector to be aware of situations when the scene differs substantially from the training domain and to include the functionality to flag the scene for human intervention.
Moreover, we also expect that the confidence of the detections matches the performance, referred to as calibration. 
With these expectations in mind, we characterise the crucial elements needed to evaluate and perform the \gls{SAOD} task. 
Specifically, the \gls{SAOD} task requires an object detector to:
\begin{compactitem}
    \item Have the functionality to reject a scene based on its image-level uncertainties through a binary indicator variable $\hat{a} \in \{0,1\}$.
    \item Produce detection-level confidences that are calibrated in terms of classification \emph{and} localisation.
    \item Be robust to domain-shift.
\end{compactitem}
For brevity, and to enable future researchers to adopt the \gls{SAOD} framework, the explicit practical specification for \glspl{SAODet} is
\begin{align}
    f_A(X) = \{\hat{a}, \{\hat{c}_i, \hat{b}_i, \hat{p}_i\}^N\},
\end{align}
where $\hat{a} \in \{0,1\}$ implies if the image should be accepted or rejected and that the predicted confidences $\hat{p}_i$ are calibrated.

\textbf{Evaluation Datasets}
As the \gls{SAOD} emulates challenging real-life situations, the evaluation needs to be performed using large-scale test datasets.
Unlike 
previous approaches on \gls{OOD} detection using around 1-2K OOD images~\cite{RegressionUncOD,VOS} for testing or calibration methods \cite{mvcalibrationod} relying on 2.5K \gls{ID} test images, 
our test set totals to 155K individual images for each of our two use-cases when combining \gls{ID} and \gls{OOD} data.
Specifically, we construct two test datasets, where each $\testdata$ in our case is the \emph{union} of the following datasets:
\begin{compactitem}
    \item $\indata$ ($45K$ Images): \gls{ID} dataset with images containing the same foreground objects as were  present in $\traindata$. 
    \item $\shiftdata$ ($3 \times 45K$ Images): domain-shift dataset obtained by applying transformations to the images from $\indata$, which preserve the semantics of the image.
    \item $\ooddata$ ($110K$ Images): \gls{OOD} dataset with images that do not contain any foreground object from $\indata$. These images tend to include objects not present in $\traindata$.
\end{compactitem}
We present exact splits in \cref{tab:datasets} for object detection in General and \gls{AV} use-cases (refer App. \ref{app:datasets} for further details).
Collected from a different dataset, our $\indata$ differs from $\traindata$, but is still semantically similar; which is reflective of a challenging real-word scenario, as domains change over time and scenes differ in terms of appearance.
For $\shiftdata$, we apply ImageNet-C style corruptions~\cite{hendrycks2019robustness} to $\indata$, where for each image we randomly choose one of 15 corruption types (fog, blur, noise, etc.) at severity levels 1, 3 and 5 as is common in practice~\cite{RegressionUncOD}.
Then, we expect that for a given input $\image \in \testdata$, a \gls{SAODet} makes the following decisions:
\begin{compactitem}
    \item if $\image \in \indata \cup \shiftdata$ for corruption severities 1 and 3, `accept' the input and provide \textit{accurate and  calibrated} detections. Penalize a rejection.
    \item if $\image \in \shiftdata$ at corruption severity 5, provide the choice to `accept' and evaluate but do not \emph{not} penalize a `rejection' as the transformed images might not contain enough cues to perform object detection reliably. 
    \item if $\image \in \ooddata$, `reject' the image and provide \textit{no} detections as, by design, the predictions would be wrong. An `accept' should be penalized in this case.
\end{compactitem}

\textbf{Models}\label{sec:models}
In terms of evaluating \gls{SAOD} on common object detectors, it would prove useful at this point to introduce the models used in our investigation. We mainly exploit a diverse set of four object detectors:
\begin{compactitem}
    \item[1.] Faster R-CNN (F-RCNN)~\cite{FasterRCNN} is a two-stage detector with a softmax classifier
    \item[2.] Rank \& Sort R-CNN (RS-RCNN)~\cite{RSLoss} is another two-stage detector but with a ranking-based loss function and sigmoid classifiers
    \item[3.] Adaptive Training Sample Selection (ATSS)~\cite{ATSS} is a common one-stage baseline with sigmoid classifiers
    \item[4.] Deformable DETR (D-DETR)~\cite{DDETR} is a transformer-based model, again using sigmoid classifiers
\end{compactitem}
%
We also evaluate two probabilistic detectors with a diagonal covariance matrix minimizing the negative log likelihood~\cite{KLLoss} (NLL-RCNN) or energy score~\cite{RegressionUncOD} (ES-RCNN), allowing us to obtain uncertainty estimates for localisation.
%
%
Please see App. \ref{app:models} for the training details of the methods as well as their accuracy on $\valdata$, $\mathcal{T}(\mathcal{D}_{\mathrm{Val}})$, $\indata$ and $\shiftdata$.

As we have now outlined clear requirements for a \gls{SAODet}, it is natural to ask how well the aforementioned object detectors perform under these requirements.
We will extensively investigate this by first introducing a simple method to extract image-level uncertainty enabling the acceptance or rejection of an image in~\cref{sec:ood}; evaluate the calibration and provide methods to calibrate such detectors in~\cref{sec:calibration}; before finally providing a complete analysis of them using the \gls{SAOD} framework in~\cref{sec:evaluation}.

 

\begin{table}[]
    \caption{Our dataset splits for SAOD. We design test sets for COCO \cite{COCO} and nuImages \cite{nuimages} as ID data (train \& val). We exploit Objects365 \cite{Objects365} and BDD100K \cite{bdd100k} for $\mathcal{D}_{\mathrm{ID}}$ and $\mathcal{T}(\mathcal{D}_{\mathrm{ID}})$, and use Objects365, iNaturalist \cite{inaturalist} and SVHN \cite{SVHN} for $\mathcal{D}_{\mathrm{OOD}}$.}
    \centering
    \scalebox{0.58}{
    \begin{tabular}{c|ccccc}
        \toprule
         Dataset&  \multirow{2}{*}{$\traindata$} & \multirow{2}{*}{$\valdata$} & \multicolumn{3}{c}{$\testdata$}\\
         \cline{4-6}
         & & & $\mathcal{D}_{\mathrm{ID}}$ & $\mathcal{T}(\mathcal{D}_{\mathrm{ID}})$ & $\mathcal{D}_{\mathrm{OOD}}$ \\
         \midrule
         SAOD-Gen & $\text{COCO}^{(train)}$ & $\text{COCO}^{(val)}$ & Obj45K & Obj45K-C & \oodname-OOD \\
         SAOD-AV & $\text{nuImages}^{(train)}$ & $\text{nuImages}^{(val)}$ & BDD45K & BDD45K-C & \oodname-OOD \\
         \bottomrule
    \end{tabular}
    }
    \vspace{-2ex}
    \label{tab:datasets}
\end{table}

\section{Obtaining Image-level Uncertainty}\label{sec:ood}
As there is no clear distinction between background and an \gls{OOD} object unless each pixel in $\traindata$ is labelled \cite{opensetelephant}, evaluating uncertainties of detectors is nontrivial at detection-level. 
Thus, different from prior work \cite{RegressionUncOD,VOS} conducting \gls{OOD} detection at detection-level, we evaluate the uncertainties on image-level \gls{OOD} detection task.
Thereby aligning the evaluation and the definition of an \gls{OOD} image.
Please see App. \ref{subsec:whyimagelevel} for further discussion.

Practically, one method to accept or reject an image is to obtain an estimate of uncertainty at the image-level through a function $\mathcal{G} : \mathcal{X} \rightarrow \mathbb{R}$ and a threshold $\bar{u} \in \mathbb{R}$, where the image is accepted if $\mathcal{G}(X) < \bar{u}$ and $\hat{a}=1$; and rejected vice-versa.
We take this approach when constructing our baseline and now specifically outline the method to do so.

\begin{table}
    \small
    \setlength{\tabcolsep}{0.45em}
    \centering
    \caption{AUROC scores (in \%) for image-level uncertainties when aggregating through different methods, where we use the uncertainty score of $1-\hat{p}_{i}$ for the detections. Here, top-$m$ refers to the average of the lowest  $m$ uncertainties for the detections. As we can see, using the most certain detections performs better. Bold and underline are best and second best respectively. }
    \label{tab:aggregate}
    \vspace{-1ex}
    \scalebox{0.9}{
    \begin{tabular}{c|c||c|c|c|c|c|c} 
    \toprule
    \midrule
         Dataset &Detector&sum&mean&top-5&top-3&top-2&min   \\ \midrule
    \multirow{4}{*}{SAOD-Gen}&F-RCNN&$20.9$&$84.1$&$93.4$&\underline{$94.1$}& $\mathbf{94.4}$&$93.8$\\
    &RS-RCNN &$85.8$&$85.8$&$94.3$&$\mathbf{94.8}$&$\mathbf{94.8}$&$93.5$\\
    &ATSS &$66.2$&$86.3$&$93.8$&$\mathbf{94.2}$&\underline{$94.0$}&$92.6$\\
    &D-DETR &$85.2$&$85.2$&$94.4$&$\mathbf{94.7}$&\underline{$94.6$}&$93.3$\\ \midrule
    \multirow{2}{*}{SAOD-AV}&F-RCNN&$27.1$&$84.1$&$96.4$&\underline{$97.3$}&$\mathbf{97.4}$&$96.0$\\
    &ATSS&$18.8$&$92.2$&$\mathbf{97.7}$&\underline{$97.6$}&$97.3$&$95.7$\\
    \midrule
    \bottomrule
    \end{tabular}
    }
    \vspace{-1ex}
\end{table}

\begin{table}
    \centering
    \small
    \setlength{\tabcolsep}{0.35em}
    \caption{AUROC scores (in \%) of different detection-level uncertainty estimates. Classification-based uncertainties perform better compared to localization and $1-\hat{p}_{i}$ performs generally the best.}
    \label{tab:different_unc}
    \vspace{-1ex}
    \scalebox{0.9}{
    \begin{tabular}{c|c||c|c|c|c|c|c} 
    \toprule
    \midrule
    \multirow{2}{*}{Dataset}&\multirow{2}{*}{Detector} & \multicolumn{3}{c|}{Classification}& \multicolumn{3}{c}{Localisation} \\
     &&$\mathrm{H}(\hat{p}^{raw}_i)$&$\mathrm{DS}$&$1-\hat{p}_{i}$&$|\Sigma|$&$\mathrm{tr}(\Sigma)$&$\mathrm{H}(\Sigma)$\\  \midrule
    &F-RCNN&$92.6$&$89.7$&$\mathbf{94.1}$ &N/A&N/A&N/A\\
    &RS-RCNN &$93.7$&$30.0$&$\mathbf{94.8}$&N/A&N/A&N/A\\ 
    SAOD&ATSS&$\mathbf{94.3}$&$36.9$ &$94.2$&N/A&N/A&N/A\\
    Gen&D-DETR&$93.9$&$73.8$&$\mathbf{94.4}$&N/A&N/A&N/A\\ \cmidrule(l){2-8}
    &NLL-RCNN&$92.4$&$89.0$&$\mathbf{94.1}$&$87.6$&$87.5$&$87.7$\\
    &ES-RCNN &$92.8$&$89.9$&$\mathbf{94.1}$&$85.0$&$85.2$&$86.4$\\ \midrule
    SAOD&F-RCNN&$\mathbf{97.3}$&$96.0$&$\mathbf{97.3}$ &N/A&N/A&N/A\\
    AV&ATSS&$97.2$&$97.1$&$\mathbf{97.6}$&N/A&N/A&N/A\\ 
    \midrule
    \bottomrule
    \end{tabular}
    }
    \vspace{-2.5ex}
\end{table}

\textbf{Obtaining Image-level Uncertainties} \label{subsec:reliability}
This can be achieved through aggregating the detection-level uncertainties.
We hypothesise that there is implicitly enough uncertainty information in the detections to produce image-level uncertainty, they just need to be extracted and aggregated in an appropriate way.
In terms of the extraction, we can obtain detection level uncertain through: the uncertainty score ($1-\hat{p}_{i}$); the entropy of the predictive classification distribution of the raw detections ($\mathrm{H}(\hat{p}^{raw}_i)$); and Dempster-Shafer \cite{DS,QyeryHardFive} ($\mathrm{DS}$). 
In addition, for probabilistic detectors, we can extract uncertainty from $\Sigma$ by taking the: determinant, trace, or entropy of the multivariate normal distribution\cite{pml1Book}.
In terms of the aggregation strategy, given the uncertainties for the detections after top-$k$ survival, we let $\mathcal{G}$ either take their: sum, mean, minimum, or their mean of the $m$ smallest uncertainty values, i.e. the most certain top-$m$ detections. For further details, please see App. \ref{app:uncertainty}.
Whilst these strategies are simple, as we will now show, they provide a suitable method to obtain image-level uncertainty, enabling effective performance on \gls{OOD} detection, a common task for evaluating uncertainty quantification.
%

To do this, we evaluate the \gls{AUROC} score between the uncertainties of the data from $\mathcal{D}_{\mathrm{ID}}$ and $\mathcal{D}_{\mathrm{OOD}}$ and display the results in \cref{tab:aggregate}; which shows that high \gls{AUROC} scores are obtained when $\mathcal{G}$ is formed by considering up to the mean(top-5) detections, with the mean(top-3) aggregation strategy of $1-\hat{p}_{i}$ performs the best.
This highlights that the detections with lowest uncertainty in each image provide useful information to reliably estimate image-level uncertainty.
We believe the poor performance for mean and sum stem from the fact that there are typically too many noisy detections (up to $k=100$) for only a few objects in the image.
We further provide assurance that $1-\hat{p}_{i}$ is the most appropriate method to extract detection-level uncertainty in \cref{tab:different_unc}, where we can see that $1-\hat{p}_{i}$ obtains higher \gls{AUROC} scores compared to $\mathrm{H}(\hat{p}^{raw}_i)$ and $\mathrm{DS}$.
We also note that classification uncertainties (except DS) perform consistently better than localisation ones for probabilistic detectors. 
We believe one of the reasons for that is the classifier is trained using both the proposals matching and not matching with any object, preventing the detector from becoming over-confident everywhere.

%
%
%

%
\begin{figure}[t]
        \centering
        \includegraphics[width=0.48\textwidth]{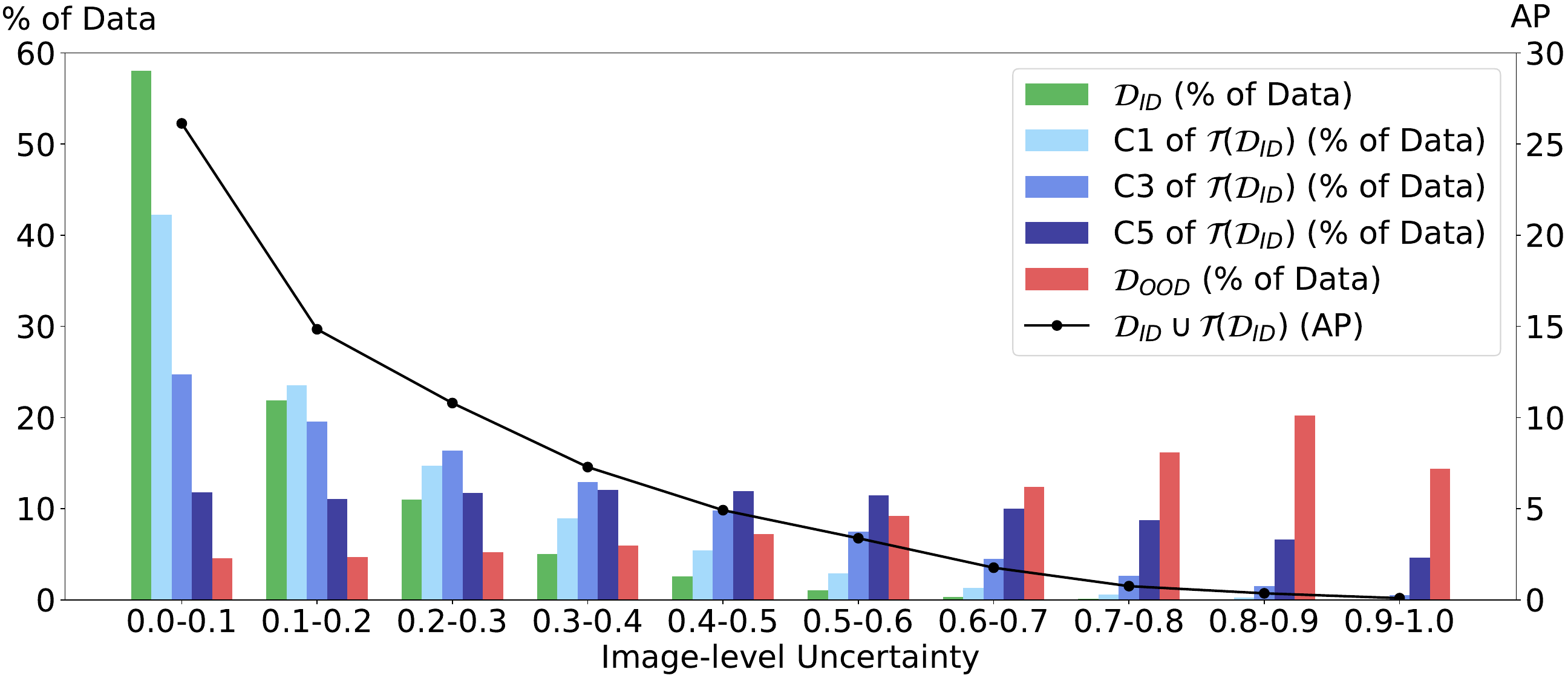}
        \vspace{-3ex}
        \caption{The distribution of image-level uncertainties obtained from F-RCNN (SAOD-Gen) on $\indata$, different severities 1, 3, 5 (C1, C3, C5) of $\shiftdata$ and $\ooddata$ vs. the accuracy in COCO-Style AP in \% (AP in short).
        App. \ref{app:uncertainty} includes more examples.
        }
        \label{fig:unc_reliability}
        \vspace{-1ex}
\end{figure}

\begin{table}
    \centering
    \small
    \setlength{\tabcolsep}{0.8em}
    \caption{\gls{AUROC} scores  (in \%) on subsets of $\ooddata$. In all cases, near-OOD (Obj365) has a lower AUROC than far-\gls{OOD}  (SVHN).}
    \label{tab:oodperf_different_subsets}
    \vspace{-2ex}
    \scalebox{0.9}{
    \begin{tabular}{c|c||c|c|c||c}
    \toprule
    \midrule
         \multirow{2}{*}{Dataset} &\multirow{2}{*}{Detector} & \multicolumn{3}{|c|}{near to far \gls{OOD}}&\multirow{2}{*}{all \gls{OOD}} \\ \cline{3-5}
          &&Obj365&iNat&SVHN&\\ \hline
    &F-RCNN&$83.7$&$97.6$&$\mathbf{99.8}$&$94.1$ \\
    SAOD&RS-RCNN &$85.6$&$97.8$&$\mathbf{99.8}$&$94.8$ \\
    Gen&ATSS &$84.5$&$97.4$&$99.5$&$94.2$ \\
    &D-DETR &$85.7$&$\mathbf{98.1}$&$99.4$&$94.4$  \\ \hline
    SAOD&F-RCNN&$\mathbf{95.2}$&$97.4$&$98.8$&$97.3$ \\
    AV&ATSS&$95.0$&$97.3$&$99.7$&$\mathbf{97.7}$  \\ 
    \midrule
    \bottomrule
    \end{tabular}
    }
    \vspace{-2ex}
\end{table}

\textbf{How Reliable are these Image-level Uncertainties?}
Though the aforementioned results show that the image-level uncertainties are effective, we now see how reliable these uncertainties are in practice.
For this, we first evaluate the detectors on different subsets of our \oodname{ }\gls{OOD} set. 
\cref{tab:oodperf_different_subsets} shows that for all detectors, the \gls{AUROC} score is lower for near-\gls{OOD} subset (Obj365) than for far-\gls{OOD} (iNat and SVHN) and is consistently very high for far-\gls{OOD} subsets (up to $99.8$ on SVHN).
%

We then consider the uncertainties of $\mathcal{D}_{\mathrm{ID}}$, $\mathcal{T}(\mathcal{D}_{\mathrm{ID}})$ and $\mathcal{D}_{\mathrm{OOD}}$ by plotting histograms of the image-level uncertainties in 10 equally-spaced bins in the range of $[0,1]$.
In \cref{fig:unc_reliability} we see that the uncertainties from $\mathcal{D}_{\mathrm{ID}}$ have a significant amount of mass in the smaller bins and vice versa for $\mathcal{D}_{\mathrm{OOD}}$, moreover the uncertainties get larger as the severity of corruption increases.
%
%
We also display \gls{AP} (black line), where it can be clearly seen that as the uncertainty increases AP decreases, implying that the uncertainty reflects the performance of the detector.
Thereby suggesting that \textit{the image-level uncertainties are reliable and effective}.
As already pointed out, this conclusion is not necessarily very surprising, since the classifiers of object detectors are generally trained not only by proposals matching the objects but also by a very large number of proposals not matching with any object, which can be $\sim$ 1000 times more \cite{Review}. 
This composition of training data prevents the classifier from becoming drastically over-confident for unseen data, enabling the detector to yield reliable uncertainties.

\textbf{Thresholding Image-level Uncertainties}
%
%
For our \gls{SAOD} baseline, we can obtain an appropriate value for $\bar{u}$ through cross-validation.
Ideally, this will require a validation set including both ID and OOD images, but unfortunately $\valdata$ consists of only \gls{ID} images.
%
%
However, given that in this case our image-level uncertainty is obtained by aggregating detection-level uncertainties, the images which have detections with high uncertainty will produce high image-level uncertainty and vice-versa.
Using this fact, if we remove the ground-truth objects from the images in $\valdata$, the resulting image-level uncertainties should be high.
We leverage this approach to construct a \emph{pseudo \gls{OOD}} dataset out of $\valdata$, by replacing the pixels inside the ground-truth bounding boxes with zeros, thereby removing them from the image and enabling us to cross-validation.

As for the metric to cross-validate $\bar{u}$ against, we observe that existing metrics such as: \gls{AUC} metrics are unsuitable to evaluate binary predictions, F-Score is sensitive to the choice of the positive class \cite{regmixup} and TPR@0.95\footnote{Which is the FPR for a fixed threshold set when TPR=0.95.} \cite{VOS,scalingOOD} requires a fixed threshold.
%
%
As an attractive candidate, Uncertainty Error \cite{unc_est_merging} computes the arithmetic mean of \gls{FP} and \gls{FN} rates.
%
However, the arithmetic mean does not heavily penalise choosing $\bar{u}$ on extreme values, potentially leading to the situation where $\hat{a} = 1$ or $\hat{a} = 0$ for all images.
To address this, we instead leverage the harmonic mean, which is sensitive to these extreme values.
Particularly, we define the \gls{BA} as the harmonic mean of \gls{TP} rate (TPR) and \gls{FP} rate (FPR), addressing the aforementioned issue and enabling us to use it to obtain a suitable $\bar{u}$.

\section{Calibration of Object Detectors}\label{sec:calibration}
Accepting or rejecting an image is only one component of the \gls{SAOD} task, in situations where the image is accepted \gls{SAOD} then requires the detections to be calibrated.
Here we define calibration as the alignment of performance and confidence of a model; which has already been extensively studied for the classification task~\cite{calibration,AdaptiveECE,verifiedunccalibration,rethinkcalibration,FocalLoss_Calibration,calibratepairwise}.
However, existing work which studies the calibration properties of an object detector~\cite{CalibrationOD, mvcalibrationod,calibrationdomainshiftod,RelaxedSE} is limited.
For object detection, the goal is to align a detector's confidence with the quality of the joint task of classification and localisation (regression).
Arguably, it is not obvious how to extend merely classification-based calibration measures such as \gls{ECE}~\cite{calibration} for object detection.
A straightforward extension would be to replace the accuracy in such measures by the precision of the detector, which is computed by validating \gls{TP}s from a specific IoU threshold.
However, this perspective, as employed by \cite{CalibrationOD}, does not account for the fact that two object detectors, while having the same precision, might differ significantly in terms of localisation quality. 

Hence, as one of the main contributions of this work, we consider the calibration of object detectors from a fundamental perspective and define Localisation-aware Calibration Error (\gls{LaECE}) which accounts for the joint nature of the task (classification and localisation).
We further analyse how calibration measures should be coupled with accuracy in object detection and adapt common post hoc calibration methods such as histogram binning~\cite{zadrozny2001obtaining}, linear regression, and isotonic regression~\cite{zadrozny2002transforming} to improve \gls{LaECE}.
%
%
%
%
%
%
%
%
%
%
%

\begin{figure}[t]
        \captionsetup[subfigure]{}
        \centering
        \begin{subfigure}[b]{0.14\textwidth}
            \includegraphics[width=\textwidth]{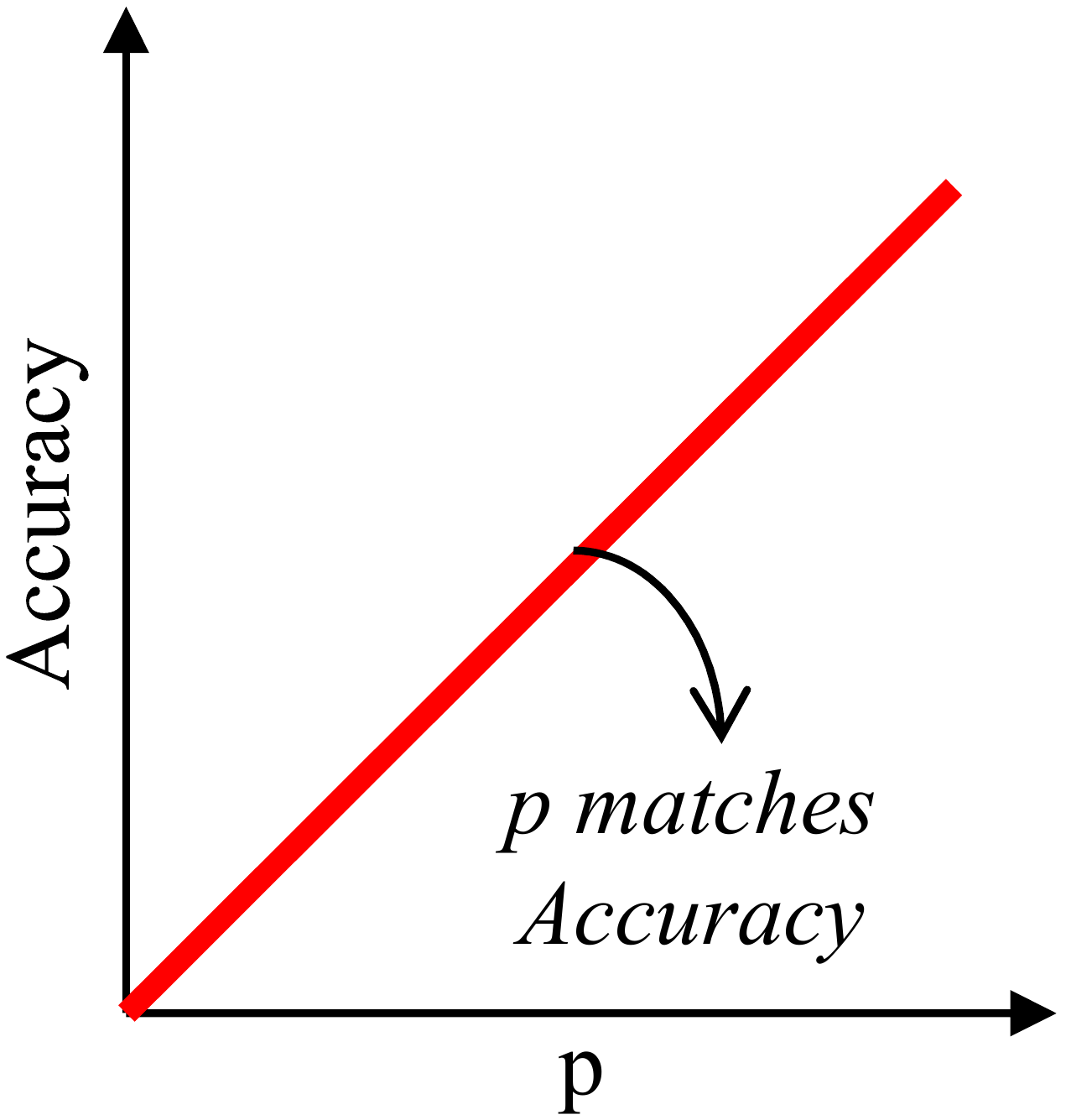}
            \caption{Classifier \cite{calibration}}
        \end{subfigure}
        \begin{subfigure}[b]{0.17\textwidth}
            \includegraphics[width=\textwidth]{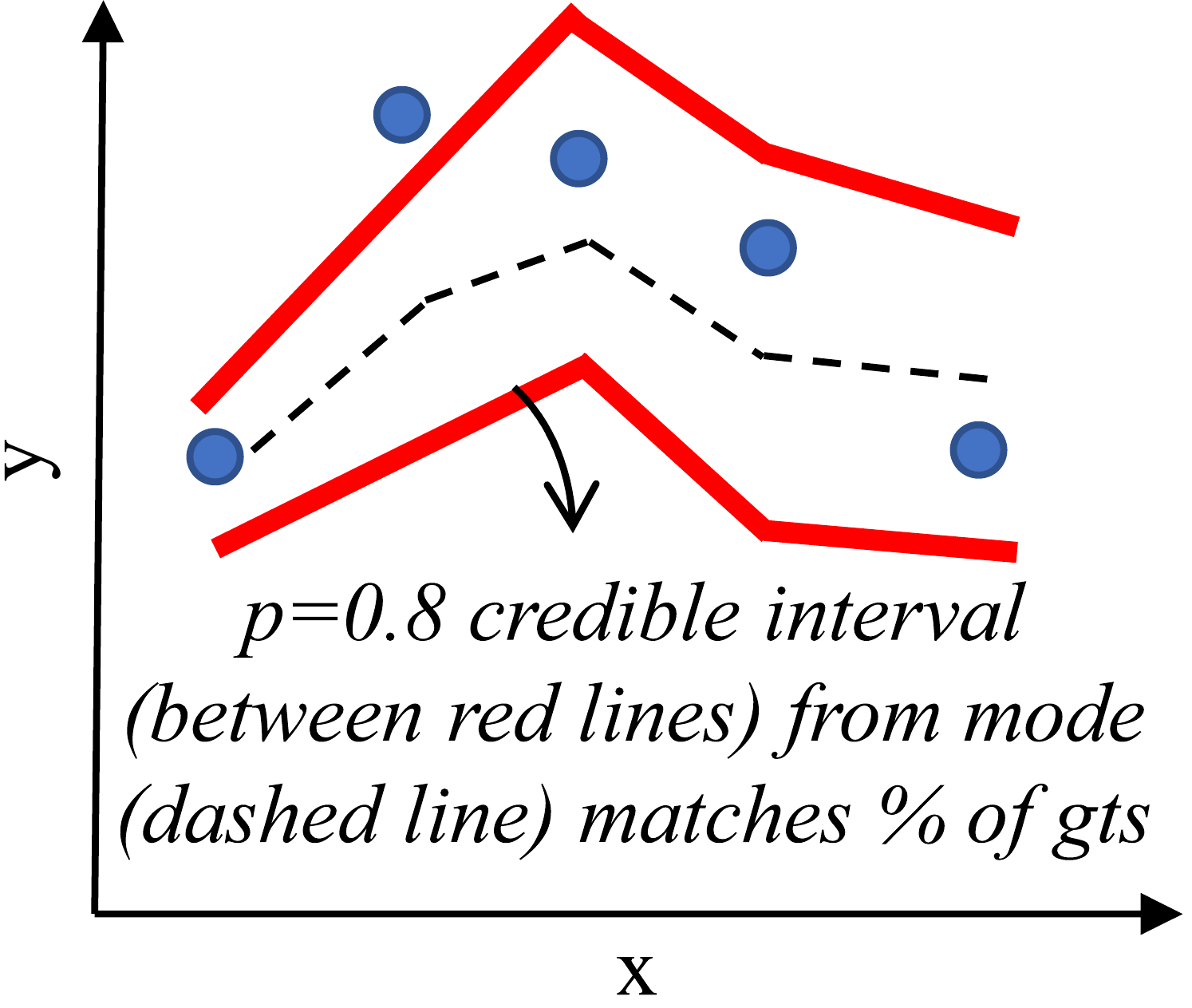}
            \caption{Regressor \cite{regressionunc}}
        \end{subfigure}
        \begin{subfigure}[b]{0.155\textwidth}
            \includegraphics[width=\textwidth]{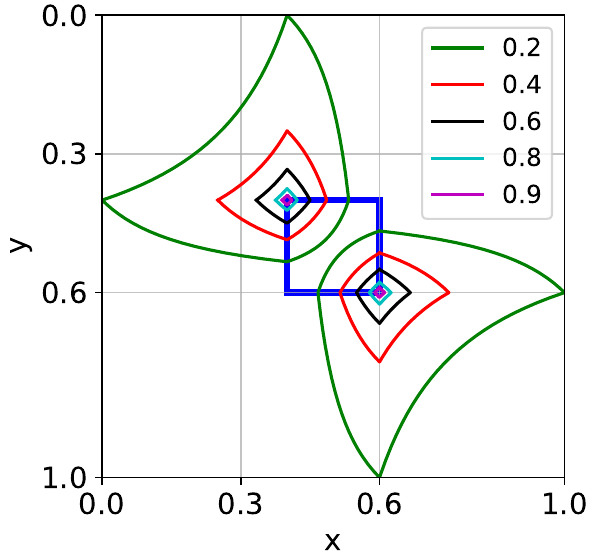}
            \caption{Detector}
        \end{subfigure}
        %
        %
        %
        \vspace{-4.0ex}
        \caption{(a) Calibrated classifier; (b) Calibrated Bayesian regressor, where empirical and predicted CDFs match; (c) Loci of constant \gls{IoU} boundary, e.g. any predicted box with top-left and bottom-right corners obtained from within the green loci has an \gls{IoU} $> 0.2$ with the blue box. The detector is calibrated if its confidence matches the classification \textit{and} the localisation quality.}
        \label{fig:calibration_def}
        \vspace{-3.0ex}
\end{figure}

\subsection{Localisation-aware ECE} \label{subsec:mECE}
To build an intuitive understanding and to appreciate the underlying complexity in developing a metric to quantify the calibration of an object detector, we first revisit its subtasks and briefly discuss what a calibrated classifier and a calibrated regressor correspond to.
For the former, a classifier is calibrated if its confidence matches its accuracy as illustrated in \cref{fig:calibration_def}(a). 
For calibrating Bayesian regressors, there are different definitions \cite{regressionunc,distributioncalibrationregression,UCE,UCE2}.
One notable definition \cite{regressionunc} requires aligning the predicted and the empirical cumulative distribution functions (cdf), implying $p\%$ credible interval from the mean of the predictive distribution should include $p\%$ of the ground truths for all $p \in [0,1]$ (\cref{fig:calibration_def}(b)).
Extending this definition to object detection is nontrivial due to the increasing complexity of the problem.
For example, a detection is represented by a tuple $\{\hat{c}_i, \hat{b}_i, \hat{p}_i\}$ with $\hat{b}_i \in \mathbb{R}^{4}$, which is not univariate as in \cite{regressionunc}.
%
Also, this definition to align the empirical and predicted cdfs does not consider the regression accuracy explicitly, and therefore not fit for our purpose.
Instead, we take inspiration from an alternative definition that aims to directly align the confidence with the regression accuracy \cite{UCE,UCE2}.
%
%

To this end, without loss of generality, we use \gls{IoU} as the measure of localisation quality for the detection boxes. 
Therefore, broadly speaking, if the detection confidence score $\hat{p}_i = 0.8$, then the localisation task is calibrated (ignoring the classification task for now)  if the average localisation performance (\gls{IoU} in our case) is $80\%$ over the entire dataset. 
%
%
To demonstrate, following~\cite{BBSampling} we plot the loci for fixed values of \gls{IoU} in \cref{fig:calibration_def}(c).
In this example, considering the blue-box to be the ground-truth, $\hat{p}_i = 0.2$ implies that a detector is calibrated if the detection box lie on the `green' loci corresponding to \gls{IoU} $= 0.2$.

Focusing back onto the joint nature of object detection, we say that an object detector
$f: X \mapsto \{\hat{c}_i, \hat{b}_i, \hat{p}_i\}^N$ is calibrated if the classification and the localisation performances \textit{jointly} match its confidence $\hat{p}_i$. More formally, 
{\small
\begin{align}\label{eq:gen_calibration_}
    \underbrace{\mathbb{P}(\hat{c}_i = c_i | \hat{p}_i)}_{\textit{Classification perf.}} \underbrace{\mathbb{E}_{\hat{b}_i \in B_i(\hat{p}_i)}[ \mathrm{IoU}(\hat{b}_i, b_{\psi(i)})]}_{\textit{Localisation perf.}} = \hat{p}_i, \forall \hat{p}_i \in [0,1]
\end{align}
}%
where $B_i(\hat{p}_i)$ is the set of \gls{TP} boxes with the confidence score of $\hat{p}_i$, and $b_{\psi(i)}$ is the ground-truth box that $\hat{b}_i$ matches with. Note that in the absence of localisation quality, the above calibration formulation boils down to the standard classification calibration definition. 

For a given $B_i(\hat{p}_i)$, the first term in Eq. \eqref{eq:gen_calibration_}, $\mathbb{P}(\hat{c}_i = c_i | \hat{p}_i)$, is the ratio of the number of correctly-classified to the total number of detections, which is simply the precision. Whereas, the second term represents the average localisation quality of the boxes in $B_i(\hat{p}_i)$.
%
%


%
%

Following the approximations used to define the well-known \gls{ECE}, we use Eq. \eqref{eq:gen_calibration_} to define \gls{LaECE}. Precisely, we discretize the confidence score space into $J=25$ equally-spaced bins~\cite{calibration,verifiedunccalibration}, and to prevent more frequent classes to dominate the calibration error, we compute the average calibration error for each class separately~\cite{FocalLoss_Calibration,verifiedunccalibration}. 
%
Thus, the calibration error for the $c$-th class is obtained as
{\small
\begin{align}
\label{eq:odcalibrationfinal1}
   \mathrm{LaECE}^c 
   & = \sum_{j=1}^{J} \frac{|\hat{\mathcal{D}}^{c}_j|}{|\hat{\mathcal{D}}^{c}|} \left\lvert \bar{p}^{c}_{j} - \mathrm{precision}^{c}(j) \times \bar{\mathrm{IoU}}^{c}(j)  \right\rvert,
\end{align}}
where $\hat{\mathcal{D}}^{c}$ denotes the set of all detections, $\hat{\mathcal{D}}_j^{c}	\subseteq \hat{\mathcal{D}}^{c}$ is the set of detections in bin $j$ and $\bar{p}_{j}^{c}$ is the average of the detection confidence scores in bin $j$ for class $c$.
Furthermore, $\mathrm{precision}^{c}(j)$ denotes the precision of the $j$-th bin for $c$-th class and $\bar{\mathrm{IoU}}^{c}(j)$ the average \gls{IoU} of \gls{TP} boxes in bin $j$.
Then, $\mathrm{LaECE}$ is computed as the average of $\mathrm{LaECE}^c$ over all the classes.
%
We highlight that for the sake of better accuracy the recent detectors \cite{IoUNet,KLLoss,FCOS,GFL,varifocalnet,GFLv2,aLRPLoss,RSLoss,RankDetNet,paa,maskscoring,yolact-plus} tend to obtain $\hat{p}_i$ by combining the classification confidence with the localisation confidence (e.g., obtained from an auxiliary IoU prediction head), which is very well aligned with our LaECE formulation, enforcing $\hat{p}_i$ to match with the joint performance in Eq. \eqref{eq:odcalibrationfinal1}.
%
%

\begin{figure}[t]
        \captionsetup[subfigure]{}
        \centering
        \begin{subfigure}[b]{0.2\textwidth}
        \includegraphics[width=\textwidth]{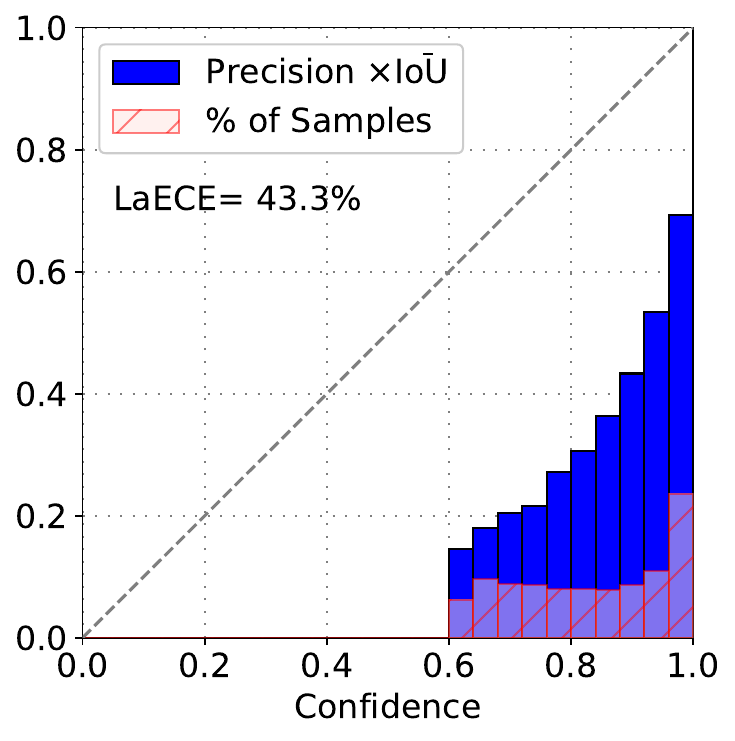}
        \caption{Base model}
        \end{subfigure}
        \begin{subfigure}[b]{0.2\textwidth}
        \includegraphics[width=\textwidth]{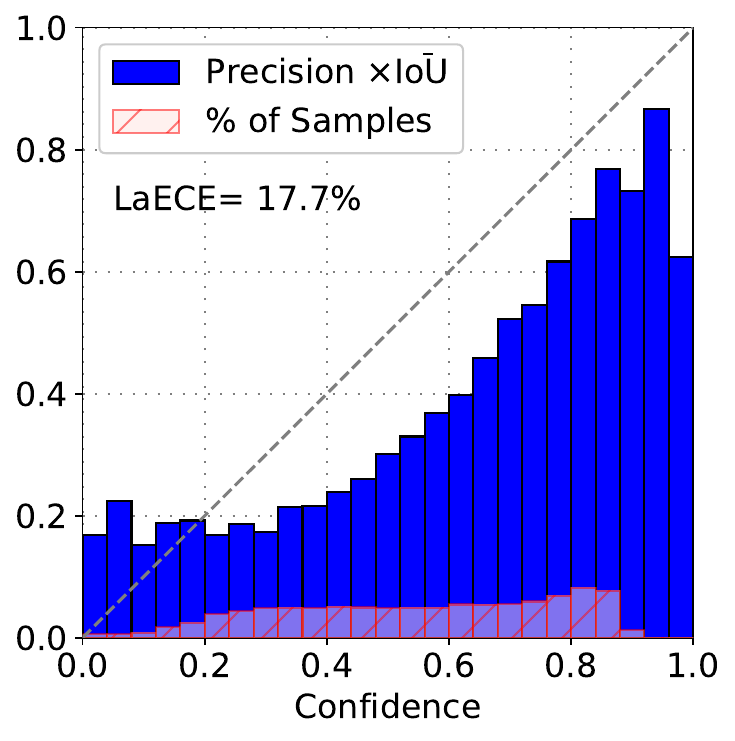}
        \caption{Calibrated by LR}
        \end{subfigure}
        \caption{Reliability diagrams of F-RCNN on $\mathcal{D}_{ID}$ for SAOD-Gen before and after calibration.}
        \label{fig:frcnnreliabilityhist}
        \vspace{-3ex}
\end{figure}

\textbf{Reliability Diagrams} We also produce reliability diagrams to provide insights on the calibration properties of a detector (\cref{fig:frcnnreliabilityhist}(a)).
To obtain a reliability diagram, we first obtain the performance, measured by the product of precision and \gls{IoU} (Eq. \eqref{eq:odcalibrationfinal1}), for each class over bins and then average the performance over the classes by ignoring the empty bins. 
Note that if a detector is perfectly calibrated with $\mathrm{LaECE}=0$, then all the histograms will lie along the diagonal in the reliability diagram since $\mathrm{LaECE}^c=0$.
Similar to classification, if the performance tends to be lower than the diagonal, then the detector is said to be over-confident as in Fig. \ref{fig:frcnnreliabilityhist}(a), and vice versa for an under-confident detector. Please see Fig. \ref{fig:frcnnreliabilityavod} for more examples. 

\subsection{Impact of Top-k Survival on Calibration} \label{subsec:relation}



%
%
Top-$k$ survival, a critical part of the post-processing step, selects $k$ detections with the highest confidence in an image. 
The value of $k$ is typically significantly larger than the number of objects, for example, $k=100$ for COCO where an average of only 7.3 ground-truth objects exist per image on the val set.
Therefore, the final detections may contain numerous low-scoring noisy detections. 
In fact, ATSS on COCO val set, for example, produces 86.4 detections on average per image after postprocessing, far more than the average number of objects per image.

Since these extra noisy detections do not impact on the widely used \gls{AP}, most works do not pay much attention to them, however, as we show below, they do have a negative impact on the calibration metric.
Thus, this may mislead a practitioner in choosing the wrong model when it comes to calibration quality. 

%

\begin{figure}[t]
        \captionsetup[subfigure]{}
        \centering
        \begin{subfigure}[b]{0.235\textwidth}
        \includegraphics[width=\textwidth]{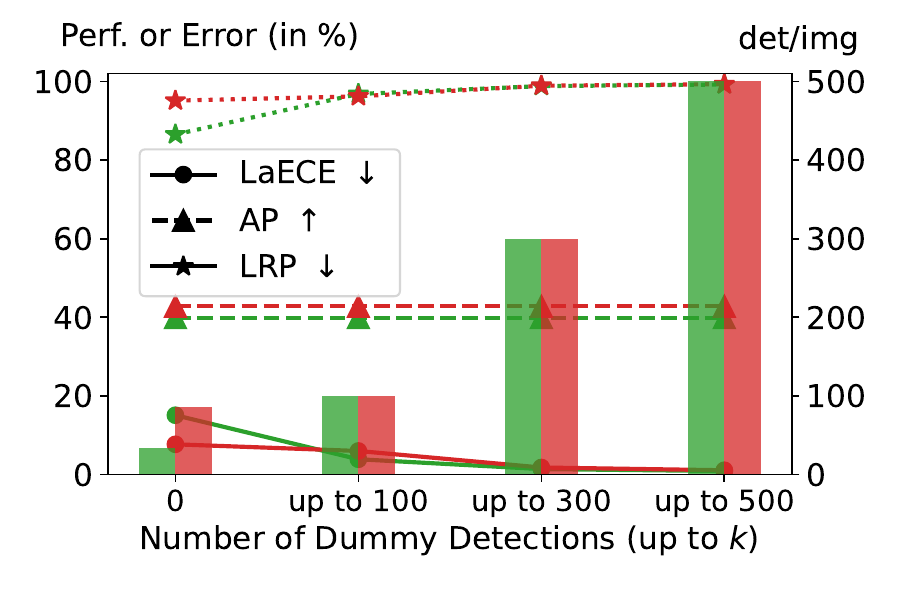}
        \caption{Adding dummy detections}
        \end{subfigure}
        \begin{subfigure}[b]{0.235\textwidth}
        \includegraphics[width=\textwidth]{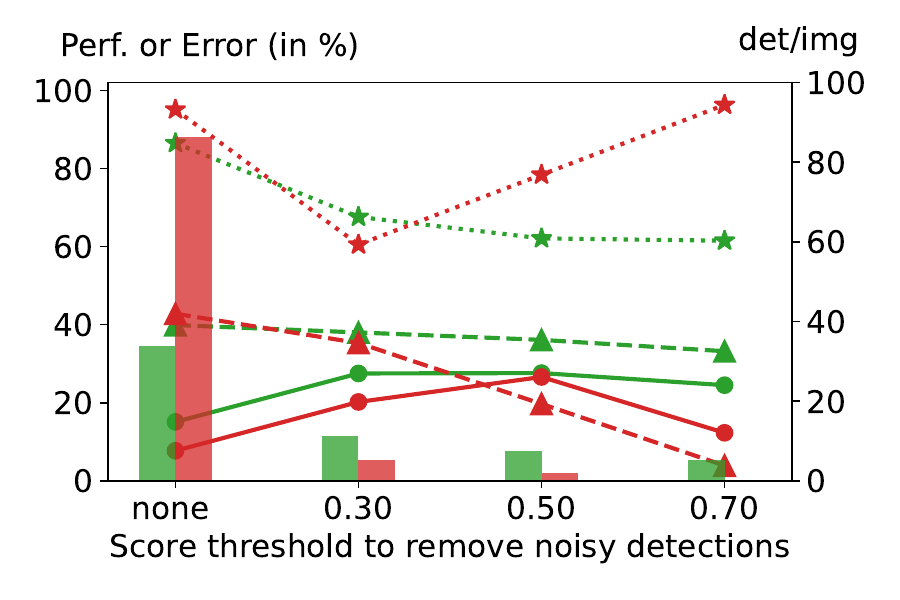}
        \caption{Thresholding detections}
        \end{subfigure}
        \caption{\textcolor{BrickRed}{Red}: ATSS, \textcolor{OliveGreen}{green}: F-RCNN, histograms present det/img using right axes, results are on COCO val set with 7.3 objects/img. \textbf{(a)} Dummy detections decrease \gls{LaECE} (solid line) artificially with no effect on AP (dashed line). \gls{LRP} (dotted line), on the other hand, penalizes dummy detections. \textbf{(b)} AP is maximized with more detections (threshold `none') while LRP Error benefits from properly-thresholded detections.
        %
        (refer App. \ref{app:calibration})
        }
        \label{fig:calibration}
        \vspace{-3ex}
\end{figure}

We design a synthetic experiment to show the impact of low-scoring noisy detections on \gls{AP} and calibration (\gls{LaECE}). Specifically, if the number of final detections is less than $k$ in an image, we insert ``dummy'' detections into the remaining space.
These dummy detections are randomly assigned a class $\hat{c}_i$, $\hat{p}_i = 0$, and only one pixel to ensure that they do \textit{not} match with any object. Hence, by design, they are ``perfectly calibrated''.
As shown in \cref{fig:calibration}(a), though these dummy detections have \textit{no} impact on the \gls{AP} (mathematical proof in App. \ref{app:calibration}), they do give an impression that the model becomes more calibrated (lower \gls{LaECE}) as $k$ increases. 
%
%
%
%
Therefore, considering that extra noisy detections are undesirable in practice, \textit{we do not advocate top-$k$ survival}, instead, we motivate the need to select a detection confidence threshold $\bar{v}$, where detections are rejected if their confidence is lower than $\bar{v}$. 

An appropriate choice of $\bar{v}$ should produce a set of thresholded-detections with a good balance of precision, recall and localisation errors\footnote{Using properly-thresholded detections is in fact similar to the Panoptic Segmentation, which is a closely-related task to object detection \cite{PanopticSegmentation,PanopticFPN}}.
In \cref{fig:calibration}(b), we present the effect of $\bar{v}$ on \gls{LRP}, where the lowest error is obtained around 0.30 for ATSS and 0.70 for F-RCNN, leading to an average of 6 detections/image for both detectors, far closer to the average number of objects compared to using $k=100$.
Consequently, to obtain $\bar{v}$ for our baseline, we use LRP-optimal thresholding \cite{LRP,LRPPAMI}, which is the threshold achieving the minimum \gls{LRP} for each class on the val set.

\begin{table*}
    \centering
    \setlength{\tabcolsep}{0.07em}
    \def\r{\hspace*{1.0ex}}
    \small
    \caption{Effect of post hoc calibration on LaECE and LRP (in \%). \xmark: Uncalibrated, HB: Histogram binning, IR: Isotonic Regresssion, LR: Linear Regression. ATSS, combining localisation and classification confidences using multiplication as in our LaECE (Eq. \eqref{eq:odcalibrationfinal1}), performs the best on both datasets before/after calibration. Aligned with \cite{FocalLoss_Calibration}, uncalibrated F-RCNN using cross-entropy loss performs the worst.}
    \vspace{-1.5ex}
    \label{tab:calibratingod}
    \scalebox{0.85}{
    \begin{tabular}{c@{\r}|@{\r}c@{\r}c@{\r}c@{\r}c@{\r}|@{\r}c@{\r}c@{\r}c@{\r}c@{\r}|@{\r}c@{\r}c@{\r}c@{\r}c@{\r}|@{\r}c@{\r}c@{\r}c@{\r}c@{\r}|@{\r}c@{\r}c@{\r}c@{\r}c@{\r}|@{\r}c@{\r}c@{\r}c@{\r}c@{\r}}
    \toprule
    \midrule
    Dataset &\multicolumn{16}{c@{\r}|@{\r}}{SAOD-Gen}&\multicolumn{8}{c}{SAOD-AV} \\
    \midrule
    Detector &\multicolumn{4}{c@{\r}|@{\r}}{F-RCNN}&\multicolumn{4}{c@{\r}|@{\r}}{RS-RCNN}&\multicolumn{4}{c@{\r}|@{\r}}{ATSS}&\multicolumn{4}{c@{\r}|@{\r}}{D-DETR}&\multicolumn{4}{c@{\r}|@{\r}}{F-RCNN}&\multicolumn{4}{c}{ATSS} \\
    Calibrator&\xmark&LR&HB&IR&\xmark&LR&HB&IR&\xmark&LR&HB&IR&\xmark&LR&HB&IR&\xmark&LR&HB&IR&\xmark&LR&HB&IR \\
    \midrule
    LaECE&$43.3$&$17.7$&$18.6$&$\mathbf{16.9}$&$32.0$&$17.4$&$19.6$&$\mathbf{17.2}$&$\mathbf{15.7}$&$16.8$&$18.7$&$16.7$&$15.9$&$\mathbf{15.7}$&$17.7$&$15.9$&$26.5$&$\mathbf{9.8}$&$10.2$&$10.2$&$16.8$&$\mathbf{9.0}$&$9.7$&$9.7$ \\
    LRP&$74.7$&$74.7$&$74.7$&$74.7$&$73.6$&$73.6$&$73.6$&$73.6$&$74.0$&$74.0$&$74.1$&$74.0$&$71.9$&$71.9$&$71.9$&$71.9$&$73.5$&$73.5$&$73.5$&$73.5$&$70.6$&$70.6$&$70.6$&$70.6$  \\
    %
    %
    \midrule
    \bottomrule
    \end{tabular}
    }
    \vspace{-3.0ex}
\end{table*}

\subsection{Post hoc Calibration of Object Detectors} \label{subsec:calibrationmethods}
For our baseline, given that \gls{LaECE} provides the calibration error of the model, we can calibrate an object detector using common calibration approaches from the classification and regression literature.
Precisely, for each class, we train a calibrator $\zeta^c:[0,1] \rightarrow [0,1]$ using the input-target pairs ($\{\hat{p}_i, t^{cal}_i\}$) from $\valdata$, where $t^{cal}_i$ is the target confidence.
As shown in App \ref{app:calibration}, \gls{LaECE} for bin $j$ reduces to
{\small
\begin{align}
\label{eq:finalECEoptimise_main}
  &\Bigg\lvert \sum_{\substack{\hat{b}_i \in \hat{\mathcal{D}}_j^c \\ \psi(i) > 0}} \big( t^{cal}_i - \mathrm{IoU}(\hat{b}_i, b_{\psi(i)}) \big) + \sum_{\substack{\hat{b}_i \in \hat{\mathcal{D}}_j^c \\ \psi(i) \leq 0}} t^{cal}_i  \Bigg\rvert.
\end{align}
}
Consequently, we seek $t^{cal}_i$ which minimises this value assuming that $\hat{p}_i$ resides in the $j$th bin.
In situations where the prediction is a TP ($\psi(i) > 0)$, Eq. \eqref{eq:finalECEoptimise_main} is minimized when $\hat{p}_{i}=t^{cal}_i=\mathrm{IoU}(\hat{b}_i, b_{\psi(i)})$ and conversely, if $\psi(i) \leq 0$, it is minimised when $\hat{p}_{i}=t^{cal}_i=0$.
%
%
We then train linear regression (LR); histogram binning (HB)~\cite{zadrozny2001obtaining}; and isotonic regression (IR)~\cite{zadrozny2002transforming} models with such pairs.
%
\cref{tab:calibratingod} shows that these calibration methods improve \gls{LaECE} in five out of six cases, and in the case where they do not improve (ATSS on SAOD-Gen), the calibration performance of the base model is already good. 
Overall, we find IR and LR perform better than HB and consequently we employ LR for \gls{SAODet}s since LR performs the best on three detectors.
\cref{fig:frcnnreliabilityhist}(b) shows an example reliability histogram after applying LR, indicating the improvement to calibration.

\section{Baseline \gls{SAODet}s and Their Evaluation} \label{sec:evaluation} \label{subsec:DAQ}
%
Using the necessary features developed in \cref{sec:ood} and \cref{sec:calibration}, namely, obtaining: image-level uncertainties, calibration methods as well as the thresholds $\bar{u}$ and $\bar{v}$, we now show how to convert standard detectors into ones that are self-aware. Then, we benchmark them using the \gls{SAOD} framework proposed in \cref{sec:saod} whilst leveraging our test datasets and \gls{LaECE}.

\begin{table}
    \def\s{\hspace*{0.5ex}}
    \def\r{\hspace*{1ex}}
    \centering
    \caption{ Evaluating \gls{SAODet}s. With higher BA and IDQs, SA-D-DETR achieves the best DAQ on SAOD-Gen. For SAOD-AV datasets, SA-ATSS outperforms SA-F-RCNN thanks to its higher IDQs. Bold:  \gls{SAODet} achieves the best, values are in \%.}
    \vspace{-1ex}
    \label{tab:evaluation}
    \scalebox{0.65}{
    \begin{tabular}{@{}c@{\r}c@{\r}|@{\r}c@{\r}|@{\r}c@{\r}|@{\r}c@{\s}c@{\s}c@{\r}|@{\r}c@{\s}c@{\s}c|@{\r}c@{\s}c}
    \toprule
    \midrule
         &Self-aware&\multirow{2}{*}{$\scriptstyle\mathrm{DAQ}\uparrow$}&$\ooddata$&\multicolumn{3}{c@{\r}|@{\r}}{$\indata$}&\multicolumn{3}{c@{\r}|@{\r}}{$\shiftdata$}&\multicolumn{2}{c}{$\valdata$} \\
         &Detector& &$\scriptstyle\mathrm{BA}\uparrow$&$\scriptstyle\mathrm{IDQ}\uparrow$&$\scriptstyle\mathrm{LaECE}\downarrow$&$\scriptstyle\mathrm{LRP}\downarrow$&$\scriptstyle\mathrm{IDQ}\uparrow$&$\scriptstyle\mathrm{LaECE}\downarrow$&$\scriptstyle\mathrm{LRP}\downarrow$&$\scriptstyle\mathrm{LRP}\downarrow$&$\scriptstyle\mathrm{AP}\uparrow$\\
    \midrule
    \multirow{4}{*}{\rotatebox[origin=c]{90}{Gen}}
    &SA-F-RCNN&$39.7$&$87.7$&$38.5$&$17.3$&$74.9$&$26.2$&$18.1$&$84.4$&$59.5$&$39.9$\\
    &SA-RS-RCNN&$41.2$&$\mathbf{88.9}$&$39.7$&$17.1$&$73.9$&$27.5$&$\mathbf{17.8}$&$83.5$&$58.1$&$42.0$\\
    &SA-ATSS&$41.4$&$87.8$&$39.7$&$16.6$&$74.0$&$27.8$&$18.2$&$83.2$&$58.5$&$42.8$\\
    &SA-D-DETR&$\mathbf{43.5}$&$\mathbf{88.9}$&$\mathbf{41.7}$&$\mathbf{16.4}$&$\mathbf{72.3}$&$\mathbf{29.6}$&$17.9$&$\mathbf{81.9}$&$\mathbf{55.9}$&$\mathbf{44.3}$\\
    \midrule
    \multirow{2}{*}{\rotatebox[origin=c]{90}{AV}} &SA-F-RCNN&$43.0$&$\mathbf{91.0}$&$41.5$&$9.5$&$73.1$&$28.8$&$7.2$&$83.0$&$54.3$&$55.0$\\
    &SA-ATSS&$\mathbf{44.7}$&$85.8$&$\mathbf{43.5}$&$\mathbf{8.8}$&$\mathbf{71.5}$&$\mathbf{30.8}$&$\mathbf{6.8}$&$\mathbf{81.5}$&$\mathbf{53.2}$&$\mathbf{56.9}$\\
    \midrule
    \bottomrule
    \end{tabular}
    }
    \vspace{-1ex}
\end{table}

\begin{table}
    \def\s{\hspace*{0.5ex}}
    \def\r{\hspace*{1ex}}
    \centering
    \caption{Ablation study by removing: LRP-Optimal thresholding~(\cref{subsec:relation}) for $\bar{v}=0.5$; LR calibration~(\cref{subsec:calibrationmethods}) for uncalibrated model; and image-level threshold $\bar{u}$~(\cref{sec:ood}) for the threshold corresponding to $\mathrm{TPR=0.95}$.}
    \vspace{-1ex}
    \label{tab:baselineablation}
    \scalebox{0.80}{
    \begin{tabular}{@{\r}c@{\r}|@{\r}c@{\r}|@{\r}c@{\r}|@{\r}c@{\r}|@{\r}c@{\r}|@{\r}c@{\r}|@{\r}c@{\r}|@{\r}c@{\r}|@{\r}c@{\r}}
    \toprule
    \midrule
    $\bar{v}$&$\scriptstyle\mathrm{LR}$&$\bar{u}$&$\scriptstyle\mathrm{DAQ}\uparrow$&$\scriptstyle\mathrm{BA}\uparrow$&$\scriptstyle\mathrm{LaECE}\downarrow$&$\scriptstyle\mathrm{LRP}\downarrow$&$\scriptstyle\mathrm{LaECE_T}\downarrow$&$\scriptstyle\mathrm{LRP_T}\downarrow$\\ \midrule
     &&&$36.0$&$83.2$&$42.7$&$76.2$&$44.1$&$84.7$\\ 
    \cmark& &&$36.5$&$83.2$&$41.7$&$\mathbf{74.8}$&$43.9$&$84.7$ \\
    \cmark&\cmark& &$39.1$&$83.2$&$\mathbf{17.2}$&$\mathbf{74.8}$&$\mathbf{18.1}$&$84.7$\\
    \cmark &\cmark&\cmark&$\mathbf{39.7}$&$\mathbf{87.7}$&$17.3$&$74.9$&$\mathbf{18.1}$&$\mathbf{84.4}$\\
    \midrule
    \bottomrule
    \end{tabular}
    }
    \vspace{-3ex}
\end{table}

\textbf{Baseline \gls{SAODet}s}
To address the requirements of a \gls{SAODet}, we make the following design choices when converting an object detector into one which is self aware:
The hard requirement of predicting whether or not to accept an image is achieved through obtaining image-level uncertainties by aggregating uncertainty scores.
Specifically, we use mean(top-3) and obtain an uncertainty threshold $\bar{u}$ through cross-validation using pseudo \gls{OOD} set approach (\cref{sec:ood}).
We only keep the detections with higher confidence than $\bar{v}$, which is set using LRP-optimal thresholding (\cref{subsec:relation}).
To calibrate the detection scores, we use linear regression as discussed in \cref{subsec:calibrationmethods}.
Thus, we convert all four  detectors that we use (\cref{sec:saod}) into ones that are self-aware, prefixed by a SA in the tables.
For further details, please see App. \ref{app:SAOD}.

\textbf{The \gls{SAOD} Evaluation Protocol}
The \gls{SAOD} task is a robust protocol unifying the evaluation of the: (i) reliability of uncertainties; (ii) the calibration and accuracy; (iii) and performance under domain shift.
To obtain quantitative values for the above, we leverage the Balanced Accuracy (\cref{sec:ood}) for (i).
For (ii) we evaluate the calibration and accuracy using \gls{LaECE}~(\cref{sec:calibration}) and the LRP~\cite{LRP} respectively, but combine them through the harmonic mean of $1-\mathrm{LRP}$ and $1-\mathrm{LaECE}$ on $X \in \indata$, which we define as the \gls{IDQ}.
Similarly, for (iii) we compute the IDQ for $X \in \mathcal{T}(\mathcal{D}_{\mathrm{ID}})$, denoted by $\mathrm{IDQ_T}$, but with the principal difference that the detector is flexible to accept or reject severe corruptions (C5) as discussed in \cref{sec:saod}.
Considering that all of these features are crucial in a safety-critical application, a lack of performance in one them needs to be heavily penalized. 
To do so, we introduce the \gls{DAQ}, a unified performance measure to evaluate \gls{SAODet}s, constructed as the the harmonic mean of $\mathrm{BA}$, $\mathrm{IDQ}$ and $\mathrm{IDQ_T}$.
The resulting DAQ is a higher-better measure with a range of $[0,1]$.

\textbf{Main Results}
Here we discuss how our \gls{SAODet}s perform in terms of the aforementioned metrics.
In terms of our hypotheses, the first evaluation we wish observe is the effectiveness of our metrics.
Specifically, we observe in \cref{tab:evaluation} that a lower \gls{LaECE} and \gls{LRP} lead to a higher \gls{IDQ}; and that a higher BA, \gls{IDQ} and \gls{IDQ}\textsubscript{T} lead to a higher \gls{DAQ}, indicating that the constructions of these metrics is appropriate.
To justify that they are reasonable, we observe that typically more complex and better performing detectors (DETR and ATSS) outperform the simpler F-RCNN, indicating that these metrics reflect the quality of the object detectors.

In terms of observing the performance of these self-aware variants, we can see that while recent state-of-the-art detectors perform very well in terms of \gls{LRP} and \gls{AP} on $\valdata$, their performance drops significantly as we expose them to our $\mathcal{D}_{ID}$ and $\mathcal{T}(\mathcal{D}_{ID})$ which involves domain shift, corruptions and OOD. 
We would also like to note that the best \gls{DAQ} corresponding to the best performing model SA-D-DETR still obtains a low score of $43.5\%$ on the SAOD-Gen dataset.
As this performance does not seem to be convincing, extra care should be taken before these models are deployed in safety-critical applications.
Consequently, our study shows that a significant amount of attention needs to be provided in building self-aware object detectors and effort to reduce the performance gap needs to be undertaken. 

\textbf{Ablation Analyses} To test which components of the \gls{SAODet} contribute the most to their improvement, we perform a simple experiment using SA-F-RCNN (SAOD-Gen).
In this experiment, we systematically remove the LRP-optimal thresholds; LR calibration; and pseudo-set approach and replace these features, with a detection-score threshold of 0.5; no calibration; and a threshold corresponding to a TPR of 0.95 respectively.
We can see in \cref{tab:baselineablation} that as hypothesized, LRP-optimal thresholding improves accuracy, LR yields notable gain in \gls{LaECE} and using pseudo-sets results in a gain for \gls{OOD} detection. 
In App. \ref{app:SAOD}, we further conduct additional experiments to (i) investigate the effect of $\bar{u}$ and $\bar{v}$ on reported metrics and (ii) how common improvement strategies for object detectors affect \gls{DAQ}.

\textbf{Evaluating Individual Robustness Aspects}
We finally note that our framework provides the necessary tools to evaluate a detector in terms of reliability of uncertainties, calibration and domain shift. 
Thereby enabling the researchers to benchmark either a \gls{SAODet} using our \gls{DAQ} measure or one of its individual components.
Specifically, (i) uncertainties can be evaluated on $\indata \cup \ooddata$ using \gls{AUROC} or \gls{BA} (\cref{tab:aggregate}); (ii) calibration can be evaluated on $\indata \cup \shiftdata$ using \gls{LaECE} (\cref{tab:calibratingod}); and (iii) $\indata \cup \shiftdata$ can be used to test detectors developed for single domain generalization~\cite{DGOD,clipgap}. 
\label{subsec:SAODEvaluation}

\vspace{-1ex}
\section{Conclusive Remarks}
\label{sec:conclusion}
\vspace{-1ex}
In this paper, we developed the \gls{SAOD} task, which requires detectors to obtain reliable uncertainties; yield calibrated confidences; and be robust to domain shift.
We curated large-scale datasets and introduced novel metrics to evaluate detectors on the \gls{SAOD} task.
Also, we proposed a metric (\gls{LaECE}) to quantify the calibration of object detectors which respects both classification \emph{and} localisation quality, addressing a critical shortcoming in the literature. 
We hope that this work inspires researchers to build more reliable object detectors for safety-critical applications.

%

%
%
%

%
%

{\small
\bibliographystyle{ieee_fullname}
\bibliography{egbib}
}
\clearpage

\section*{APPENDICES}
\tableofcontents
\renewcommand{\thefigure}{A.\arabic{figure}}
\renewcommand{\thetable}{A.\arabic{table}}
\renewcommand{\theequation}{A.\arabic{equation}}
\renewcommand{\thealgorithm}{A.\arabic{algorithm}}
\renewcommand{\thesection}{A}
\newpage
\section{Details of the Test Sets} \label{app:datasets}
This section provides the details of our test sets summarized in \cref{tab:datasets}.
To give a general overview, while constructing our datasets, we impose restrictions for a more principled evaluation:
\begin{compactitem}
\item We ensure that there is at least one ID object in the images of $\indata$ (and also in the ones in $\shiftdata$) to avoid a situation that an ID image does not include any ID object.
\item $\mathcal{D}_{\mathrm{OOD}}$ images does not include any foreground object. Besides, we use detection datasets with different ID classes (iNat, Obj365 and SVHN) than our $\mathcal{D}_{\mathrm{Train}}$ to promote OOD objects in OOD images.
\end{compactitem}
In the following, we present how we curate each of these test splits, that are (i) Obj45K and BDD45K as $\mathcal{D}_{\mathrm{ID}}$; (ii) Obj45K-C and BDD45K-C as $\mathcal{T}(\mathcal{D}_{\mathrm{ID}})$; and (iii) \oodname-OOD as $\mathcal{D}_{\mathrm{OOD}}$.

\subsection{Obj45K and BDD45K  Splits}
We construct $\indata$ from different but semantically similar datasets; thereby introducing domain-shift to be reflective of the challenges faced by detectors in practice such as distribution shifts over time or lack of data in a particular environment.
To do so, we employ Objects365\cite{Objects365} for our SAOD-Gen use-case using COCO as \gls{ID} data and BDD100K for our SAOD-AV use-case with nuImages comprising the \gls{ID} data.
In the following, we discuss the specific details how we constructed our Obj45K and BDD45K splits from these datasets.
\subsubsection{Obj45K Split}
We rely on Objects365 \cite{Objects365} to construct our Gen-OD ID test set.
Similar to COCO \cite{COCO}, which we use for training and validation in our Gen-OD setting, Objects365 is a general object detection dataset.
On the other hand, Object365 includes 365 different classes, which is significantly larger than the 80 different classes in COCO dataset.
Therefore, using images from Objects365 to evaluate a model trained on COCO requires a proper matching between the classes of COCO with those of Objects365.
Fortunately, by design, Objects365 already includes most of the classes of COCO in order to facilitate using these datasets together.
However, we inspect the classes in those datasets more thoroughly to prove a more proper one-to-many matching from COCO classes to Objects365 classes.
As an example, examining the objects labelled as \texttt{chair} in COCO dataset, we observe that wheelchairs also pertain to the \texttt{chair} class of COCO.
However, in Objects365 dataset, \texttt{Wheelchair} and \texttt{Chair} are different classes.
Therefore, in this case, we match \texttt{chair} class of COCO not only with \texttt{Chair} but also with \texttt{Wheelchair} of Objects365.
Having said that, we also note that due to high numbers of images and classes in those datasets, it is not practical to have a manual inspection over all images and classes.
In the following, we present our resulting matching between COCO and Objects365 classes:
\begin{lstlisting}[language=python]
'person':'Person',
'bicycle':'Bicycle',
'car':['Car', 'SUV','Sports Car',
'Formula 1 '],
'motorcycle':'Motorcycle',
'airplane':'Airplane',
'bus':'Bus',
'train':'Train',
'truck':['Truck','Pickup Truck',
'Fire Truck', 'Ambulance',
'Heavy Truck'],
'boat':['Boat', 'Sailboat', 'Ship'],
'traffic light':'Traffic Light',
'fire hydrant':'Fire Hydrant',
'stop sign':'Stop Sign',
'parking meter':'Parking meter',
'bench':'Bench',
'bird':['Wild Bird', 'Duck',
'Goose', 'Parrot', 'Chicken'],
'cat':'Cat',
'dog':'Dog',
'horse':'Horse',
'sheep':'Sheep',
'cow':'Cow',
'elephant':'Elephant',
'bear':'Bear',
'zebra':'Zebra',
'giraffe':'Giraffe',
'backpack':'Backpack',
'umbrella':'Umbrella',
'handbag':'Handbag/Satchel',
'tie':['Tie', 'Bow Tie'],
'suitcase': 'Luggage',
'frisbee':'Frisbee',
'skis':'Skiboard',
'snowboard':'Snowboard',
'sports ball':['Baseball', 'Soccer',
'Basketball', 'Billards',
'American Football', 'Volleyball',
'Golf Ball','Table Tennis ','Tennis'],
'kite':'Kite',
'baseball bat':'Baseball Bat',
'baseball glove':'Baseball Glove',
'skateboard':'Skateboard',
'surfboard':'Surfboard',
'tennis racket':'Tennis Racket',
'bottle':'Bottle',
'wine glass':'Wine Glass',
'cup':'Cup',
'fork':'Fork',
'knife':'Knife',
'spoon':'Spoon',
'bowl':'Bowl/Basin',
'banana':'Banana',
'apple':'Apple',
'sandwich':'Sandwich',
'orange':'Orange/Tangerine',
'broccoli':'Broccoli',
'carrot':'Carrot',
'hot dog':'Hot dog',
'pizza':'Pizza',
'donut':'Donut',
'cake':'Cake',
'chair':['Chair', 'Wheelchair'],
'couch':'Couch',
'potted plant':'Potted Plant',
'bed':'Bed',
'dining table':'Dinning Table',
'toilet':['Toilet', 'Urinal'],
'tv':'Moniter/TV',
'laptop':'Laptop',
'mouse':'Mouse',
'remote':'Remote',
'keyboard':'Keyboard',
'cell phone':'Cell Phone',
'microwave':'Microwave',
'oven':'Oven',
'toaster':'Toaster',
'sink':'Sink',
'refrigerator':'Refrigerator',
'book':'Book',
'clock':'Clock',
'vase':'Vase',
'scissors':'Scissors',
'teddy bear' : 'Stuffed Toy', 
'hair drier':'Hair Dryer',
'toothbrush':'Toothbrush'
\end{lstlisting}

Having matched the ID classes, we label the remaining classes of Objects365 either as ``OOD'' or ``ambiguous''.
Specifically, a class is labelled as \gls{OOD} if COCO classes (or nuImages classes that we are interested in) do not contain that class  and they will be discussed in Section \ref{subsec:OODdata}.
Subsequently, we label a class as an ambiguous class in the cases that we cannot confidently categorize the class neither as \gls{ID} nor as \gls{OOD}.
As an example, having examined quite a few COCO images with \texttt{bottle} class, we haven't observed a flask, which is an individual class of Objects365 (\texttt{Flask}).
Still, as there might be instances of flask labelled as \texttt{bottle} class in COCO, we categorize \texttt{Flask} class of Objects365 as ambiguous and do not use any of the images in Objects that has a  \texttt{Flask} object in it.
Following this, we identify the following 25 out of 365 classes in Objects365 as ambiguous:
\begin{lstlisting}[language=python]
'Nightstand', 'Desk', 'Coffee Table',
'Side Table', 'Watch', 'Stool',
'Machinery Vehicle', 'Tricycle',
'Carriage', 'Rickshaw', 'Van',
'Traffic Sign','Speed Limit Sign',
'Crosswalk Sign', 'Flower', 'Telephone',
'Tablet', 'Flask', 'Briefcase',
'Egg tart', 'Pie', 'Dessert', 'Cookies',
'Wallet/Purse'
\end{lstlisting}

\begin{figure*}[t]
        \captionsetup[subfigure]{}
        \centering
        \begin{subfigure}[b]{0.68\textwidth}
        \includegraphics[width=\textwidth]{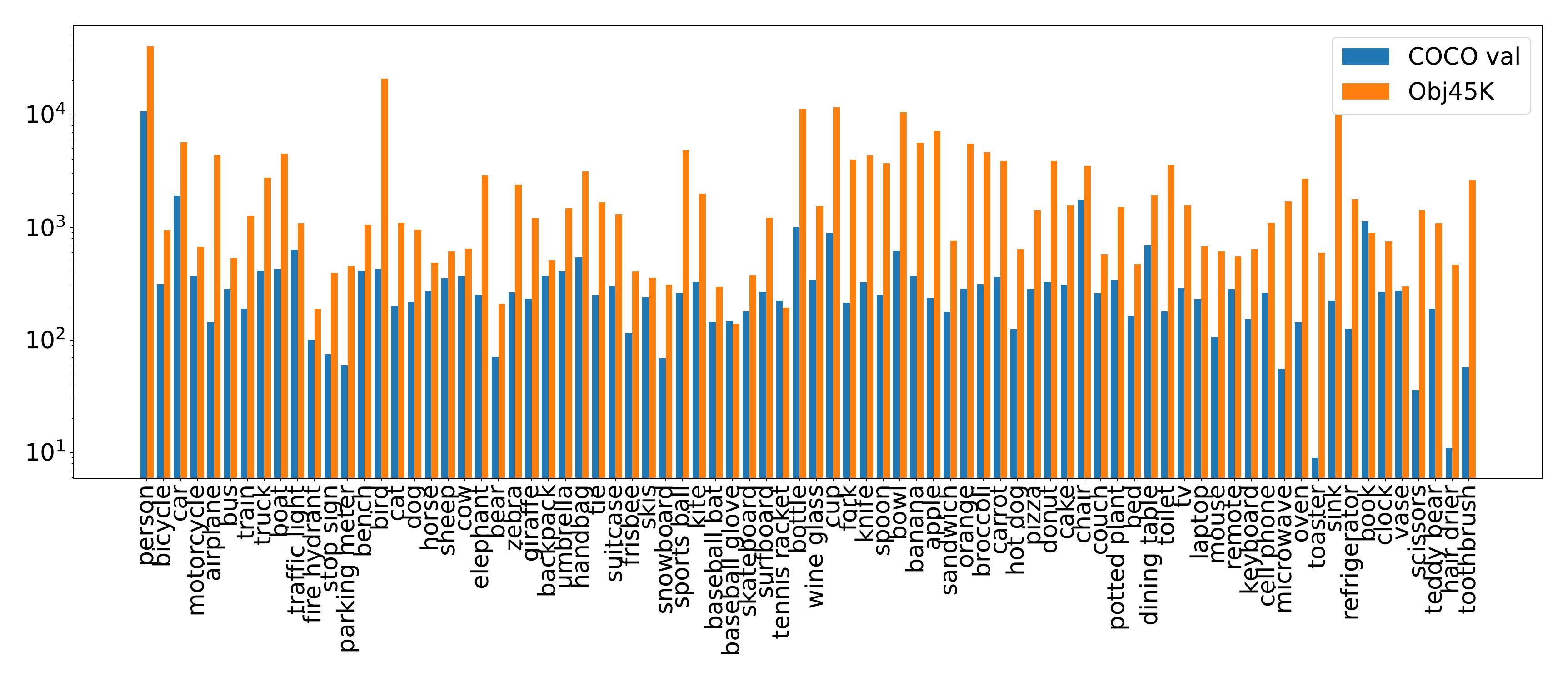}
        \caption{COCO val vs. Obj45K}
        \end{subfigure}
        \begin{subfigure}[b]{0.30\textwidth}
        \includegraphics[width=\textwidth]{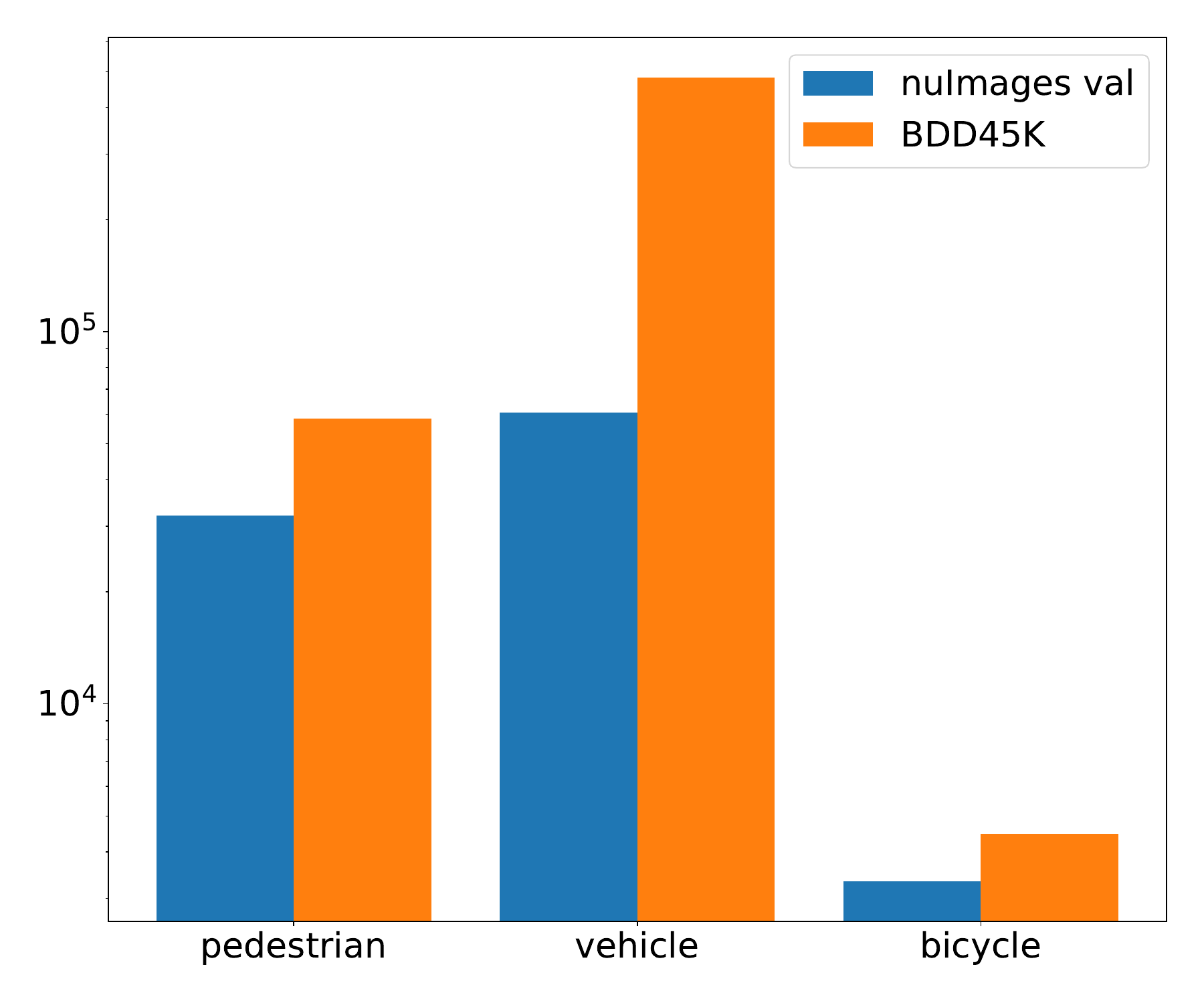}
        \vspace{8pt}
        \caption{nuImages val vs. BDD45K}
        \end{subfigure}
        \caption{Distribution of the objects over classes from our test sets and existing val sets. For both SAOD-Gen and SAOD-AV use-cases, our $\indata$ have more objects nearly for all classes to provide a thorough evaluation. Note that y-axes are in log-scale.}
        \label{fig:datacomp}
\end{figure*}

Finally, we collect 45K images for Obj45K split from validation set of Objects365 that contains (i) at least one ID object based on the one-to-many matching between classes of COCO and Objects365; and (ii) no object from an ambiguous class.
Compared to COCO val set with 5K images with 36K annotated objects, our Gen-OD ID test has 45K images with 237K objects, significantly outnumbering the val set which is commonly used to analyse and test the models mainly in terms of robustness aspects.
\cref{fig:datacomp}(a) compares the number of objects for Obj45K split and COCO val set, showing that the number of objects for each class of our Obj45K split is for almost classes (except 2 of 80 classes) larger than the COCO val set.
This large number of objects enables us to evaluate the models thoroughly.

\subsubsection{BDD45K Split}

\begin{figure*}[t]
        \captionsetup[subfigure]{}
        \centering
        \begin{subfigure}[b]{0.45\textwidth}
        \includegraphics[width=\textwidth]{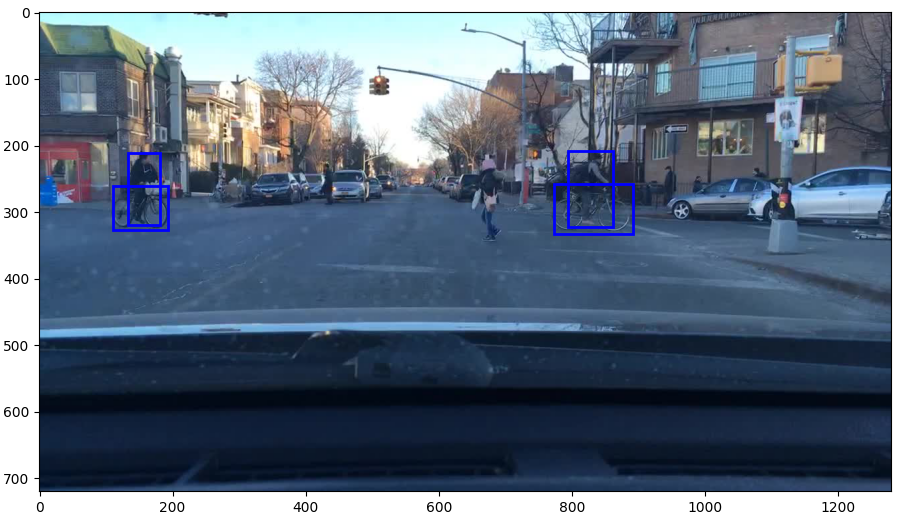}
        \caption{BDD100K-style annotations}
        \end{subfigure}
        \begin{subfigure}[b]{0.45\textwidth}
        \includegraphics[width=\textwidth]{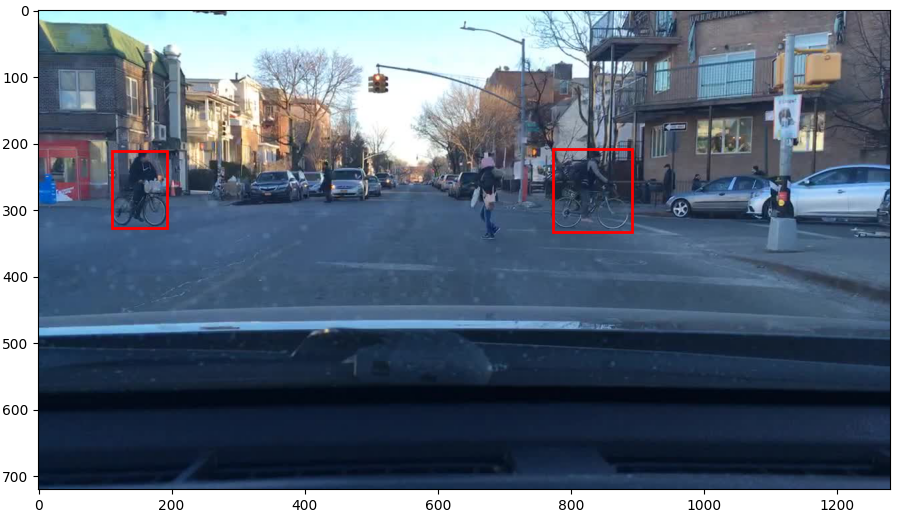}
        \caption{NuImages-style annotations (obtained by Hungarian matching)}
        \end{subfigure}
        \caption{Aligning the annotations of certain classes in BDD100K and nuImages datasets while curating our BDD45K test set. The riders and ridables (bicycles or motorcycles) need to be combined properly in (a). In this example, both of the rider objects are properly assigned to the corresponding bicycle objects by our simple method relying on Hungarian algorithm. In (b), which we use as a test image in our BDD45K, the bounding boxes are combined by finding the smallest enclosing bounding box and the objects are labelled as bicycles.}
        \label{fig:riders}
\end{figure*}

%
Considering that the widely-used AV datasets \cite{bdd100k,nuimages,kitti,waymo,Cityscapes} have \texttt{pedestrian}, \texttt{vehicle} and \texttt{bicycle} in common, we consider these three classes as ID classes of our SAOD-AV use-case\footnote{Accordingly, we the models for SAOD-AV for these three classes.}.
Then, similar to how we obtain Obj45K, we match these classes of nuImages with the classes of BDD100K, resulting in the following one-to-many matching:
%

%
%
\begin{lstlisting}[language=python]
'pedestrian': ['pedestrian',
"other person"],
'vehicle': ['car', 'truck', 'bus',
'motorcycle', 'train', "trailer",
"other vehicle"],
'bicycle': 'bicycle'
\end{lstlisting}

On the other hand, we observe a key difference in annotating \texttt{bicycle} and \texttt{motorcycle} classes between nuImages and BDD100K datasets. 
Specifically, while BDD100K has an additional class \texttt{rider} that is annotated separately from \texttt{bicycle} and \texttt{motorcycle} objects, the riders of bicycle and motorcycle are instead included in the annotated bounding box of \texttt{bicycle} and \texttt{motorcycle} objects in nuImages dataset.
In order to align the annotations of these classes between BDD100K and nuImages and provide a consistent evaluation, we aim to rectify the bounding box annotations of these classes in BDD100K dataset such that they follow the annotations of nuImages.
Particularly, there should be no rider class but \texttt{bicycle} and \texttt{motorcycle} objects include their riders in the resulting annotations.
To do so, we use a simple matching algorithm on BDD100K images to combine bicycle and motorcycle objects with their riders.
In particular, given an image, we identify first objects from \texttt{bicycle}, \texttt{motorcycle} and \texttt{rider} categories.
Then, we group \texttt{bicycle} and \texttt{motorcycle} objects as ``rideables'' and compute IoU between each rideable and rider object.
Given this matrix of representing the proximity between each rideable and rider object in terms of their IoUs, we assign riders to rideables by maximizing the total IoU using the Hungarian assignment algorithm \cite{Hungarian}.
Furthermore, we include a sanity check to avoid possible matching errors, e.g., in which a rideable object might be combined with a rider in a further location in the image due to possible annotation errors.
Specifically, our simple sanity is to require a minimum IoU overlap of $0.10$ between a rider and its assigned rideable in the resulting assignment from the Hungarian algorithm.
Otherwise, if any of the riders is assigned to a rideable object with an IoU less than $0.10$ in an image, we simply do not include this image in our BDD45K test set.
Finally, exploiting the assignment result, we obtain the bounding box annotation using the smallest enclosing bounding box including both the bounding box of the rider and that of the rideable object.
As for the category annotation of the object, we simply use the category of the rideable, which is either \texttt{bicycle} or \texttt{motorcycle}.
\cref{fig:riders} presents an example in which we convert BDD100K annotations of these specific classes into the nuImages format.
To validate our approach, we manually examine more than 2500 images in BDD45K test set and observe that it is effective to align the annotations of nuImages and BDD100K.

\begin{figure*}[t]
        \centering
        \includegraphics[width=0.9\textwidth]{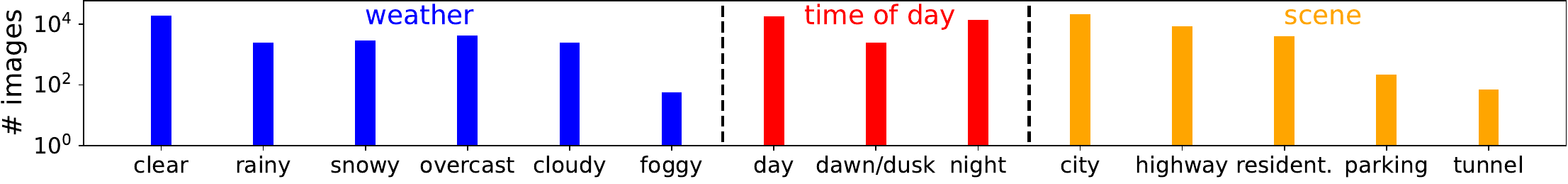}
        \caption{The diversity of BDD45K split in terms of weather, time of day and scene counts. 
        }
        \label{fig:bdd45k_cov}
\end{figure*}

Overall, using this strategy, we collect 45K images from training and validation sets of BDD100K and construct our BDD45K split.
We would like to highlight that our BDD45K dataset is diverse and extensive, where (i) it is larger compared to 16K images of nuImages val set; and (ii) it includes 543K objects in total, significantly larger than the number of objects from these 3 classes in nuImages val set with 96K objects.
Please refer to \cref{fig:datacomp}(b) for quantitative comparison.
In terms of diversity, our BDD45K ($\mathcal{D}_{\mathrm{ID}}$) comes from a different distribution than nuImages ($\mathcal{D}_{\mathrm{Train}}$); thereby introducing natural covariate shift.
\cref{fig:bdd45k_cov} illustrates that our BDD45K is very diverse and it is collected from different cities using different camera types than nuImages ($\mathcal{D}_{\mathrm{Train}}$).
As a result, as we will see in \cref{app:models}, the accuracy of the models drops significantly from $\valdata$ to $\indata$ even before the corruptions are employed.
We note that ImageNet-C corruptions are then applied to this dataset, further increasing the domain shift.

\subsection{Obj45K-C and BDD45K-C  Splits}
While constructing Obj45K-C and BDD45K-C as $\mathcal{T}(\mathcal{D}_{\mathrm{ID}})$, we use the following 15 different corruptions from 4 main groups \cite{hendrycks2019robustness}:
\begin{itemize}
    \item \textit{Noise.} gaussian noise, shot noise, impulse noise, speckle noise
    \item \textit{Blur.} defocus blur, motion blur, gaussian blur
    \item \textit{Weather.} snow, frost, fog, brightness
    \item \textit{Digital.} contrast, elastic transform, pixelate, jpeg compression
\end{itemize}
Then, given an image for a particular severity level that can be 1, 3 or 5, we randomly sample a transformation and apply to the image. 
In such a way, we obtain 3 different copies of Obj45K and BDD45K during evaluation.
%
%

\begin{figure*}[t]
        \captionsetup[subfigure]{}
        \centering
        \begin{subfigure}[b]{0.23\textwidth}
        \includegraphics[width=\textwidth]{Images/exampleimages/avodexample1_cleanimage_.jpg}
        \caption{Clean Image - BDD45K}
        \end{subfigure}
        \begin{subfigure}[b]{0.23\textwidth}
        \includegraphics[width=\textwidth]{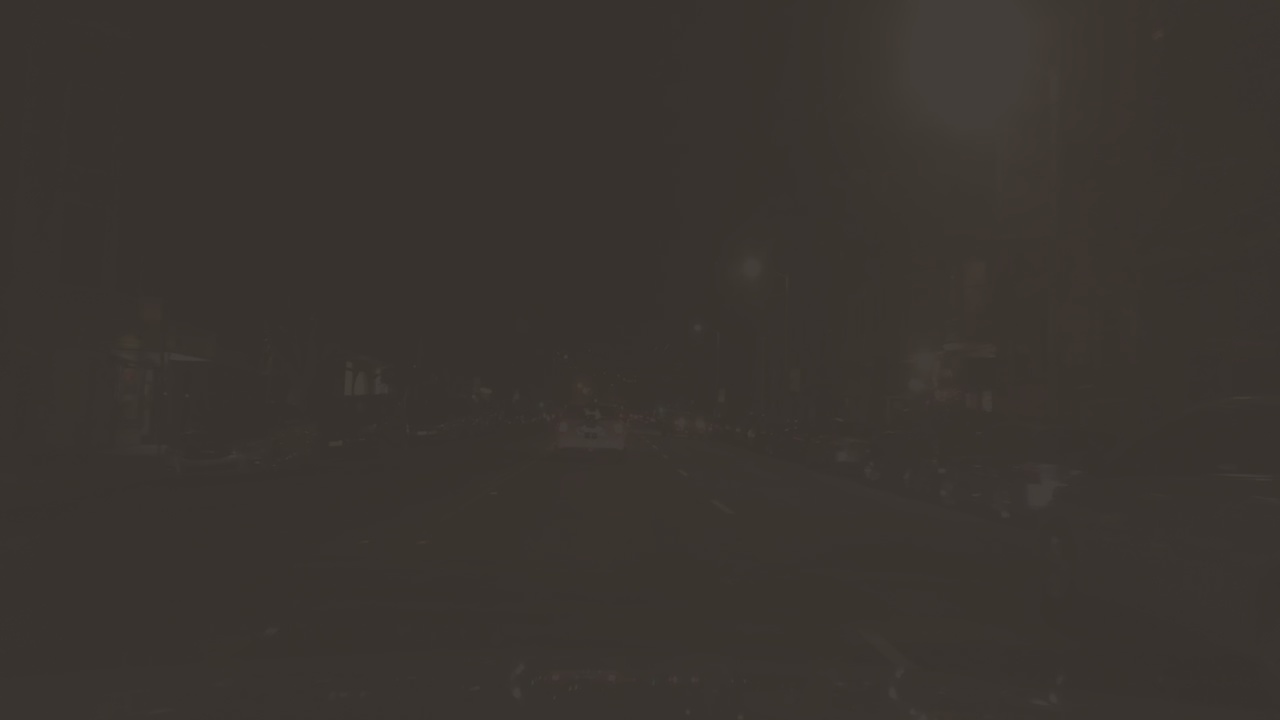}
        \caption{Constrast}
        \end{subfigure}
        \begin{subfigure}[b]{0.23\textwidth}
        \includegraphics[width=\textwidth]{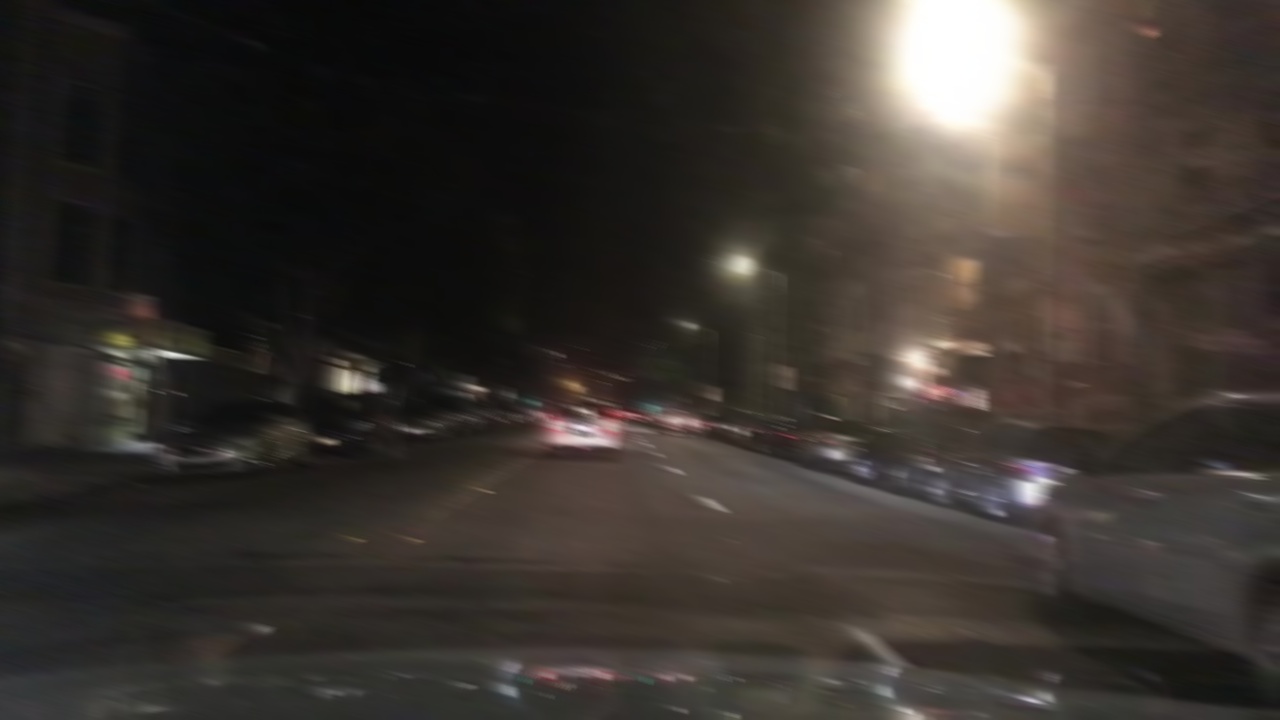}
        \caption{Motion Blur}
        \end{subfigure}
        \begin{subfigure}[b]{0.23\textwidth}
        \includegraphics[width=\textwidth]{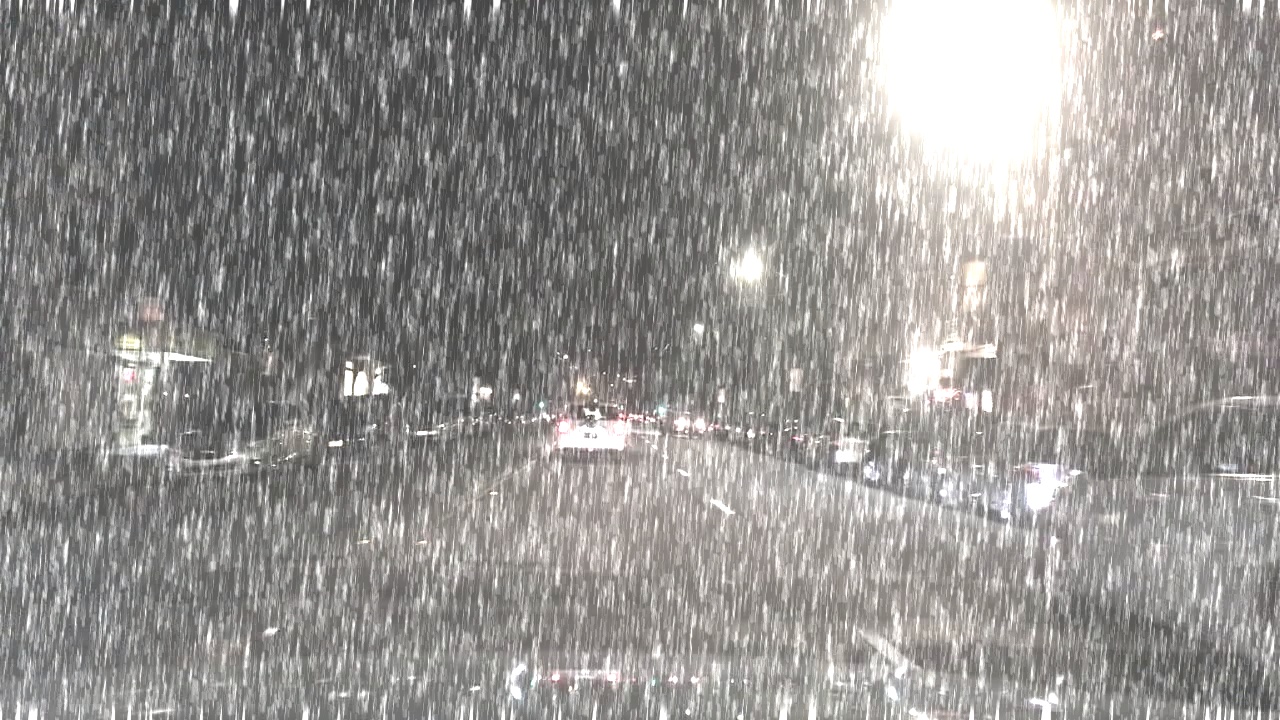}
        \caption{Snow}
        \end{subfigure}

        \begin{subfigure}[b]{0.23\textwidth}
        \includegraphics[width=\textwidth]{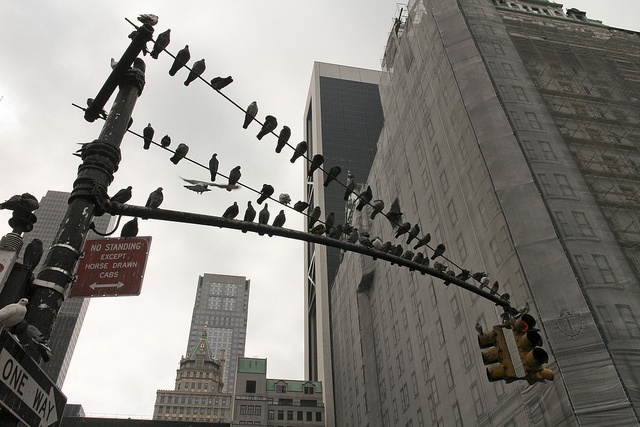}
        \caption{Clean Image - Obj45K}
        \end{subfigure}
        \begin{subfigure}[b]{0.23\textwidth}
        \includegraphics[width=\textwidth]{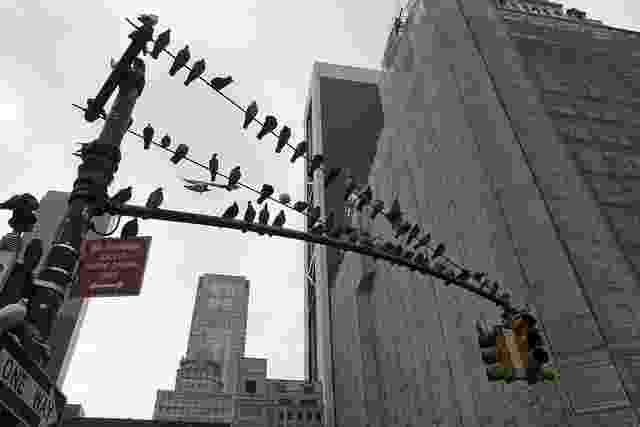}
        \caption{JPEG compression}
        \end{subfigure}
        \begin{subfigure}[b]{0.23\textwidth}
        \includegraphics[width=\textwidth]{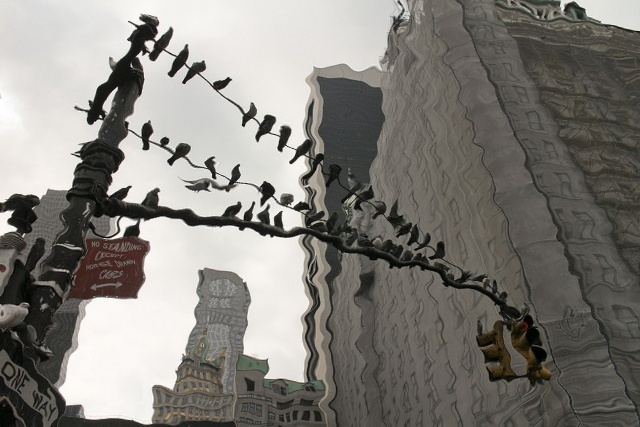}
        \caption{Elastic transform}
        \end{subfigure}
        \begin{subfigure}[b]{0.23\textwidth}
        \includegraphics[width=\textwidth]{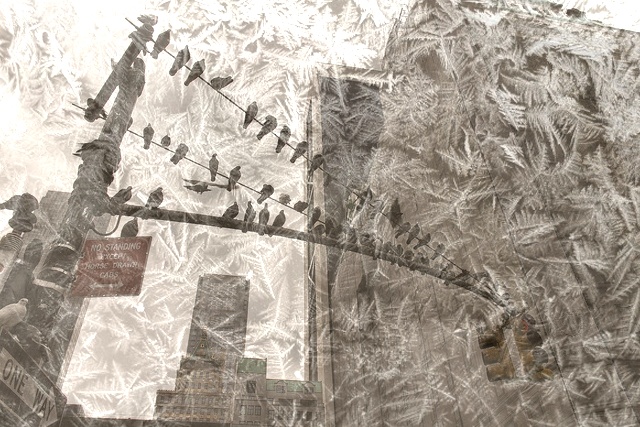}
        \caption{Frost}
        \end{subfigure}
        \caption{Clean and corrupted images using different transformations at severity 5 from AV-OD (upper row) and Gen-OD (lower row) use-cases. We do not penalize a detector if it can infer that it is uncertain and rejects such images with high corruption severity.}
        \label{fig:examplecorrimages}
\end{figure*}

We outline in the definition of the SAOD task (\cref{sec:saod}) that an image with a corruption severity 5 might not contain enough cues to perform object detection reliably and that a \gls{SAODet} is flexible to accept or reject such images as long as it yields accurate and calibrated detections on the accepted ones.
To provide insight of providing this flexibility, \cref{fig:examplecorrimages} presents example corruptions with severity 5.
Note that several cars in the corrupted images above and birds in the ones below are not visible any more due to the severity of the corruption.
As notable examples, some of the cars in \cref{fig:examplecorrimages}(b) and the birds in \cref{fig:examplecorrimages}(h) \cref{fig:examplecorrimages}(h) are not visible.
As a result, instead of enforcing the detector to predict all of the objects accurately, we do not penalize a detector as long as it can infer that it is uncertain and rejects such images with high corruption severity.

\subsection{\oodname-OOD Split} \label{subsec:OODdata}
This split is designed to evaluate the reliability of the uncertainties.
Following similar work~\cite{RegressionUncOD,VOS}, we ensure that the images in our OOD test set do not include any object from ID classes.
Specifically, in order to use \oodname-OOD within both SAOD-Gen and SAOD-AV datasets, we select an image to \oodname-OOD if the image does not include an object from either of the ID classes of Obj45K or BDD45K ($\indata$).
Then, we collect 110K images from three different detection datasets as detailed below:
\begin{itemize}
\item \textit{SVHN subset of \oodname-OOD.} We include all 46470 full numbers (not cropped digits) using both training and test sets of SVHN dataset in our OOD test set.

\item \textit{iNaturalist OOD subset of \oodname-OOD.} We use the validation set of iNaturalist 2017 object detection dataset to obtain our iNaturalist dataset.
Specifically, we include 28768 images in our OOD test set with the following classes:
\begin{lstlisting}
'Actinopterygii', 'Amphibia', 
'Animalia', 'Arachnida',
'Insecta', 'Mollusca', 'Reptilia'
\end{lstlisting}

\item \textit{Objects365 OOD subset of \oodname-OOD.} To select images for our OOD test set from Objects365 dataset, we use the following classes as OOD:
\begin{lstlisting}[language=python]
'Sneakers', 'Other Shoes', 'Hat',
'Lamp', 'Glasses', 'Street Lights',
'Cabinet/shelf', 'Bracelet', 
'Picture/Frame', 'Helmet', 'Gloves',
'Storage box', 'Leather Shoes', 'Flag',
'Pillow', 'Boots', 'Microphone',
'Necklace', 'Ring', 'Belt',
'Speaker', 'Trash bin Can', 'Slippers',
'Barrel/bucket', 'Sandals', 'Bakset',
'Drum', 'Pen/Pencil', 'High Heels',
'Guitar', 'Carpet', 'Bread', 'Camera',
'Canned', 'Traffic cone', 'Cymbal',
'Lifesaver', 'Towel', 'Candle', 
'Awning', 'Faucet', 'Tent', 'Mirror',
'Power outlet', 'Air Conditioner',
'Hockey Stick', 'Paddle', 'Ballon',
'Tripod', 'Hanger',
'Blackboard/Whiteboard','Napkin',
'Other Fish', 'Toiletry', 'Tomato',
'Lantern', 'Fan', 'Pumpkin',
'Tea pot', 'Head Phone', 'Scooter',
'Stroller', 'Crane', 'Lemon',
'Surveillance Camera', 'Jug', 'Piano',
'Gun', 'Skating and Skiing shoes',
'Gas stove', 'Strawberry',
'Other Balls', 'Shovel', 'Pepper',
'Computer Box', 'Toilet Paper',
'Cleaning Products', 'Chopsticks',
'Pigeon', 'Cutting/chopping Board',
'Marker', 'Ladder', 'Radiator',
'Grape', 'Potato', 'Sausage',
'Violin', 'Egg', 'Fire Extinguisher',
'Candy', 'Converter', 'Bathtub',
'Golf Club', 'Cucumber',
'Cigar/Cigarette ', 'Paint Brush',
'Pear', 'Hamburger', 
'Extention Cord', 'Tong', 'Folder',
'earphone', 'Mask', 'Kettle', 
'Swing', 'Coffee Machine', 'Slide',
'Onion', 'Green beans', 'Projector', 
'Washing Machine/Drying Machine',
'Printer', 'Watermelon', 'Saxophone',
'Tissue', 'Ice cream', 'Hotair ballon',
'Cello', 'French Fries', 'Scale',
'Trophy', 'Cabbage', 'Blender',
'Peach', 'Rice', 'Deer', 'Tape', 
'Cosmetics', 'Trumpet', 'Pineapple',
'Mango', 'Key', 'Hurdle',
'Fishing Rod', 'Medal', 'Flute',
'Brush', 'Penguin', 'Megaphone',
'Corn', 'Lettuce', 'Garlic', 
'Swan', 'Helicopter', 'Green Onion',
'Nuts', 'Induction Cooker',
'Broom', 'Trombone', 'Plum',
'Goldfish', 'Kiwi fruit',
'Router/modem', 'Poker Card',
'Shrimp', 'Sushi', 'Cheese',
'Notepaper', 'Cherry', 'Pliers',
'CD', 'Pasta', 'Hammer',
'Cue', 'Avocado', 'Hamimelon',
'Mushroon', 'Screwdriver', 'Soap',
'Recorder', 'Eggplant',
'Board Eraser', 'Coconut',
'Tape Measur/ Ruler', 'Pig',
'Showerhead', 'Globe', 'Chips',
'Steak', 'Stapler', 'Campel', 
'Pomegranate', 'Dishwasher',
'Crab', 'Meat ball', 'Rice Cooker',
'Tuba', 'Calculator',
'Papaya', 'Antelope', 'Seal',
'Buttefly', 'Dumbbell',
'Donkey', 'Lion', 'Dolphin',
'Electric Drill', 'Jellyfish',
'Treadmill', 'Lighter',
'Grapefruit', 'Game board',
'Mop', 'Radish',
'Baozi', 'Target', 'French',
'Spring Rolls', 'Monkey', 'Rabbit',
'Pencil Case', 'Yak',
'Red Cabbage', 'Binoculars',
'Asparagus', 'Barbell',
'Scallop', 'Noddles',
'Comb', 'Dumpling',
'Oyster', 'Green Vegetables',
'Cosmetics Brush/Eyeliner Pencil',
'Chainsaw', 'Eraser', 'Lobster',
'Durian', 'Okra', 'Lipstick',
'Trolley', 'Cosmetics Mirror',
'Curling', 'Hoverboard',
'Plate', 'Pot',
'Extractor', 'Table Teniis paddle'
\end{lstlisting}
Using both training and validation sets of Objects365, we collect 35190 images that \textit{only} contains objects from above classes.

\end{itemize}

Consequently, our resulting \oodname-OOD is both diverse and extensive compared to the datasets introduced in previous work~\cite{RegressionUncOD,VOS} which includes around 1-2K images and is collected from a single dataset.
\renewcommand{\thesection}{B}
\section{Details of the Used Object Detectors} \label{app:models}
%
%
\begin{table}
    \centering
    \small
    \setlength{\tabcolsep}{0.1em}
    \caption{COCO-style AP of the used object detectors on validation set ($\valdata$) and test set ($\indata$), along with their corrupted versions ($\mathcal{T}(\valdata)$ and $\mathcal{T}(\indata)$).}
    \label{tab:general_od_id_corruption}
    \begin{tabular}{|c|c||c|c|c|c||c|c|c|c|} \hline
         \multirow{2}{*}{Dataset}&\multirow{2}{*}{Detector}&\multirow{2}{*}{$\valdata$}&\multicolumn{3}{|c||}{$\mathcal{T}(\valdata)$}&\multirow{2}{*}{$\indata$}&\multicolumn{3}{|c|}{$\mathcal{T}(\mathcal{D}_{\mathrm{ID}})$}\\\cline{4-6} \cline{8-10}
         &&&C1&C3&C5&&C1&C3&C5 \\ \hline
    &F-RCNN&$39.9$&$31.3$&$20.3$&$10.8$&$27.0$&$20.3$&$12.8$&$6.9$\\
    &RS-RCNN&$42.0$&$33.7$&$21.8$&$11.6$&$28.6$&$21.7$&$13.7$&$7.3$\\
    SAOD&ATSS&$42.8$&$33.9$&$22.3$&$11.9$&$28.8$&$22.0$&$14.0$&$7.3$\\
    Gen&D-DETR&$44.3$&$36.2$&$24.0$&$12.2$&$30.5$&$23.4$&$15.4$&$8.0$\\  \cline{2-10}
    &NLL-RCNN&$40.1$&$31.0$&$20.0$&$11.6$&$26.9$&$20.3$&$12.9$&$6.8$ \\
    &ES-RCNN&$40.3$&$31.6$&$20.3$&$11.7$&$27.2$&$20.6$&$13.0$&$6.9$\\\hline
    SAOD&F-RCNN&$55.0$&$44.9$&$31.1$&$16.7$&$23.2$&$19.8$&$12.8$&$7.2$\\
    AV&ATSS&$56.9$&$47.1$&$34.1$&$18.9$&$25.1$&$21.7$&$14.8$&$8.6$\\ \hline
    \end{tabular}
\end{table}
%
%
%
Here we demonstrate the details of the selected object detectors and ensure that their performance is inline with their expected results.
We build our SAOD framework upon the mmdetection framework \cite{mmdetection} since it enables us using different datasets and models also with different design choices.
As presented in \cref{sec:saod}, we use four conventional and two probabilistic object detectors.
We exploit all of these detectors for our SAOD-Gen setting by training them on the COCO training set as $\traindata$.
We train all the detectors with the exception of D-DETR. 
As for D-DETR, we directly employ the trained D-DETR model released in mmdetection framework.
This D-DETR model is trained for 50 epochs with a batch size of 32 images on 16 GPUs (2 images/GPU) and corresponds to the vanilla D-DETR (i.e., not its two-stage version and without iterative bounding box refinement).

While training the detectors, we incorporate the multi-scale training data augmentation used by D-DETR into them in order to obtain stronger baselines.
Specifically, the multi-scale training data augmentation is sampled randomly from two alternatives: (i) randomly resizing the shorter side of the image within the range of [480, 800] by limiting its longer size to 1333 and keeping the original aspect ratio; or (ii) a sequence of
\begin{itemize}
    \item randomly resizing the shorter side of the image within the range of [400, 600] by limiting its longer size to 4200 and keeping the original aspect ratio,
    \item random cropping with a size of [384, 600],
    \item randomly resizing the shorter side of the cropped image within the range of [480, 800] by limiting its longer size to 1333 and keeping the original aspect ratio.
\end{itemize}
Unless otherwise noted, we train all of the detectors (as aforementioned, with the exception of  D-DETR, which is trained for 50 epochs following its recommended settings \cite{DDETR}) for 36 epochs using 16 images in a batch on 8 GPUs. 
Following the previous works, we use the initial learning rates of $0.020$ for F-RCNN, NLL-RCNN and ES-RCNN; $0.010$ for ATSS; and $0.012$ for RS-RCNN.
We decay the learning rate by a factor of 10 after epochs 27 and 33.  
As a backbone, we use a ResNet-50 with FPN \cite{FeaturePyramidNetwork} for all the models, as is common in practice. 
At test time, we simply rescale the images to 800 $\times$ 1333 and do not use any test-time augmentation.
For the rest of the design choices, we follow the recommended settings of the detectors.

As for SAOD-AV, we train F-RCNN \cite{FasterRCNN} and ATSS \cite{ATSS} on nuImages training set by following the same design choices.
We note that these models are trained using the annotations of the three classes (\texttt{pedestrian}, \texttt{vehicle} and \texttt{bicycle}) in nuImages dataset.

We display baseline results in \cref{tab:general_od_id_corruption} on $\valdata$, $\mathcal{T}(\mathcal{D}_{\mathrm{Val}})$, $\indata$ and $\shiftdata$ data splits, which shows the performance on the COCO val set ($\valdata$ of SAOD-Gen in the table) is inline or higher with those published in the corresponding papers.
%
We would like to note that the performance on $\valdata$ is lower than that on $\indata$ due to (i) more challenging nature of Object365/BDD100K compared to COCO/nuImages and (ii) the domain shift between them.
As an example, AP drops $\sim 30$ points from $\valdata$ (nuImages) to $\indata$ (BDD45K) even before the corruptions are applied.
As expected, we also see a decrease in performance with increasing severity of corruptions.
%
\renewcommand{\thesection}{C}
\section{Further Details on Image-level Uncertainty}
\label{app:uncertainty}
This section presents further details on image-level uncertainty including the motivation behind; the definitions of the used uncertainty estimation techniques; and more analyses.
\begin{figure}[t]
        \centering
        \includegraphics[width=0.48\textwidth]{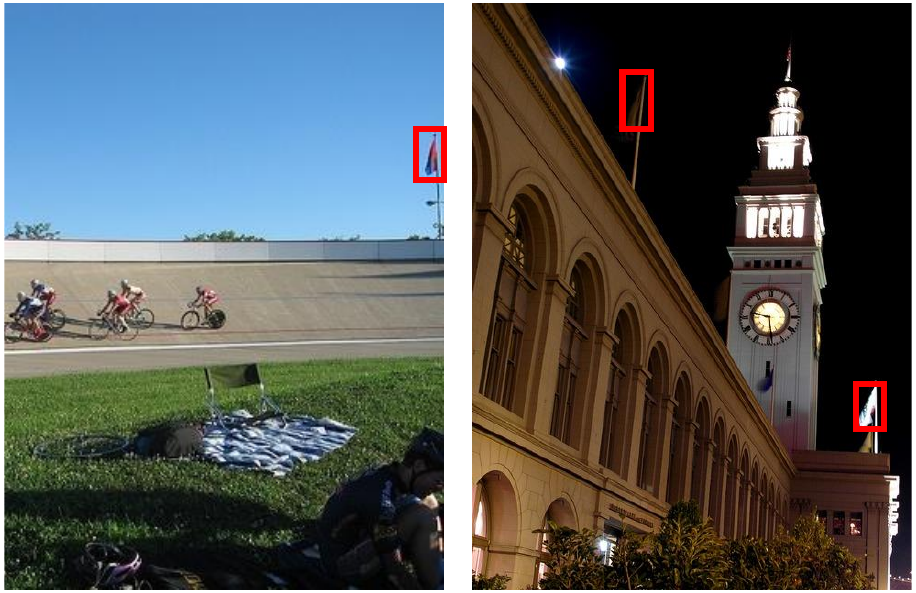}
        \caption{(Left) An example image from the ID test set and (Right) an example image from the OOD test set used by \cite{VOS}. The \texttt{flag} is not an \gls{ID} class but exists in both ID and OOD test sets as indicated by red bounding boxes. Labelling the detections corresponding to the \texttt{flag} as ID or OOD is non-trivial without labelling every pixel in the images of the training set.  Conversely, current works label all detections from an ID image as ID and those from an OOD image as OOD; and then compute AUROC in detection-level using the measured uncertainty. This way of detection-level OOD detection evaluation might not be ideal for object detection.
        }
        \label{fig:flag}
\end{figure} 

\subsection{Why is Detection-Level OOD Detection for Object Detection Nontrivial?} 
\label{subsec:whyimagelevel}
As we motivated in \cref{sec:introduction} and \cref{sec:ood}, evaluating the reliability of uncertainties using OOD detection task in detection-level is conceptually non-trivial for object detection.
This is because there is no clear definition as to which detections can be considered ID and which cannot.
To elaborate further, unknown objects may appear in two forms at test time: (i) ``known-unknowns'', which can manifest as background and unlabelled objects in the training set or (ii) ``unknown-unknowns'', completely unseen objects, which are not present in the training data.
It is not possible to split these unknown objects into the two categories without having labels for every pixel in the training set~\cite{opensetelephant}.
Current evaluation \cite{RegressionUncOD,VOS} however, does not adhere to this and instead defines ``an image'' with no ID object as OOD but assumes ``any detection'' in an OOD image is an OOD detection, and vice versa for an ID image; thereby decreasing the reliability of the evaluation.
\cref{fig:flag} presents an example from an existing ID and OOD test splits \cite{VOS} to illustrate why the reliability of the evaluation decreases.
Conversely, as we have followed, evaluating the reliability of the uncertainties for object detectors based on OOD detection at the image-level aligns with the definition of OOD images, which is again at image-level.

\subsection{Definitions} \label{app:uncertainty_def}
Here, we provide the definitions of the detection-level uncertainty estimation methods for classification and localisation as well as the aggregation techniques we used to obtain image-level uncertainty estimates.

\subsubsection{Detection-Level Uncertainties}
In the following, we present how we obtain detection-level uncertainties from classification and localisation heads.
We note that all of these uncertainties, except the uncertainty score, are computed on the raw detections represented by $\{\hat{b}^{raw}_i, \hat{p}^{raw}_i \}^{N^{raw}}$ in \cref{sec:relatedwork} and then propagated through the post-processing steps.
The uncertainty score is, instead, directly computed using the confidence of the final detections ($\hat{p}_i$).
In such a way, we obtain the uncertainty values of top-$k$ final detections, which are then aggregated for image-level uncertainty estimates.

\paragraph{Classification Uncertainties} \label{subsubsec:detectionlevelunc}
We use the following detection-level classification uncertainties:
\begin{compactitem}
\item \textit{The entropy of the predictive distribution.} 
The standard configuration of F-RCNN, NLL-RCNN and ES-RCNN employ a softmax classifier over $K$ \gls{ID} classes and background; resulting in a $K+1$-dimensional categorical distribution.
Denoting this distribution by $\hat{p}^{raw}_i$ (\cref{sec:relatedwork}), the entropy of $\hat{p}^{raw}_i$ is:
\begin{align}
\label{eq:entropy}
    \mathrm{H}(\hat{p}^{raw}_i) = - \sum_{j=1}^{K+1} \hat{p}^{raw}_{ij} \log \hat{p}^{raw}_{ij},
\end{align}
such that $\hat{p}^{raw}_{ij}$ is the probability mass in $j$th class in $\hat{p}^{raw}_{i}$.
As for the object detectors which exploit class-wise sigmoid classifiers, the situation is more complicated since the prediction $\hat{p}^{raw}_{i}$ comprises of $K$ Bernoulli random variables, instead of a single distribution unlike the softmax classifier.
Therefore, we will discuss and analyse different ways of computing entropy for the detectors using sigmoid classifiers in \cref{subsubsec:sigmoidentropy}.
\item \textit{Dempster-Shafer.} 
We use the logits as the evidence to compute DS \cite{DS}.
Accordingly, denoting the $j$th logit (i.e., for class $j$) of the $i$th detection obtained from a softmax-based detector by $s_{ij}$, we compute the uncertainty by
\begin{align}
    \mathrm{DS} = \frac{K+1}{K+1 + \sum_{j=1}^{K+1} \exp{(s_{ij})}} ,
\end{align}
and similarly, for a sigmoid-based classifier yielding $K$ logits, we simply use
\begin{align}
\label{eq:DS}
    \mathrm{DS} = \frac{K}{K + \sum_{j=1}^{K} \exp{(s_{ij})}}.
\end{align}
\item \textit{Uncertainty score.} 
While $\mathrm{H}(\hat{p}^{raw}_i)$ and $\mathrm{DS}$ are computed on the raw detections, we compute uncertainty score based on final detections using the detection confidence score as $1-\hat{p}_{i}$.
\end{compactitem}
\paragraph{Localisation Uncertainties}
We utilise the covariance matrix $\Sigma$ predicted by the probabilistic detectors (NLL-RCNN and ES-RCNN) to compute the uncertainty of a detection in the localisation head. 
As described in \cref{sec:saod}, our models predicts a diagonal covariance matrix,
\begin{align}
    \Sigma = 
\begin{bmatrix}
    \sigma_{1}^2  &  0 & 0 & 0 \\
    0 & \sigma_{2}^2 & 0 & 0\\
    0 & 0 & \sigma_{3}^2 & 0\\
    0 & 0 & 0 & \sigma_{4}^2
\end{bmatrix},
\end{align}
for each detection such that $\sigma_{i}^2$ with $0< i \leq 4$ is the predicted variance of the Gaussian for $i$th bounding box parameter.
Considering that an increase in $\Sigma$ should imply more uncertainty of the localisation head, we define the following uncertainty measures for localisation exploiting $\Sigma$,.
\begin{compactitem}
\item \textit{The determinant of the predicted covariance matrix}
\begin{align}
    |\Sigma| =  \prod_{i=1}^4 \sigma_{i}^2 
\end{align}
\item \textit{The trace of the predicted covariance matrix} 
\begin{align}
    \mathrm{tr}(\Sigma) = \sum_{i=1}^4 \sigma_{i}^2 
\end{align}
\item \textit{The entropy of the predicted multivariate Gaussian distribution \cite{pml1Book}} 
\begin{align}
    \mathrm{H}(\Sigma) = 2 +  2\ln(2 \pi) + \frac{1}{2} \ln(|\Sigma|)
\end{align}
\end{compactitem}
\subsubsection{Aggregation Strategies to Obtain Image-Level Uncertainties} \label{subsubsec:aggregation}
In this section, given detection-level uncertainties $\{u_i\}^N$ where $u_i$ is the detection-level uncertainty for the $i$th detection, we present our aggregation strategies to obtain the image-level uncertainty $\mathcal{G}(X)$.
Note that $u_i$ corresponds to a detection-level uncertainty after post-processing with top-$k$ survival (\cref{sec:relatedwork}), hence implying $N \leq k$. 
In particular, we use the following aggregation techniques that enables us to obtain reliable image-level uncertainties from different detectors:
\begin{compactitem}
\item \textit{sum:} 
\begin{align}
    \mathcal{G}(X) =  \sum \limits_{i=1}^N u_i 
\end{align}

\item \textit{mean:}
\begin{align}
    \mathcal{G}(X) = \frac{1}{N} \sum \limits_{i=1}^N u_i
\end{align}

\item \textit{mean(top-$m$):} Denoting $\phi(i)$ as the index of the $i$th smallest uncertainties,
\begin{align}
    \mathcal{G}(X) = 
\begin{cases}
    \frac{1}{m} \sum \limits_{i=1}^m u_{\phi(i)} , & \text{if } N \geq m\\
    \frac{1}{N} \sum \limits_{i=1}^N u_{\phi(i)} , & \text{if } 0 < N < m
\end{cases}
\end{align}

\item \textit{min:} Similarly, denoting $u_{\phi(1)}$ is the smallest uncertainty or the most certain one,
\begin{align}
    \mathcal{G}(X) = u_{\phi(1)}
\end{align}

\end{compactitem}
\blockcomment{
\begin{compactitem}
\item \textit{sum} 
\begin{align}
    \mathcal{G}(X) = 
\begin{cases}
    \sum \limits_{i=1}^N u_i , & \text{if } N \neq 0\\
    \kappa,              &\text{if } N=0
\end{cases}
\end{align}

\item \textit{mean}
\begin{align}
    \mathcal{G}(X) = 
\begin{cases}
    \frac{1}{N} \sum \limits_{i=1}^N u_i , & \text{if } N \neq 0\\
    \kappa,              &\text{if } N=0
\end{cases}
\end{align}
\item \textit{mean(top-$m$)}
\begin{align}
    \mathcal{G}(X) = 
\begin{cases}
    \frac{1}{m} \sum \limits_{i=1}^m u_{\phi(i)} , & \text{if } N \geq m\\
    \frac{1}{N} \sum \limits_{i=1}^N u_{\phi(i)} , & \text{if } 0 < N < m\\
    \kappa,              &\text{if } N=0
\end{cases}
\end{align}

\item \textit{min}
\begin{align}
    \mathcal{G}(X) = 
\begin{cases}
    u_{\phi(1)} , & \text{if } N \geq 0\\
    \kappa,              &\text{if } N=0
\end{cases}
\end{align}

\end{compactitem}
}

Finally, we consider the extreme case in which all of the detections are eliminated in the background removal stage, the first step of the post-processing.
It is also worth mentioning that this case can be avoided by reducing the score threshold of the detectors, which is typically $0.05$.
However, using off-the-shelf detectors by keeping their hyper-parameters as they are, we observe rare cases that a detector may not yield any detection for an image.
To give an intuition how rare these cases are, we haven't observed any image with no detection for D-DETR and RS-RCNN and there very few images for F-RCNN.
However, for the sake of completeness, we assign a large uncertainty value (typically $10^{12}$) that ensures that the image is classified as OOD in such cases.
%
%
%
%

%
%
\subsection{More Analyses on Image-Level Uncertainty} \label{app:uncertainty_analyses}
This section includes more analyses on obtaining image-level uncertainties.
We use our Gen-OD setting and report AUROC following \cref{sec:ood} unless explicitly otherwise noted.
\subsubsection{Computing Detection-level Uncertainty for Sigmoid-based Classifiers}
\label{subsubsec:sigmoidentropy}

\begin{table}
    \centering
    \setlength{\tabcolsep}{0.5em}
    \caption{AUROC values for different variations of computing entropy as the uncertainty for sigmoid-based detectors. Applying softmax to the logits to obtain $K$-dimensional categorical distribution performs the best for all detectors.}
    \label{tab:entropy}
    \begin{tabular}{|c||c|c|c|} \hline
     Detector&average&max class& categorical\\ \hline
    RS-RCNN &$73.3$&$91.2$&$\mathbf{93.7}$\\
    ATSS &$79.9$&$27.5$&$\mathbf{94.3}$\\
    D-DETR &$63.4$&$27.9$& $\mathbf{93.9}$ \\\hline
    \end{tabular}
\end{table}
Unlike the detectors using $K+1$-class softmax classifiers, the detectors employing sigmoid-based classifiers, such as RS R-CNN, ATSS and D-DETR, yield $K$ different Bernoulli random variables each of which corresponds to one of the $K$ different classes in $\traindata$.
%
%
In this case, one can think of different ways to compute the detection-level uncertainty.
Here, we analyse the effect of three different methods to obtain the uncertainty for a detection as (i) the average over the entropies of $K$ Bernoulli random variables; (ii) the entropy of the maximum-scoring class; and (iii) obtaining a categorical distribution over the classes through softmax first, and then, computing the entropy of this categorical distribution.
More precisely, for the $j$th class and the $i$th detection, denoting the predicted logit and corresponding probability (obtained through sigmoid)  by $\hat{s}_{ij}$ and $\hat{p}^{raw}_{ij}$ respectively, we define the following uncertainties:
\begin{compactitem}
\item \textit{The average of the entropies of the $K$ Bernoulli random variables:}
\begin{align}
     \frac{1}{K} \sum_{j=1}^{K} \left( \hat{p}^{raw}_{ij} \log \hat{p}^{raw}_{ij} + (1-\hat{p}^{raw}_{ij}) \log (1 - \hat{p}^{raw}_{ij}) \right);
\end{align}
\item \textit{The entropy of the maximum-scoring class:} 
\begin{align}
      \hat{p}^{raw}_{ik} \log \hat{p}^{raw}_{ik} + (1-\hat{p}^{raw}_{ik}) \log (1 - \hat{p}^{raw}_{ik})
\end{align}
with $k$ being the maximum scoring class for detection $i$; and
\item \textit{The entropy of the categorical distribution:} It is simply obtained by first applying softmax over $K$-dimensional logits, yielding a categorical distribution, and then, computing the entropy following Eq. \eqref{eq:entropy} with $K$ classes. 
\end{compactitem}

\cref{tab:entropy} indicates that the entropy of the categorical distribution performs consistently the best for all three sigmoid-based detectors.
\begin{table}
    \centering
    \setlength{\tabcolsep}{0.1em}
    \caption{Combining classification and localisation uncertainties. Please refer to the text for more details.}
    \label{tab:combine}
    \begin{tabular}{|c|c|c|c|c||c|} \hline
    Detector&$\mathrm{H}(\hat{p}^{raw}_i)$&$\mathrm{H}(\Sigma)$&Balanced&Norm.&AUROC\\ \hline
    \multirow{5}{*}{NLL-RCNN}&\cmark& & & & $92.4$ \\
    & &\cmark& & & $87.7$ \\ 
    &\cmark&\cmark& && $89.9$ \\
    &\cmark&\cmark&\cmark&& $92.2$ \\
    &\cmark &\cmark&\cmark&\cmark& $\mathbf{93.2}$ \\
     \hline \hline
    \multirow{5}{*}{ES-RCNN}&\cmark&&& & $92.8$ \\
    & &\cmark & & &$86.4$ \\ 
    &\cmark &\cmark& && $89.3$\\
    &\cmark &\cmark &\cmark & &$91.8$ \\
    &\cmark &\cmark &\cmark& \cmark& $\mathbf{93.2}$ \\ \hline
    \end{tabular}
\end{table}
\begin{table*}[t]
    \small
    \setlength{\tabcolsep}{0.4em}
    \centering
    \caption{AUROC scores of different aggregations to obtain image-level uncertainty. $|\Sigma|$ can be computed for probabilistic detectors, hence N/A for others. mean(top-m) refers to average of the uncertainties of the most $m$ certain detections (the detections with the least uncertainty based on $1-\hat{p}_{i}$ or $|\Sigma|$). Using few most certain detections perform better for both detection-level uncertainty estimation methods. Underlined \& Bold: best of a detector, bold: second best. }
    \label{tab:aggregate_}
    \begin{tabular}{|c|c||c|c||c|c||c|c||c|c||c|c||c|c|} \hline
         Dataset &\multirow{2}{*}{Detector} & \multicolumn{2}{|c|}{sum}&\multicolumn{2}{|c|}{mean}&\multicolumn{2}{|c|}{mean(top-5)}&\multicolumn{2}{|c|}{mean(top-3)}&\multicolumn{2}{|c|}{mean(top-2)}&\multicolumn{2}{|c|}{min}    \\ \cline{3-14}
         ($\mathcal{D}_{\mathrm{ID}}$ vs. $\mathcal{D}_{\mathrm{OOD}}$)&&$1-\hat{p}_{i}$& $|\Sigma|$&$1-\hat{p}_{i}$& $|\Sigma|$&$1-\hat{p}_{i}$& $|\Sigma|$&$1-\hat{p}_{i}$& $|\Sigma|$&$1-\hat{p}_{i}$& $|\Sigma|$&$1-\hat{p}_{i}$& $|\Sigma|$\\ \hline
    \multirow{6}{*}{SAOD-Gen}&F R-CNN \cite{FasterRCNN}&$20.9$&N/A&$84.1$&N/A&$93.4$&N/A&$\mathbf{94.1}$& N/A&\underline{$\mathbf{94.4}$}&N/A&$93.8$&N/A\\
    &RS R-CNN \cite{RSLoss}&$85.8$&N/A&$85.8$&N/A&$94.3$&N/A&\underline{$\mathbf{94.8}$}&N/A&\underline{$\mathbf{94.8}$}&N/A&$93.5$&N/A\\
    &ATSS \cite{ATSS}&$66.2$&N/A&$86.3$&N/A&$93.8$&N/A&\underline{$\mathbf{94.2}$}&N/A&$\mathbf{94.0}$&N/A&$92.6$&N/A\\
    &D-DETR \cite{DDETR}&$85.2$&N/A&$85.2$&N/A&$94.4$&N/A&$\mathbf{94.7}$& N/A&$94.6$&N/A&$93.3$&N/A\\ \cline{2-14}
    &NLL R-CNN \cite{KLLoss}&$22.6$&$41.6$&$83.8$&$74.9$&$93.4$&$87.4$&$\mathbf{94.1}$&$87.6$&\underline{$\mathbf{94.4}$}&$87.5$&$93.7$&$87.0$\\
    &ES R-CNN \cite{RegressionUncOD}&$22.1$&$24.5$&$84.6$&$32.9$ &$93.4$&$83.8$&$\mathbf{94.1}$&$85.0$&\underline{$\mathbf{94.4}$}&$85.7$&$93.8$&$86.3$\\ \hline
    \multirow{2}{*}{SAOD-AV}&F R-CNN \cite{FasterRCNN}&$27.1$&N/A&$84.1$&N/A&$96.4$&N/A&$\mathbf{97.3}$& N/A&\underline{$\mathbf{97.4}$}&N/A&$96.0$&N/A\\
    &ATSS \cite{ATSS}&$18.8$&N/A&$92.2$&N/A&\underline{$\mathbf{97.7}$}&N/A&$\mathbf{97.6}$&N/A&$97.3$&N/A&$95.7$&N/A\\\hline
    \end{tabular}
\end{table*}

\subsubsection{Combining Classification and Localisation Uncertainties}
Considering that the probabilistic object detectors yield an uncertainty both for classification and localisation for each detection, here we investigate whether there is a benefit in combining such uncertainties for a detection. 
Basically, we combine the entropy of the predictive classification distribution ($\mathrm{H}(\hat{p}^{raw}_i)$) and the entropy of the predictive Gaussian distribution for localisation ($\mathrm{H}(\Sigma)$).
Assuming these two distributions independent, the entropy of the joint distribution can be obtained by the summation over the entropies, i.e., $\mathrm{H}(\hat{p}^{raw}_i) + \mathrm{H}(\Sigma)$. 
However, this way of combining tends to overestimate the contribution of the localisation as the localisation output involves four random variables but the classification is univariate. 
Consequently, we first increase the contribution of the classification by multiplying its entropy by 4, which results in a positive effect (``balanced'' in Table \ref{tab:combine}).
We also find it useful to normalize the uncertainties between $0$ and $1$ using their minimum and maximum scores obtained on the validation set.
This normalisation shows for the both cases that the resulting performance is better compared to using only classification and localisation (``Norm.'' in Table \ref{tab:combine}). 
%
%

We also would like to note that the resulting AUROC does not outperform using only the uncertainty score ($1-\hat{p}_{i}$), which yields $94.1$ AUROC score with mean(top-3) for both of the detectors ES-RCNN and NLL-RCNN as demonstrated in \cref{tab:aggregate}.
Hence, instead of combining classification and localisation uncertainties, we still suggest using $1-\hat{p}_{i}$ to obtain the detection-level uncertainties.

\subsubsection{The Effect of Aggregation Techniques on a Localisation Uncertainty Estimate} 
In \cref{tab:aggregate}, we investigated the aggregation methods using the uncertainty score, which is a classification-based uncertainty estimate.
Here in \cref{tab:aggregate_}, we present an extended version of \cref{tab:aggregate} including the effect of the same aggregation methods on $|\Sigma|$ as one example of our localisation uncertainty estimation methods.
\cref{tab:aggregate_} also validates our conclusion on $|\Sigma|$ to use the most certain detections in obtaining image-level uncertainties. 

\begin{figure*}[t]
        \captionsetup[subfigure]{}
        \centering
        \begin{subfigure}[b]{0.48\textwidth}
        \includegraphics[width=\textwidth]{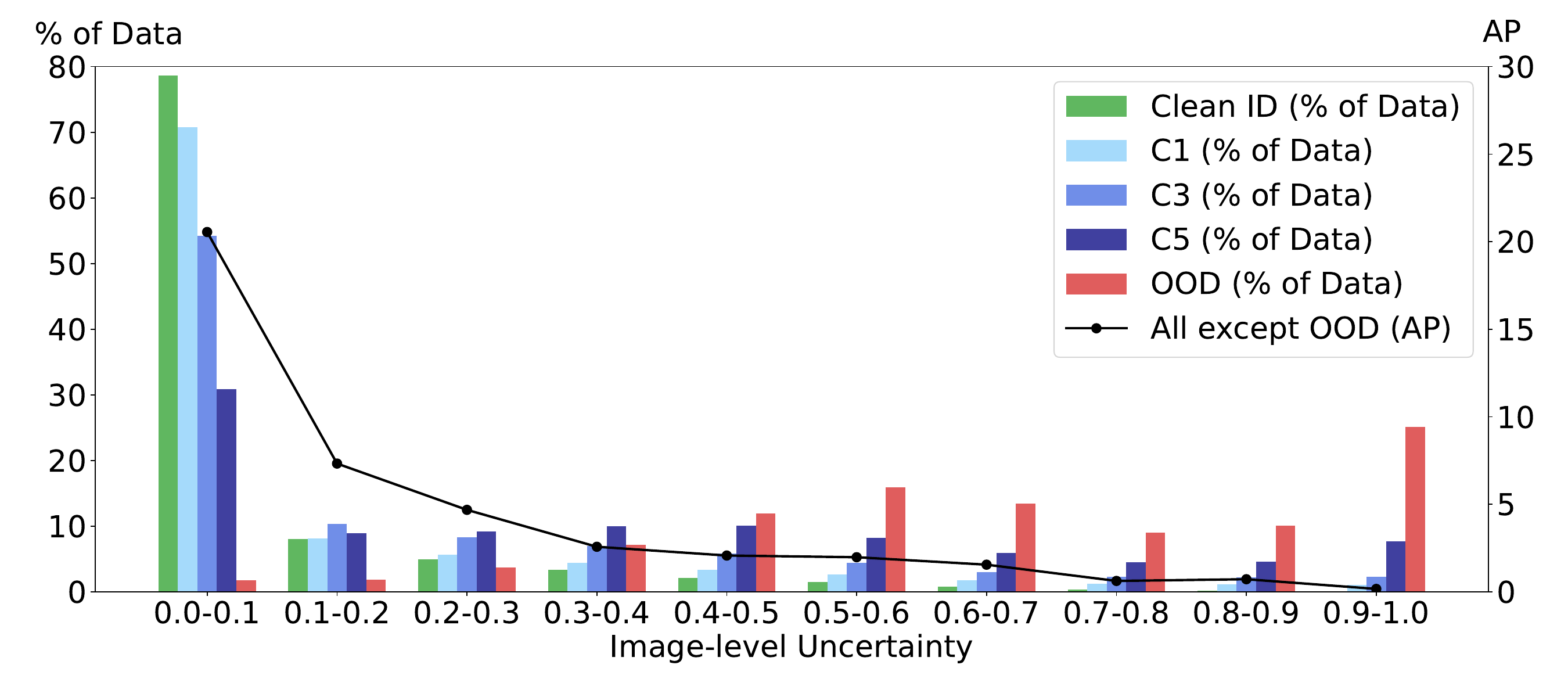}
        \caption{F-RCNN}
        \end{subfigure}
        \begin{subfigure}[b]{0.48\textwidth}
        \includegraphics[width=\textwidth]{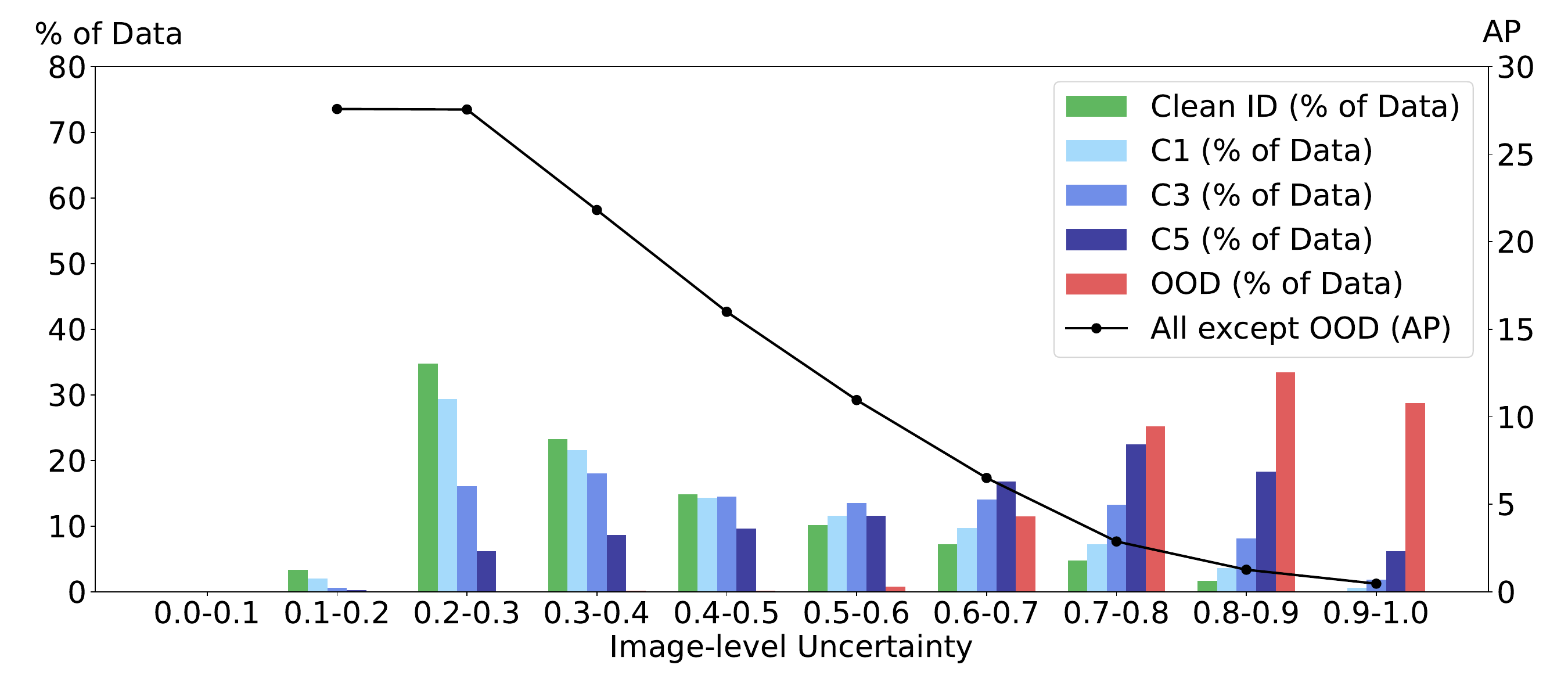}
        \caption{ATSS}
        \end{subfigure}
        \caption{The distribution of the image-level uncertainties obtained from different detectors on clean ID, corrupted ID with severities 1, 3, 5 and OOD data on SAOD-AV dataset. }
        \label{fig:unc_reliable_all}
\end{figure*}

\subsubsection{On the Reliability of Image-level Uncertainties} \label{subsubsec:analyses_reliability}

%
Here, similar to our analysis on SAOD-Gen, we present the distribution of the image-level uncertainties for different subsets of $\testdata$ on SAOD-AV.
Fig. \ref{fig:unc_reliable_all} confirms that the following observations obtained on SAOD-Gen are also valid for SAOD-AV: (i) the distribution gets more right-skewed as the subset moves away from the $\indata$ and (ii) \gls{AP} (black line) perfectly decreases as the uncertainties increase (refer to \cref{sec:ood} for details).
These figures confirm on a different dataset that the image-level uncertainties are reliable and effective.
\begin{table}
    \small
    \centering
    \setlength{\tabcolsep}{0.5em}
    \caption{Effectiveness of our pseudo-OOD set approach compared to using TPR@0.95.}
    \label{tab:threshold_ood_app}
    \begin{tabular}{|c|c|c||c||c|c|} \hline
    Task&Detector&Method&BA&TPR&TNR\\ \hline
    \multirow{8}{*}{Gen-OD}&\multirow{2}{*}{F-RCNN}&TPR@0.95&$83.2$&$98.5$&$72.0$\\ 
    &&pseudo-OOD&$\mathbf{87.7}$&$94.7$&$81.6$\\ \cline{2-6}
    &\multirow{2}{*}{RS-RCNN}&TPR@0.95&$84.0$&$98.3$&$73.4$\\
    &&pseudo-OOD&$\mathbf{88.9}$&$92.8$&$85.3$\\ \cline{2-6}
    &\multirow{2}{*}{ATSS}&TPR@0.95&$84.7$&$96.9$&$75.2$\\
    &&pseudo-OOD&$\mathbf{87.8}$&$93.1$&$83.0$\\ \cline{2-6}
    &\multirow{2}{*}{D-DETR}&TPR@0.95&$85.8$&$97.2$&$76.8$\\
    &&pseudo-OOD&$\mathbf{88.9}$&$90.0$&$87.8$\\\hline
    \multirow{4}{*}{SAOD-AV}&\multirow{2}{*}{F-RCNN}&TPR@0.95&$80.9$&$97.7$&$69.1$\\ 
    &&pseudo-OOD&$\mathbf{91.0}$&$94.1$&$88.2$\\ \cline{2-6}
    &\multirow{2}{*}{ATSS}&TPR@0.95&$83.5$&$96.7$&$73.5$\\
    &&pseudo-OOD&$\mathbf{85.8}$&$95.9$&$77.6$\\ \hline
    \end{tabular}
\end{table}

\subsubsection{The Effectiveness of Using Pseudo OOD val set for Image-level Uncertainty Thresholding} 
\label{subsubsec:imagethreshold}
In order to compute the image-level uncertainty threshold $\bar{u}$ and decide whether or not to accept an image, we presented a way to construct pseudo-OOD val set in \cref{sec:ood} as $\valdata$ only includes ID images.
Here, we discuss the effectiveness of this pseudo-set approach.
To do so, we also prefer to have a baseline to compare our method against and demonstrate its effectiveness.
However, to the best of our knowledge, there is no such a method that obtains a threshold relying only on the ID data for OOD detection task. 
As a result, inspired from the performance measure TPR@0.95 \cite{VOS}, we simply set the threshold $\bar{u}$ to the value that corresponds to TPR@0.95, and use it as a baseline.
Note that this approach only relies on the ID val set= and hence there is no need for OOD val set, which is similar to our pseudo-OOD approach.
\cref{tab:threshold_ood_app} compares our pseudo-OOD approach with TPR@0.95 baseline; suggesting, on average, more than $4.5$ \gls{BA} gain over the baseline method; thereby confirming the effectiveness of our approach.

\renewcommand{\thesection}{D}

\section{Further Details on Calibration of Object Detectors}
\label{app:calibration}
This section provides further details and analyses on calibration of object detectors.
\begin{figure*}[t]
        \captionsetup[subfigure]{}
        \centering
        \begin{subfigure}[b]{0.24\textwidth}
            \includegraphics[width=\textwidth]{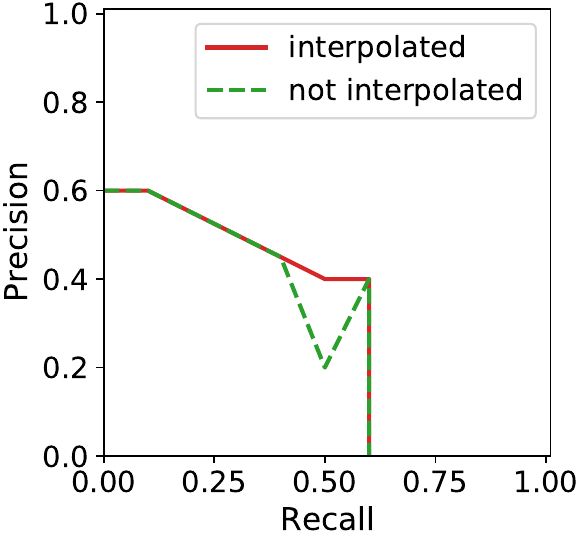}
            \caption{Interpolating PR Curve}
        \end{subfigure}
        \begin{subfigure}[b]{0.24\textwidth}
            \includegraphics[width=\textwidth]{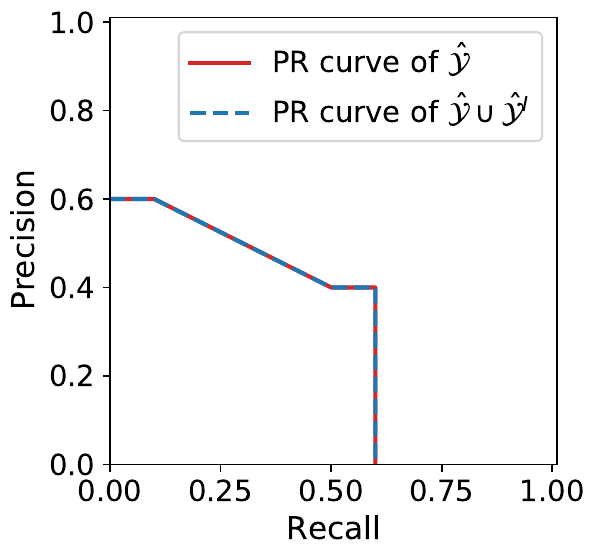}
            \caption{Case 1 of the proof}
        \end{subfigure}
        \begin{subfigure}[b]{0.24\textwidth}
            \includegraphics[width=\textwidth]{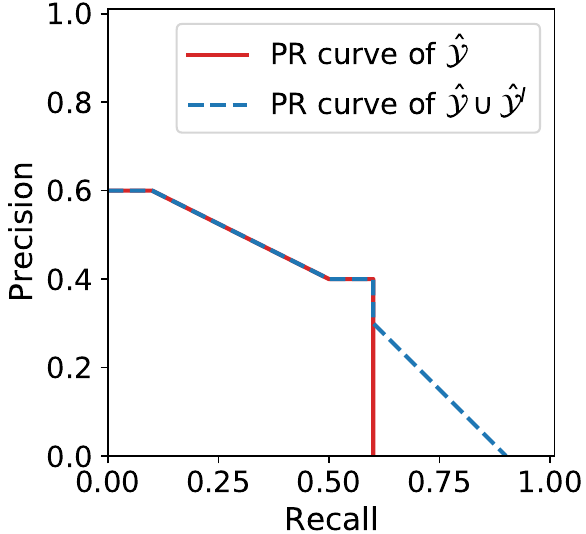}
            \caption{Case 2 of the proof}
        \end{subfigure}
        \begin{subfigure}[b]{0.24\textwidth}
            \includegraphics[width=\textwidth]{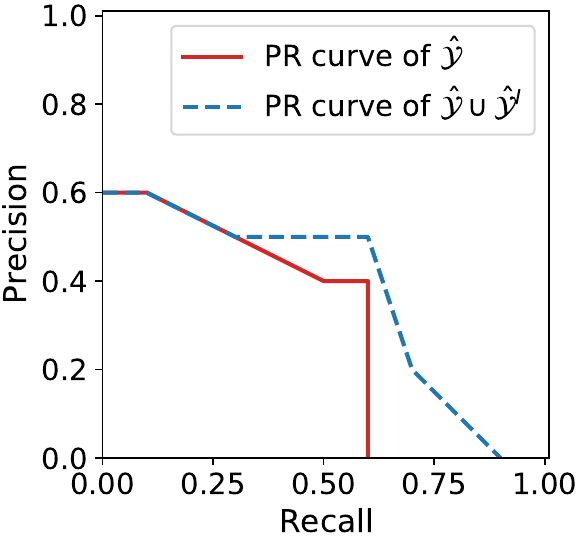}
            \caption{Case 3 of the proof}
        \end{subfigure}
        \caption{Illustrations of \textbf{(a)} non-interpolated and interpolated PR curves. Typically, the area under the interpolated PR curve is used as the AP value in object detection; \textbf{(b), (c), (d)} corresponding to each of the three different cases we consider in the proof of Theorem \ref{theorem:AP}. Following Theorem \ref{theorem:AP}, in all three cases, the area under the red curve is smaller or equal to that of the blue curve. 
        }
        \label{fig:ap_proof}
\end{figure*}
\subsection{How does AP benefit from low-scoring detections?}
Here we show that enlarging the detection set with low-scoring detections provably does not decrease AP.
Thereby confirming the practical limitations previously discussed by Oksuz et al. \cite{LRPPAMI}.
As a result, instead of top-k predictions and AP, we require a thresholded detection set in SAOD task and employ the LRP Error as a measure of accuracy to enforce this type of output.

Before proceeding, below we provide a formal definition of AP as a basis to our proof.

\paragraph{Definition of AP.} AP is defined as the area under the Precision-Recall curve \cite{PASCAL,LRP,LRPPAMI}. 
Here we formalize how to obtain this curve and the resulting AP in object detection given detections and the ground truths.
Considering the common practice, we will first focus on the AP of a single class and then discuss further after our proof.
More precisely, computing AP for class $c$ from an IoU threshold of $\tau$, two sets are required:
\begin{compactitem}
\item \textit{A set of detections obtained on the test set:} This set is represented by tuples $\hat{\mathcal{Y}}=\{\hat{b}_i, \hat{p}_i, X_i\}^{N_c}$, where $\hat{b}_i$ and $ \hat{p}_i$ are the bounding box and confidence score of the $i$th detection respectively.
$X_i$ is the image id that the $i$th detection resides and $N_c$ is the number of all detections across the dataset from class $c$. 
We assume that the detections obtained from a single image is less than $k$ where $k$ represents the upper-bound within the context of the top-$k$ survival (\cref{sec:relatedwork}), that is, there can be up to $k$ detections from each image.
\item \textit{A set of ground truths of the test set:} This set is represented by tuples $\mathcal{Y}=\{b_i, X_i\}^{M_c}$, where $b_i$ is the bounding box of the ground truth and $X_i$ is similarly the image id.
$M_c$ is the number of total ground truth objects from class $c$ across the dataset.
\end{compactitem}
Then, the detections are matched with the ground truths to identify TP and FP detections using a matching algorithm.
For the sake of completeness, we provide a matching algorithm that is used by the commonly-used COCO benchmark \cite{COCO}:
\begin{enumerate}
    \item All detections in $\hat{\mathcal{Y}}$ are first sorted with respect to the confidence score in descending order
    \item Then, going over this sorted list of detections, the $j$th detection is identified as a TP if there exists a ground truth that satisfies the following two conditions: 
    \begin{compactitem}
        \item The ground truth is not previously assigned to any other detections with a larger confidence score than that of $j$,
        \item The IoU between the ground truth and $j$th detection is more than $\tau$, the TP validation threshold.
    \end{compactitem}
    Note that the second condition also implies that the $j$th detection and the ground truth that it matches with should reside in the same image.
    If there is a single ground truth satisfying these two conditions, then $j$ is matched with that ground truth; else if there are more than one ground truths ensuring these conditions, then the $j$th detection is matched with the ground truth that $j$ has the largest IoU with.
    \item Upon completing this sorted list, the detections that are not matched with any ground truths are identified as FPs.
\end{enumerate}

This matching procedure enables us to determine which detections are TP or FP. 
Now, let  $L=[L_1, ..., L_{N_c}]$ be a binary vector that represents whether $j$th detection is a TP or FP and assume that $L$ is also sorted with respect to the confidence scores of the detections.
Specifically,  $L_i=1$ if the $i$th detection is a TP, else the $i$th detection is FP and $L_i=0$.
Consequently, we need precision and recall pairs in order to obtain the Precision-Recall curve, area under which corresponds to the AP.
Noting that the precision is the ratio between the number of TPs and number of all detections; and recall is the ratio between the number of TPs and number of ground truths, we can obtain these pairs by leveraging $L$. 
Denoting the precision and recall vectors by $Pr=[Pr_1, ..., Pr_{N_c}]$ and $Re=[Re_1, ..., Re_{N_c}]$ respectively, the $i$th element of these vectors can be obtained by:
\begin{align}
    \label{eq:pr}
    Pr_i = \frac{\sum_{k=1}^i L_k}{i}  \text{, and }   Re_i = \frac{\sum_{k=1}^i L_k}{M_c}.
\end{align}
Since these obtained precision values $Pr_i$ may not be monotonically decreasing function of recall, there can be wiggles in the Precision-Recall curve.
Therefore, it is common in object detection \cite{PASCAL,COCO,LVIS} to interpolate the precisions $Pr$ to make it monotonically decreasing with respect to the recall $Re$.
Denoting the interpolated precision vector by $\bar{Pr}=[\bar{Pr}_1, ..., \bar{Pr}_{N_c}]$, its $i$th element $\bar{Pr}_i$ is obtained as follows:
\begin{align}
    \label{eq:interpolation}
    \bar{Pr}_i = \max \limits_{i : Re_i \geq Re_k} (Pr_i).
\end{align}
Finally, Eq. \eqref{eq:interpolation} also allows us to interpolate the PR curve to the precision and recall axes.
Namely, we include the pairs that (i) $\bar{Pr}_1$ with recall $0$; and (ii) precision $0$ with recall $\bar{Re}_{N_c}$.
This allows us to obtain the final Precision-Recall curve using these two additional points as well as the vectors $\bar{Pr}_i$ and $Re_i$.
Then, the area under this curve corresponds to the Average Precision of the detection set $\hat{\mathcal{Y}}$ for the IoU validation threshold of $\tau$, which we denote as $\mathrm{AP}_\tau(\hat{\mathcal{Y}})$.
As an example, \cref{fig:ap_proof}(a) illustrates a PR curve before and after interpolation.
Based on this definition, we now prove that low-scoring detections do not harm AP.

\begin{theorem}
\label{theorem:AP}
Given two sets of detections $\hat{\mathcal{Y}}=\{\hat{b}_i, \hat{p}_i, X_i\}^{N_c}_{i=1}$, $\hat{\mathcal{Y}}^{'}=\{\hat{b}_j, \hat{p}_j, X_j\}^{N_c^{'}}_{j=1}$ and denoting $p_{min} = \min \limits_{\{\hat{b}_i, \hat{p}_i, X_i\} \in \hat{\mathcal{Y}}} \hat{p}_i$, $p_{max}^{'} = \max \limits_{\{\hat{b}_j, \hat{p}_j, X_j\} \in \hat{\mathcal{Y}}^{'}} \hat{p}_j$, if $p_{max}^{'} < p_{min}$, then $\mathrm{AP}_\tau(\hat{\mathcal{Y}}) \leq \mathrm{AP}_\tau(\hat{\mathcal{Y}} \cup \hat{\mathcal{Y}}^{'})$.
\end{theorem}{}
\begin{proof}
We denote the recall and precision values to compute $\mathrm{AP}_\tau(\hat{\mathcal{Y}})$ by $Pr=[Pr_1, ..., Pr_{N_c}]$ and $Re=[Re_1, ..., Re_{N_c}]$, and similarly the interpolated precision is $\bar{Pr}=[\bar{Pr}_1, ..., \bar{Pr}_{N_c}]$.
We aim to obtain these vectors  for $\mathrm{AP}_\tau(\hat{\mathcal{Y}} \cup \hat{\mathcal{Y}}^{'})$ to be able to compare the resulting $\mathrm{AP}_\tau(\hat{\mathcal{Y}} \cup \hat{\mathcal{Y}}^{'})$ with $\mathrm{AP}_\tau(\hat{\mathcal{Y}})$.
To do so, we introduce $Pr^{'}$, $Re^{'}$ and $\bar{Pr}^{'}$ as the precision, recall and the interpolated precision vectors of $\mathrm{AP}_\tau(\hat{\mathcal{Y}} \cup \hat{\mathcal{Y}}^{'})$ respectively.

By definition, the numbers of elements in $Pr^{'}$, $Re^{'}$ and $\bar{Pr}^{'}$ are equal to the number of detections in $\hat{\mathcal{Y}} \cup \hat{\mathcal{Y}}^{'}$, which is simply $N_c+N_c^{'}$.
More precisely, we need to determine the following three vectors to be able to obtain $\mathrm{AP}_\tau(\hat{\mathcal{Y}} \cup \hat{\mathcal{Y}}^{'})$:
\begin{align}
 Pr^{'}=\{Pr_1^{'}, ..., Pr_{N_c}^{'}, Pr_{N_c+1}^{'}, ... , Pr_{N_c+N_c^{'}}^{'}\}   \\
 Re^{'}=\{Re_1^{'}, ..., Re_{N_c}^{'}, Re_{N_c+1}^{'}, ... , Re_{N_c+N_c^{'}}^{'}\} \\
 \bar{Pr}^{'}=\{\bar{Pr}_1^{'}, ..., \bar{Pr}_{N_c}^{'}, \bar{Pr}_{N_c+1}^{'}, ... , \bar{Pr}_{N_c+N_c^{'}}^{'}\}
\end{align}
As an additional insight to those three vectors, $p_{max}^{'} < p_{min}$ implies the following:
\begin{compactitem}
    \item The first ${N_c}$ elements of $Pr^{'}$, $Re^{'}$ and $\bar{Pr}^{'}$ account for the precision, recall and interpolated precision values computed on the detection from $\hat{\mathcal{Y}}$; and 
    \item their elements between ${N_c}+1$ to the last element ($N_c + N_c^{'}$) correspond to the precision, recall and interpolated precision values computed on the detections from $\hat{\mathcal{Y}}^{'}$.
\end{compactitem}

Note that by definition, computing precision and recall on the $i$th detection only considers the detections with higher scores than that of $i$ (and ignores the ones with lower scores than that of $i$), since the list of labels denoted by $L$ in Eq. \eqref{eq:pr} is sorted with respect to the confidence scores.
As a result, the following holds for precision and recall values (but not the interpolated precision):
\begin{align}
    \label{eq:pr_1}
    Pr_i^{'} = Pr_i  \text{, and }   Re_i^{'} = Re_i \text{ for $i \leq N_c$}.
\end{align}
Then, the difference between $\mathrm{AP}_\tau(\hat{\mathcal{Y}})$ and $\mathrm{AP}_\tau(\hat{\mathcal{Y}} \cup \hat{\mathcal{Y}}^{'})$ depends on two aspects: 
\begin{enumerate}
    \item  $Pr_i^{'}$ and $Re_i^{'}$ for $N_c < i \leq N_c+N_c^{'}$
    \item the interpolated precision vector $\bar{Pr}^{'}$ of $\hat{\mathcal{Y}} \cup \hat{\mathcal{Y}}^{'}$, to be obtained using $Pr^{'}$ and $Re^{'}$ based on Eq. \eqref{eq:interpolation}
\end{enumerate}

For the rest of the proof, we enumerate all possible three cases for $\hat{\mathcal{Y}}^{'}$ and identify these aspects. 

\paragraph{Case (1): $\hat{\mathcal{Y}}^{'}$ does not include any TP.} 

This case suggests that the detections in $\hat{\mathcal{Y}}^{'}$ are all FPs, and neither the number of TPs nor the number of FNs change for $N_c < i \leq N_c+N_c^{'}$, implying:
\begin{align}
Re_i^{'}=Re_{N_c}^{'} \text{, for } N_c < i \leq N_c+N_c^{'}.
\end{align}
As for the precision, it is monotonically decreasing as $i$ increases between $N_c < i \leq N_c+N_c^{'}$ since the number of FPs increases, that is,
\begin{align}
    Pr_{i-1}^{'} > Pr_{i}^{'} \text{, for } N_c < i \leq N_c+N_c^{'}.
\end{align}

Having identified $Pr_i^{'}$ and $Re_i^{'}$ for $N_c < i \leq N_c+N_c^{'}$, now let's obtain the interpolated precision $\bar{Pr}^{'}$.
To do so, we focus on $\bar{Pr}^{'}$ in two parts: Up to and including its $N_c$th element and its remaining part.
Since $Pr_{N_c}^{'} > Pr_{i}^{'} \text{, for } N_c < i \leq N_c+N_c^{'}$, the low-scoring detections in $\hat{\mathcal{Y}}^{'}$ do not affect $\bar{Pr}_i^{'}$ for $i \leq N_c$ considering Eq. \eqref{eq:interpolation}, implying:
\begin{align}
    \bar{Pr}_i^{'} = \bar{Pr}_i \text{, for } i \leq N_c.
\end{align}
As for $N_c < i \leq N_c+N_c^{'}$, since $Re_k  = Re_i$, $\bar{Pr}_i^{'} = \bar{Pr}_{N_c}^{'}$ holds.

As a result, the detections from $\hat{\mathcal{Y}}^{'}$ will have all equal recall and interpolated precision, which is also equal to $\bar{Pr}_{N_c}$ and $Re_{N_c}$; implying that they do not introduce new points to the Precision-Recall curve used to obtain $\mathrm{AP}_\tau(\hat{\mathcal{Y}})$.
Therefore, $\mathrm{AP}_\tau(\hat{\mathcal{Y}}) = \mathrm{AP}_\tau(\hat{\mathcal{Y}} \cup \hat{\mathcal{Y}}^{'})$ in this case.

\cref{fig:ap_proof}(b) illustrates this case to provide more insight. 
In particular, when there is no TP in the low-scoring detections ($\hat{\mathcal{Y}}^{'}$), then no new points are introduced compared to the PR curve of $\hat{\mathcal{Y}}$ and the resulting AP after including low-scoring detections does not change.

\paragraph{Case (2): $\hat{\mathcal{Y}}^{'}$ includes TPs and $\max \limits_{N_c < i \leq N_c+N_c^{'}} (Pr_i^{'}) \leq \min \limits_{ i \leq N_c} (\bar{Pr}_i)$.} 

Note that $\max \limits_{N_c < i \leq N_c+N_c^{'}} (Pr_i^{'}) \leq \min \limits_{ i \leq N_c} (\bar{Pr}_i)$ implies that the interpolated precisions computed on the detection set $\hat{\mathcal{Y}}$ ($\bar{Pr}_i^{'}$ for $i \leq N_c$) will not be affected by the detections in $\hat{\mathcal{Y}}^{'}$.
As a result, Eq. \eqref{eq:pr_1} can simply be extended to the interpolated precisions:
\begin{align}
    \label{eq:pr_1_}
    \bar{Pr}_i^{'} = \bar{Pr}_i \text{ for $i \leq N_c$}.
\end{align}
Considering the area under the curve of the pairs $\bar{Pr}_i^{'}$ and $Re_i^{'} = Re_i$ for $i \leq N_c$, it is already guaranteed that $\mathrm{AP}_\tau(\hat{\mathcal{Y}}) \leq \mathrm{AP}_\tau(\hat{\mathcal{Y}} \cup \hat{\mathcal{Y}}^{'})$ completing the proof for this case.

To provide more insight, we also briefly explore the effect of remaining detections, that are the detections in $N_c < i \leq N_c+N_c^{'}$ and include TPs.
Assume that $j$th detection is the TP with the highest confidence score within the detections for $N_c < i \leq N_c+N_c^{'}$.
Then, for the $j$th detection $0< \bar{Pr}_j^{'} < \bar{Pr}_{N_c}^{'}$ as $\max \limits_{N_c < i \leq N_c+N_c^{'}} (Pr_i^{'}) \leq \min \limits_{ i \leq N_c} (\bar{Pr}_i)$ by definition.
Moreover, since the number of TPs increases and the number of ground truths is a fixed number, $Re_j^{'} > Re_{N_c}^{'}$.
This implies that, the PR curve now has $\bar{Pr}_j^{'} > 0$ precision for some $Re_j^{'}$.
Note that the precision was implicitly $0$  for $Re_j^{'}$ for the detection set $\hat{\mathcal{Y}}$ since this new ground truth could not be retrieved regardless of the number of predictions.
Accordingly, the additional area under the PR curve of $\hat{\mathcal{Y}} \cup \hat{\mathcal{Y}}^{'}$  compared to that of $\hat{\mathcal{Y}}$ increases and it is guaranteed that $\mathrm{AP}_\tau(\hat{\mathcal{Y}}) < \mathrm{AP}_\tau(\hat{\mathcal{Y}} \cup \hat{\mathcal{Y}}^{'})$ in this case.
As depicted in \cref{fig:ap_proof}(c), the area-under-the PR curve of $\hat{\mathcal{Y}}$ is extended towards higher recall (compare blue curve with the red one) resulting in a larger $\mathrm{AP}_\tau(\hat{\mathcal{Y}} \cup \hat{\mathcal{Y}}^{'})$ compared to $\mathrm{AP}_\tau(\hat{\mathcal{Y}})$.

\paragraph{Case (3): $\hat{\mathcal{Y}}^{'}$ includes TPs and $\max \limits_{N_c < i \leq N_c+N_c^{'}} (Pr_i^{'}) > \min \limits_{ i \leq N_c} (\bar{Pr}_i)$.} Unlike case (ii), this case implies that upon merging $\hat{\mathcal{Y}}$ and $\hat{\mathcal{Y}}^{'}$, some of the $\bar{Pr}_i$ of $\hat{\mathcal{Y}}$ with $Pr_j^{'} > \bar{Pr}_i$ will be replaced by a larger value  due to Eq. \eqref{eq:interpolation}, i.e., $\bar{Pr}_i^{'} > \bar{Pr}_i$ for some $i$ while the rest will be equal similar to Case (ii).
This simply implies that $\mathrm{AP}_\tau(\hat{\mathcal{Y}}) < \mathrm{AP}_\tau(\hat{\mathcal{Y}} \cup \hat{\mathcal{Y}}^{'})$.

\cref{fig:ap_proof}(d) includes an illustration for this case demonstrating that the PR curve of $\hat{\mathcal{Y}}$ is extended in both of the axes: (i) owing to the interpolation thanks to a TP in $\hat{\mathcal{Y}}^{'}$ with higher precision in $\hat{\mathcal{Y}} \cup \hat{\mathcal{Y}}^{'}$, it is extended in precision axis; and (ii) thanks to a new TP in $\hat{\mathcal{Y}}^{'}$, it is extended in recall axis.
Note that in our proof for this case, we only discussed the extension in precision since each of the extensions is sufficient to show $\mathrm{AP}_\tau(\hat{\mathcal{Y}}) < \mathrm{AP}_\tau(\hat{\mathcal{Y}} \cup \hat{\mathcal{Y}}^{'})$.
\end{proof}

\paragraph{Discussion} 
Theorem \ref{theorem:AP} can also be extended to COCO-style AP.
To be more specific and revisit the definition of COCO-style AP, first the class-wise COCO-style APs are obtained by averaging over the APs computed over $\tau \in \{0.50,0.55,...,0.95\}$ for a single class. 
Then, the detector COCO-style AP is again the average of the class-wise APs.
Considering that the arithmetic mean is a monotonically increasing function, Theorem \ref{theorem:AP} also applies to the class-wise COCO-style AP and the detector COCO-style AP.
More precisely, in the case that either Case (1) applies for some (or all) of the classes and the detections for the remaining classes stay the same, then following Case (1), COCO-style AP does not change.
That is also the reason why we do not observe a change in COCO-Style AP in \cref{fig:calibration}(a) once we add dummy detections that are basically FPs with lower scores.
If at least for a single class Case (2) or (3) apply, then COCO-style AP increases considering the monotonically increasing nature of the arithmetic average.
Following from this, we observe some decrease in COCO-style AP when we remove the detections in \cref{fig:calibration}(b) when we threshold and remove some TPs.
As a result, we conclude that AP, including COCO-Style AP, encourages the detections with lower scores.

\begin{figure*}[t]
        \captionsetup[subfigure]{}
        \centering
        \begin{subfigure}[b]{0.48\textwidth}
            \includegraphics[width=\textwidth]{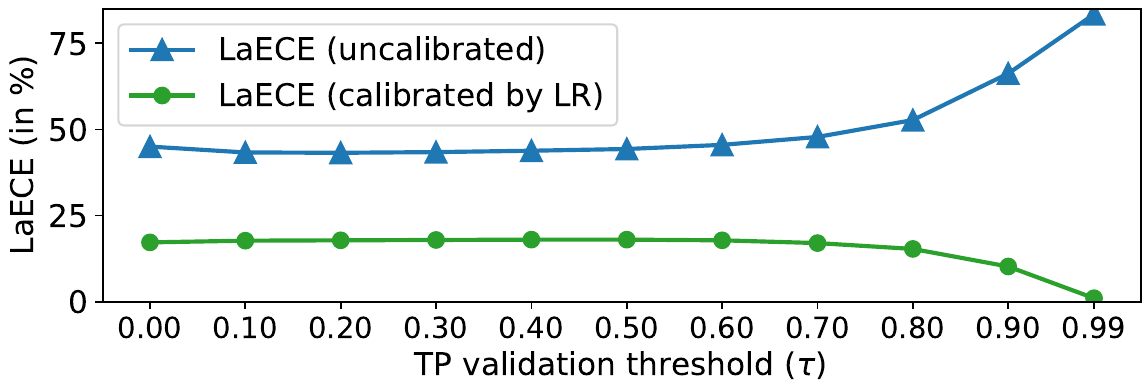}
            \caption{Sensitivity of LaECE to $\tau$}
        \end{subfigure}
        \begin{subfigure}[b]{0.48\textwidth}
            \includegraphics[width=\textwidth]{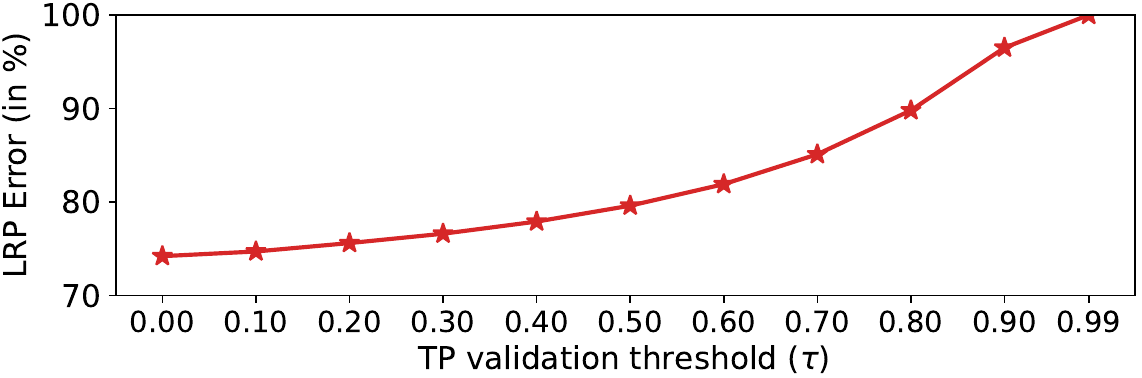}
            \caption{Sensitivity of LRP Error to $\tau$}
        \end{subfigure}
        \caption{Sensitivity analysis of \gls{LaECE} and LRP Error. We use the detections of F-RCNN on our Obj45K split. \textbf{(a)} For both calibrated and uncalibrated case, we observe that the \gls{LaECE} is not sensitive for $\tau \in [0.0, 0.5]$. When $\tau$ gets larger, the misalignment between detection scores and performance increases for the uncalibrated case, while the calibration becomes an easier problem since most of the detections are now \gls{FP}. In the extreme case that $\tau = 1$ (a perfect localisation is required for a TP), there is no TP and it is sufficient to assign a confidence score of $0.00$ to all of the detections to obtain $0$ LaECE. \textbf{(b)} Sensitivity analysis of LRP Error. As also previously analysed \cite{LRPPAMI}, when $\tau$ increases, number of \gls{FP}s increases and \gls{LRP} increases. In the extreme case when $\tau \approx 1$, \gls{LRP} approximates 1.
        }
        \label{fig:sensitivity}
\end{figure*}
\subsection{Sensitivity of \gls{LaECE} to TP validation threshold}

Here we analyse the sensitivity of \gls{LaECE} to the \gls{TP} validation threshold $\tau$.
Please note that we normally obtain class-wise LRP-optimal thresholds $\bar{v}$ considering a specific $\tau$ on $\valdata$, then use the resulting detections while measuring the LRP Error and \gls{LaECE} on the test set using the same IoU validation threshold $\tau$.
Namely, we use $\tau$ for two purposes: (i) to obtain the thresholds $\bar{v}$; and (ii) to evaluate the resulting detections in terms of $\gls{LaECE}$ and LRP Error. 
As we aim to understand how $\gls{LaECE}$, as a performance measure, behaves under different specifications of the TP validation threshold $\tau$ here, we decouple these two purposes of $\tau$ by fixing the detection confidence threshold $\bar{v}$ to the value obtained from a TP validation threshold of $0.10$.
This enables us to fix the detection set as the input of \gls{LaECE} and only focus on how the performance measures behave when only $\tau$ changes.

Specifically, we use F-RCNN detector, validate $\bar{v}$ on COCO validation set and obtain the detections on Obj45K test set using $\bar{v}$.
Then, given this set of detections, for different values of $\tau \in [0,1]$, we compute:
\begin{itemize}
    \item \gls{LaECE} for uncalibrated detection confidence scores;
    \item \gls{LaECE} for calibrated detection confidence scores using linear regression (LR); and
    \item LRP Error.
\end{itemize}
\cref{fig:sensitivity} demonstrates how these three quantities change for $\tau \in [0,1]$. 
Note that both for the uncalibrated and calibrated cases, \gls{LaECE} is not sensitive for $\tau \in [0.0, 0.5]$.
As for $\tau \in [0.5, 1.0]$, LaECE increases for the uncalibrated case due to the fact that the detection task becomes more challenging once a larger TP validation threshold $\tau$ is required and that the uncalibrated detections implies more over-confidence as $\tau$ increases.
Conversely, in this case, the calibration task becomes easier as the most of the detections are now FP.
As an insight, please consider the the extreme case that $\tau = 1$ in which a perfect localisation is required for a TP.
In this case, there is no TP and it is sufficient for a calibrator to assign a confidence score of $0.00$ to all detections and achieve perfect LaECE that is $0$.
Finally, as also analysed before \cite{LRPPAMI}, when $\tau$ increases, the detection task becomes more challenging, and therefore LRP Error, as the lower-better measure of accuracy, also increases.
This is because the number of TPs decreases and the number of FPs increases as $\tau$ increases.

While choosing the TP validation threshold $\tau$ for our \gls{SAOD} framework, we first consider that a proper $\tau$ should decompose the false positive and localisation errors properly. 
Having looked at the literature, the general consensus of object detection analysis tools \cite{Analyzer,TIDE} to split the false positive and localisation errors is achieved by employing an IoU of $0.10$.
As a result, following these works, we set $\tau=0.10$ throughout the paper unless otherwise noted.
Still, the TP validation threshold $\tau$ should be chosen by the requirements of the specific application.

%

%
\subsection{Derivation of Eq. \eqref{eq:finalECEoptimise_main}}
\label{app:subsec_cal_targetsetting}
In \cref{subsec:calibrationmethods}, we claim that the \gls{LaECE} for a bin reduces to: 
{\small
\begin{align}
\label{eq:finalECEoptimise__}
  &\Bigg\lvert \sum_{\substack{\hat{b}_i \in \hat{\mathcal{D}}_j^c \\ \psi(i) > 0}} \big( t^{cal}_i - \mathrm{IoU}(\hat{b}_i, b_{\psi(i)}) \big) + \sum_{\substack{\hat{b}_i \in \hat{\mathcal{D}}_j^c \\ \psi(i) \leq 0}} t^{cal}_i  \Bigg\rvert,
\end{align}
}
which allows us to set the target $t^{cal}_i$ as:
{\small
\begin{align}
\label{eq:targetcalibration}
    t^{cal}_i = 
\begin{cases}
    \mathrm{IoU}(\hat{b}_i, b_{\psi(i)}) , & \text{if } b_{\psi(i)} > 0\text{ ($i$ is true positive),} \\
    0,              &\text{otherwise ($i$ is false positive).} 
\end{cases}
\end{align}
}
In this section, we derive \eqref{eq:finalECEoptimise__} given the definition of \gls{LaECE} in Eq. \eqref{eq:odcalibrationfinal1} to justify our claim.

To start with, Eq. \eqref{eq:odcalibrationfinal1} defines \gls{LaECE} for class $c$ as
{\small
\begin{align}
\label{eq:odcalibrationfinal11}
   \mathrm{LaECE}^c 
   & = \sum_{j=1}^{J} \frac{|\hat{\mathcal{D}}^{c}_j|}{|\hat{\mathcal{D}}^{c}|} \left\lvert \bar{p}^{c}_{j} - \mathrm{precision}^{c}(j) \times \bar{\mathrm{IoU}}^{c}(j)  \right\rvert,
\end{align}}
which can be expressed as,
{\scriptsize
\begin{align}
 \sum_{j=1}^J \frac{|\hat{\mathcal{D}}_j^c|}{|\hat{\mathcal{D}}^c|} \left\lvert \bar{p}_{j} - \frac{\sum_{\hat{b}_k \in \hat{\mathcal{D}}_j^c, \psi(k) > 0} 1}{|\hat{\mathcal{D}}_j^c|} \frac{\sum_{\hat{b}_k \in \hat{\mathcal{D}}_j^c, \psi(k) > 0} \mathrm{IoU}(\hat{b}_k, b_{\psi(k)})}{\sum_{\hat{b}_k \in \hat{\mathcal{D}}_j^c, \psi(k) > 0} 1}  \right\rvert,
\end{align}
}%
as 
\begin{align}
 \mathrm{precision}^{c}(j) = \frac{\sum_{\hat{b}_k \in \hat{\mathcal{D}}_j^c, \psi(k) > 0} 1}{|\hat{\mathcal{D}}_j^c|} ,
\end{align}
and,
\begin{align}
 \bar{\mathrm{IoU}}^{c}(j) =  \frac{\sum_{\hat{b}_k \in \hat{\mathcal{D}}_j^c, \psi(k) > 0} \mathrm{IoU}(\hat{b}_k, b_{\psi(k)})}{\sum_{\hat{b}_k \in \hat{\mathcal{D}}_j^c, \psi(k) > 0} 1}.
\end{align}
The expression ${\sum_{\hat{b}_k \in \hat{\mathcal{D}}_j^c, \psi(k) > 0} 1}$ (in the nominator of $\mathrm{precision}^{c}(j)$ and the denominator of $\bar{\mathrm{IoU}}^{c}(j)$) corresponds to the number of TPs. 
Canceling out these terms yield
{
\begin{align}
\label{eq:beforepj}
 \sum_{j=1}^J \frac{|\hat{\mathcal{D}}_j^c|}{|\hat{\mathcal{D}}^c|} \left\lvert \bar{p}_{j} - \frac{\sum_{\hat{b}_k \in \hat{\mathcal{D}}_j^c, \psi(k) > 0} \mathrm{IoU}(\hat{b}_k, b_{\psi(k)})}{|\hat{\mathcal{D}}_j^c|} \right\rvert. 
\end{align}
}

$\bar{p}_{j}$, the average of the confidence score in bin $j$, can similarly be obtained as: 
\begin{align}
\bar{p}_{j} = \frac{\sum_{\hat{b}_k \in \hat{\mathcal{D}}_j^c} \hat{p}_{k}}{|\hat{\mathcal{D}}_j^c|},
\end{align}
and replacing  $\bar{p}_{j}$ in Eq. \eqref{eq:beforepj} yields
{
\begin{align}
\label{eq:afterpj}
 \sum_{j=1}^J \frac{|\hat{\mathcal{D}}_j^c|}{|\hat{\mathcal{D}}^c|} \left\lvert \frac{\sum_{\hat{b}_k \in \hat{\mathcal{D}}_j^c} \hat{p}_{k}}{|\hat{\mathcal{D}}_j^c|} - \frac{\sum_{\hat{b}_k \in \hat{\mathcal{D}}_j^c, \psi(k) > 0} \mathrm{IoU}(\hat{b}_k, b_{\psi(k)})}{|\hat{\mathcal{D}}_j^c|} \right\rvert. 
\end{align}
}
Since $a|x|=|ax|$ if $a \geq 0$, we take $\frac{|\hat{\mathcal{D}}_j^c|}{|\hat{\mathcal{D}}^c|}$ inside the absolute value where $|\hat{\mathcal{D}}_j^c|$ terms cancel out: 
{
\begin{align}
 \sum_{j=1}^J \left\lvert \frac{\sum_{k \in \hat{\mathcal{D}}_j^c} \hat{p}_{k}}{|\hat{\mathcal{D}}^c|} - \frac{\sum_{\hat{b}_k \in \hat{\mathcal{D}}_j^c, \psi(k) > 0} \mathrm{IoU}(\hat{b}_k, b_{\psi(k)})}{|\hat{\mathcal{D}}^c|} \right\rvert. 
\end{align}
}
Splitting $\sum_{\hat{b}_k \in \hat{\mathcal{D}}_j} \hat{p}_{k}$ for true positives and false positives as $\sum_{\hat{b}_k \in \hat{\mathcal{D}}_j, \psi(k) > 0} \hat{p}_{k}$ and $\sum_{\hat{b}_k \in \hat{\mathcal{D}}_j, \psi(k) \leq 0}  \hat{p}_{k}$ respectively, we have 
\begin{align}
\label{eq:LaECEtarget}
\begin{split}
  \sum_{j=1}^J&\left\lvert \frac{\sum_{\hat{b}_k \in \hat{\mathcal{D}}_j^c, \psi(k) > 0} \hat{p}_{k} + \sum_{\hat{b}_k \in \hat{\mathcal{D}}_j^c, \psi(k) \leq 0} \hat{p}_{k} }{|\hat{\mathcal{D}}^c|}  \right. \\
  &\left. - \frac{\sum_{\hat{b}_k \in \hat{\mathcal{D}}_j^c, \psi(k) > 0,} \mathrm{IoU}(\hat{b}_k, b_{\psi(k)})}{|\hat{\mathcal{D}}^c|}  \right\rvert.
 \end{split}
\end{align}
Considering that Eq. \eqref{eq:LaECEtarget} is minimized when the error for each bin $j$ is minimized as 0, we now focus on a single bin $j$. Note also that for each bin $j$, $|\hat{\mathcal{D}}^c|$ is a constant.
As a result, minimizing the following expression minimizes the error for each bin, and also \gls{LaECE},
{\scriptsize
\begin{align}
\label{eq:LaECEtarget_}
\begin{split}
  \left\lvert \sum_{\hat{b}_k \in \hat{\mathcal{D}}_j^c, \psi(k) > 0} \hat{p}_{k} + \sum_{\hat{b}_k \in \hat{\mathcal{D}}_j^c, \psi(k) \leq 0} \hat{p}_{k}  - \sum_{\hat{b}_k \in \hat{\mathcal{D}}_j^c, \psi(k) > 0,} \mathrm{IoU}(\hat{b}_k, b_{\psi(k)})  \right\rvert.
 \end{split}
\end{align}
}
By rearranging the terms, we have
{\small
\begin{align}
\label{eq:finalECEoptimise}
  &\left\lvert \sum_{\hat{b}_k \in \hat{\mathcal{D}}_j^c, \psi(k) > 0} \left( \hat{p}_{k} - \mathrm{IoU}(\hat{b}_k, b_{\psi(k)}) \right) + \sum_{\hat{b}_k \in \hat{\mathcal{D}}_j^c, \psi(k) \leq 0} \hat{p}_{k}  \right\rvert,
\end{align}
}
which reduces to Eq. \eqref{eq:finalECEoptimise_main} once by setting $\hat{p}_{k}$ by $t^{cal}_k$.
This concludes the derivation and validates how we construct the targets $t^{cal}_k$ while obtaining the pairs to train the calibrator.

\begin{figure*}[t]
        \captionsetup[subfigure]{}
        \centering
        \begin{subfigure}[b]{0.30\textwidth}
        \includegraphics[width=\textwidth]{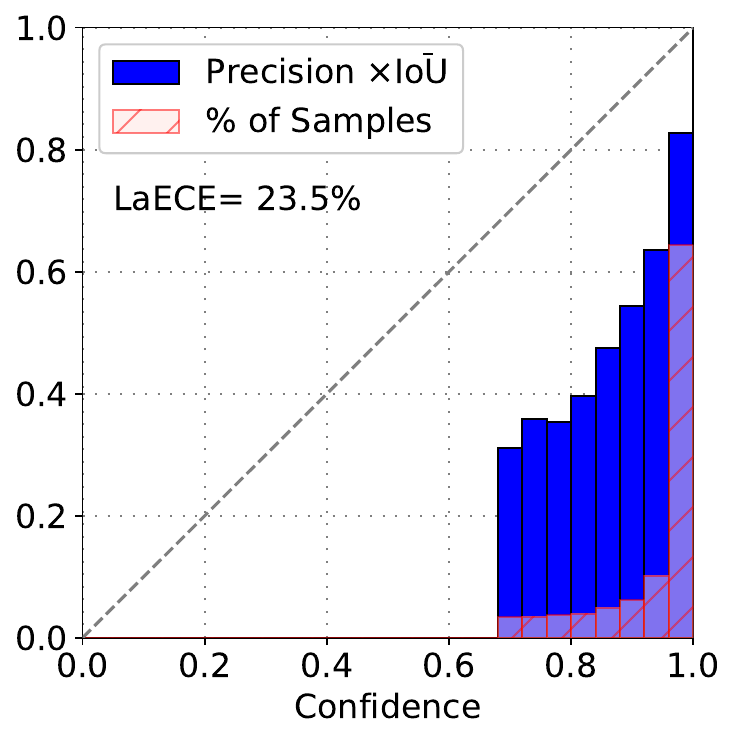}
        \caption{Uncalibrated}
        \end{subfigure}
        \begin{subfigure}[b]{0.30\textwidth}
        \includegraphics[width=\textwidth]{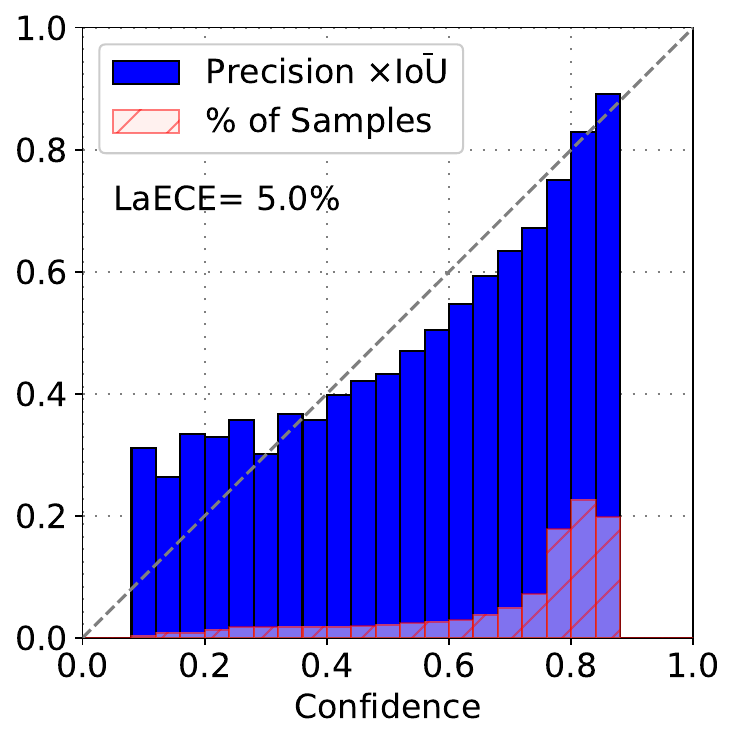}
        \caption{Calibrated by linear regression}
        \end{subfigure}
        \begin{subfigure}[b]{0.30\textwidth}
        \includegraphics[width=\textwidth]{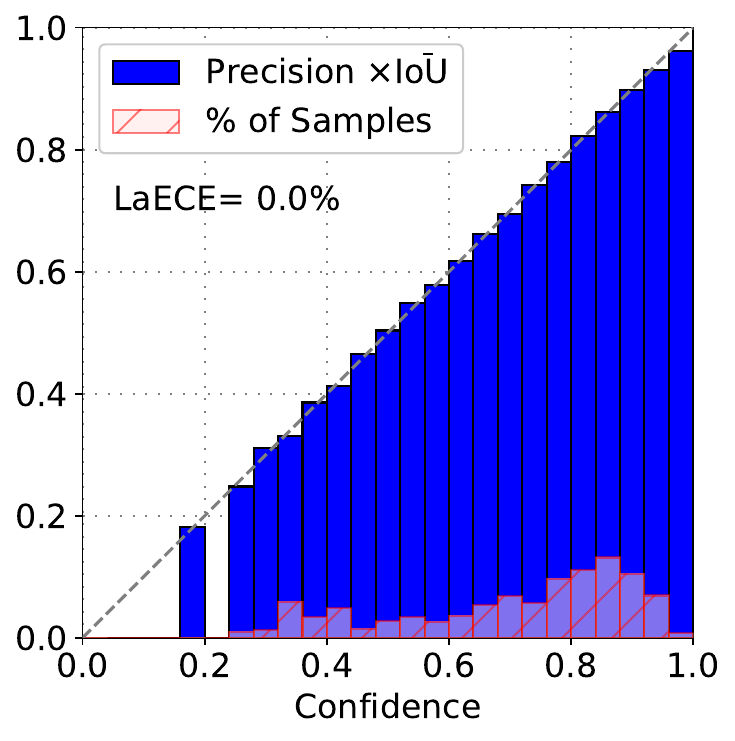}
        \caption{Calibrated by isotonic regression}
        \end{subfigure}
        
        \begin{subfigure}[b]{0.30\textwidth}
        \includegraphics[width=\textwidth]{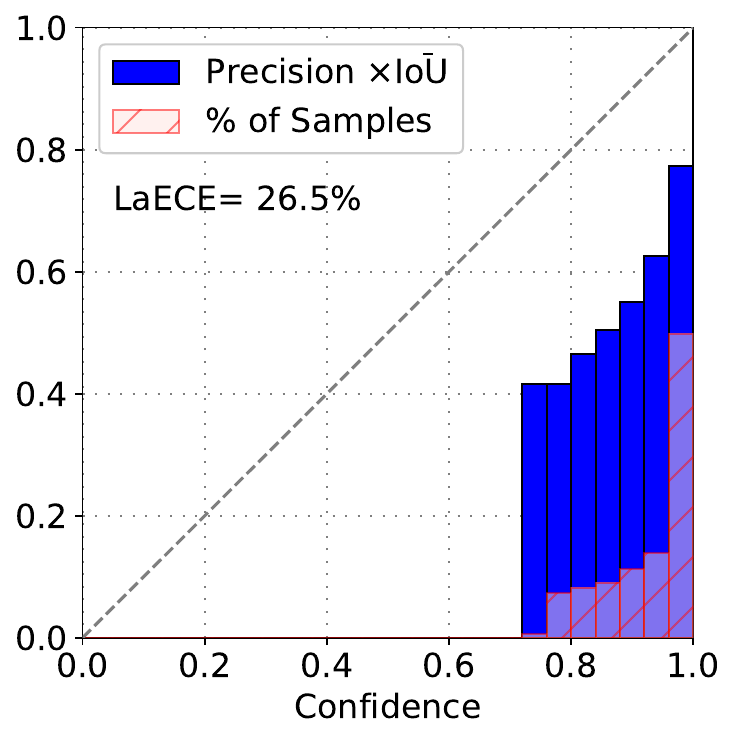}
        \caption{Uncalibrated}
        \end{subfigure}
        \begin{subfigure}[b]{0.30\textwidth}
        \includegraphics[width=\textwidth]{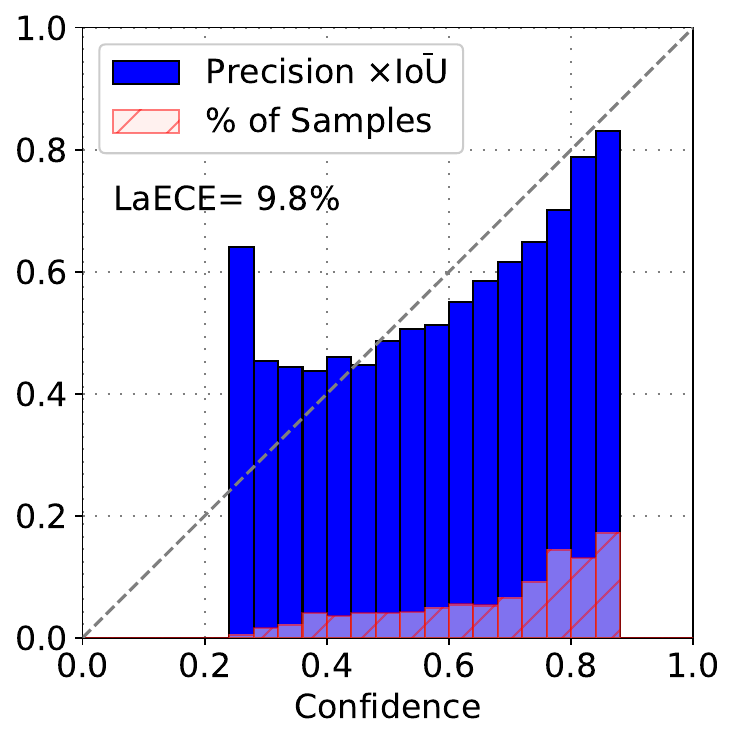}
        \caption{Calibrated by linear regression}
        \end{subfigure}
        \begin{subfigure}[b]{0.30\textwidth}
        \includegraphics[width=\textwidth]{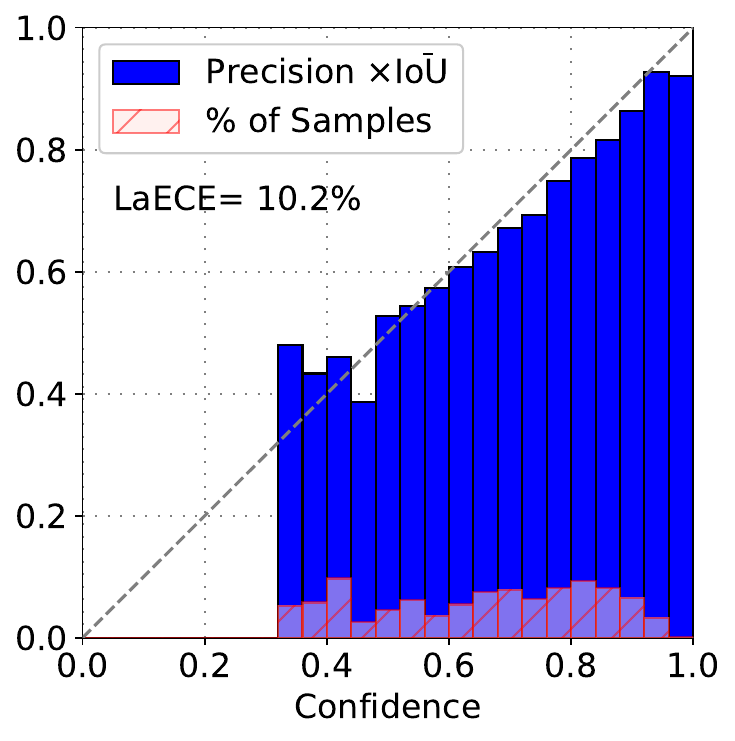}
        \caption{Calibrated by isotonic regression}
        \end{subfigure}

        \begin{subfigure}[b]{0.30\textwidth}
        \includegraphics[width=\textwidth]{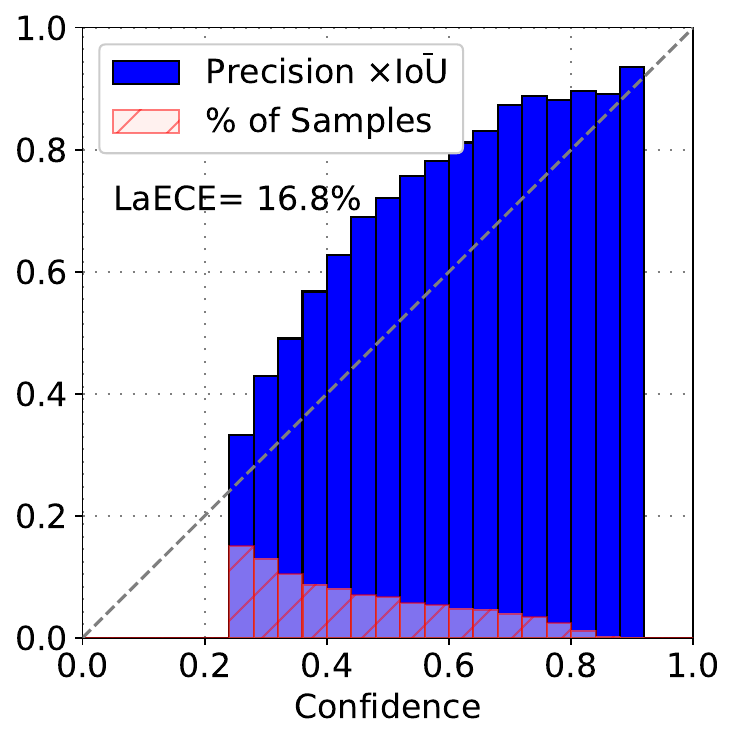}
        \caption{Uncalibrated}
        \end{subfigure}
        \begin{subfigure}[b]{0.30\textwidth}
        \includegraphics[width=\textwidth]{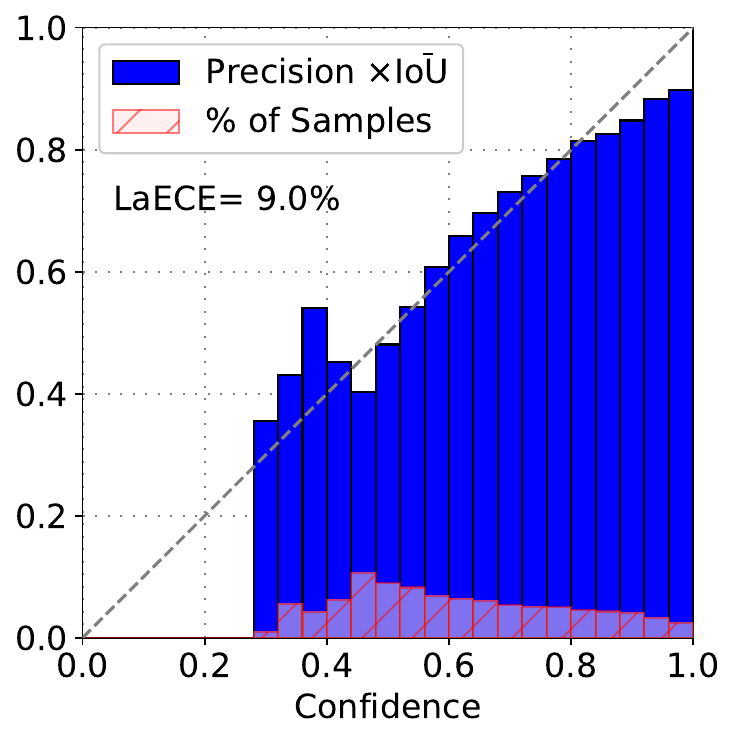}
        \caption{Calibrated by linear regression}
        \end{subfigure}
        \begin{subfigure}[b]{0.30\textwidth}
        \includegraphics[width=\textwidth]{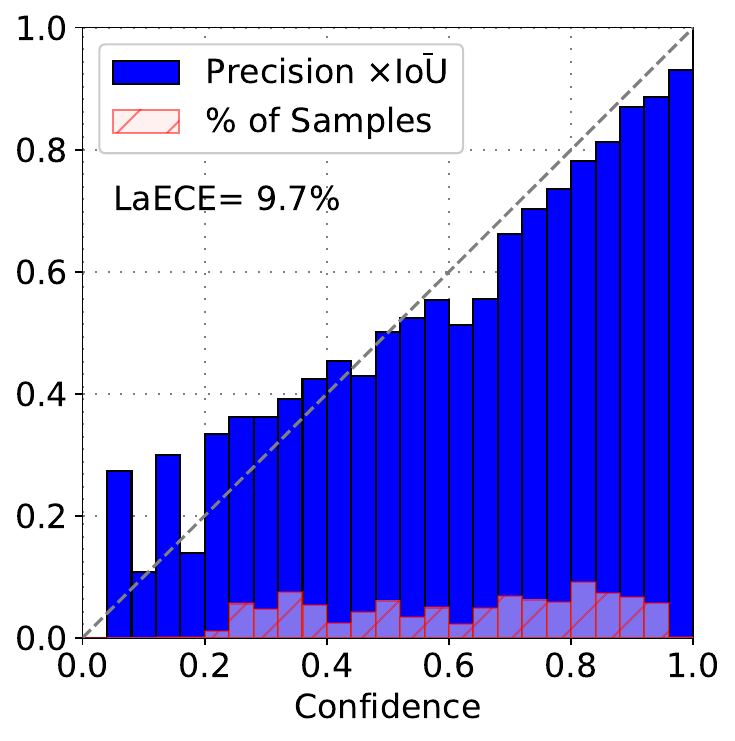}
        \caption{Calibrated by isotonic regression}
        \end{subfigure}
        \caption{(First row) Reliability diagrams of F R-CNN on SAOD-AV $\valdata$, which is used to obtain the set that we used for training the calibrators. (Second row) Reliability diagrams of F R-CNN on SAOD-AV $\indata$ (BDD45K). (Third row)  Reliability diagrams of ATSS on SAOD-AV $\indata$ (BDD45K). Linear regression and isotonic regression improve the calibration performance of both over-confidence F-RCNN (compare (e) and (f) with (d)) and under-confidence ATSS (compare (h) and (i) with (g)). }
        \label{fig:frcnnreliabilityavod}
\end{figure*}

\subsection{More Examples of Reliability Diagrams}
We provide more examples of reliability diagrams in \cref{fig:frcnnreliabilityavod} on F-RCNN and ATSS on SAOD-AV.
To provide insight on how the error on the set that the calibrator is trained with, \cref{fig:frcnnreliabilityavod} (a-c) show the reliability diagrams on the val set as the split used to train the calibrator.
On the val set, we observe that the isotonic regression method for calibration results in an LaECE of $0.0$; thereby overfitting to the training data (\cref{fig:frcnnreliabilityavod}(b)).

On the other hand, the linear regression method ends up with a training LaECE of $5.0$ (\cref{fig:frcnnreliabilityavod}(c)).
Consequently, we observe that linear regression performs slightly better than isotonic regression on the BDD45K test split (\cref{fig:frcnnreliabilityavod} (e,f)).
Besides, when we compare \cref{fig:frcnnreliabilityavod}(e,f) with \cref{fig:frcnnreliabilityavod}(d), we observe that both isotonic regression and linear regression decrease the overconfidence of the baseline F-RCNN \cref{fig:frcnnreliabilityavod}.
As a different type of calibration error, ATSS shown in \cref{fig:frcnnreliabilityavod}(g) is under-confident.
Again, linear regression and isotonic regression improves the calibration performance of ATSS. 
This further validates on SAOD-AV that such post-hoc calibration methods are effective. 

\begin{table}
    \centering
    \small
    \setlength{\tabcolsep}{0.4em}
    \caption{Dummy detections decrease LaECE superficially with no effect on AP due to top-$k$ survival. LRP Error penalizes dummy detections and requires the detections to be thresholded properly. COCO val set is used.}
    \label{tab:dummydet}
    \begin{tabular}{|c|c|c||c|c|c|} \hline
         Detector&Dummy det.&det/img.&$\mathrm{LaECE} \downarrow$&$\mathrm{AP} \uparrow$&$\mathrm{LRP} \downarrow$ \\ \hline
    \multirow{4}{*}{F-RCNN}&None&$33.9$&$15.1$&$39.9$&$86.5$\\
    &up to 100&$100$&$3.9$&$39.9$&$96.8$\\
    &up to 300&$300$&$1.4$&$39.9$&$98.8$\\
    &up to 500&$500$&$0.9$&$39.9$&$99.2$\\ \hline
    \multirow{4}{*}{ATSS}&None&$86.4$&$7.7$&$42.8$&$95.1$\\
    &up to 100&$100$&$6.0$&$42.8$&$96.2$\\
    &up to 300&$300$&$1.8$&$42.8$&$98.9$\\
    &up to 500&$500$&$1.1$&$42.8$&$99.3$\\ \hline
    \end{tabular}
\end{table}
\begin{table}
    \centering
    \small
    \setlength{\tabcolsep}{0.4em}
    \caption{To avoid superficial LaECE gain (Table \ref{tab:dummydet}); we adopt LRP Error that requires the detections to be thresholded properly. We use LRP-optimal thresholding to obtain class-wise thresholds. Results are from COCO val set. with 7.3 objects/image on average.}
    \label{tab:threshold}
    \begin{tabular}{|c|c|c||c|c|c|} \hline
         Detector&Threshold&det/img.&$\mathrm{LaECE} \downarrow$&$\mathrm{AP} \uparrow$&$\mathrm{LRP} \downarrow$ \\ \hline
    \multirow{5}{*}{F-RCNN}&None&$33.9$&$15.1$&$39.9$&$86.5$\\
    &0.30&$11.2$&$27.5$&$38.0$&$67.6$\\
    &0.50&$7.4$&$27.6$&$36.1$&$62.1$\\
    &0.70&$5.2$&$24.5$&$33.2$&$61.5$\\ \cline{2-6}
    &LRP-opt.&$6.1$&$26.1$&$34.6$&$61.1$\\\hline
    \multirow{5}{*}{ATSS}&None&$86.4$&$7.7$&$42.8$&$95.1$\\
    &0.30&$5.2$&$20.2$&$35.3$&$60.5$\\
    &0.50&$2.0$&$26.6$&$19.7$&$78.4$\\
    &0.70&$0.3$&$12.3$&$3.9$&$96.3$\\ \cline{2-6}
    &LRP-opt.&$6.0$&$18.3$&$36.7$&$60.2$\\ \hline
    \end{tabular}
\end{table}

\subsection{Numerical Values of Fig. \ref{fig:calibration}} 
Tables \ref{tab:dummydet} and \ref{tab:threshold} present the numerical values used in the Fig. \ref{fig:calibration}(a) and Fig. \ref{fig:calibration}(b) respectively. 
Please refer to \cref{subsec:relation} for the details of the tables and discussion.

\renewcommand{\thesection}{E}

\section{Further Details on \gls{SAOD} and \gls{SAODet}s} \label{app:SAOD}
This section provides further details and analyses on the \gls{SAOD} task and the \gls{SAODet}s.
\begin{algorithm}
    \caption{Making an object detector self-aware  \label{alg:train}}
    \small
    \begin{algorithmic}[1]
        \Procedure{MakingSelfAware}{$\traindata$, $\valdata$}
        \State Train a standard detector $f(\cdot)$ on $\traindata$
        \State Obtain pseudo OOD set $\valdata^-$ by replacing objects in $\valdata$ with zeros (\cref{sec:ood})
        \State Remove the images with no objects from $\valdata$, denote the resulting set by $\valdata^+$
        \State Make inference on $\valdata^-$ by including top-100 detections from each image, i.e., $\mathcal{D}_{val}^- = \{ f(X_i)\}_{X_i \in \hat{D}_{100}^-}$
        \State Make inference on $\valdata^+$ by including top-100 detections from each image, i.e., $\mathcal{D}_{val}^+ = \{ f(X_i)\}_{X_i \in \hat{D}_{100}^+}$ 
        \State Cross-validate $\bar{u}$, the image-level uncertainty threshold, on $\mathcal{D}_{val}^+$ and $\mathcal{D}_{val}^-$ using mean(top-3) of the uncertainty scores against the Balanced Accuracy as the performance measure (\cref{sec:ood})
        \State Cross-validate $\bar{v}^c$, the detection-level threshold of class $c$, on $\hat{D}_{100}^+$ using LRP-optimal thresholding (\cref{subsec:relation})
        \State Remove all detections of class $c$ in $\hat{D}_{100}^+$ with score less than $\bar{v}^c$ to obtain thresholded detections $\hat{D}_{thr}$ 
        \State Using $\hat{D}_{thr}$, train a linear regression calibrator $\zeta^c(\cdot)$ for each class $c$ (\cref{subsec:calibrationmethods})
        \State \textbf{return} $f(\cdot)$, $\bar{u}$, $\{\bar{v}^c\}_{c=1}^{C}$, $\{\zeta^c(\cdot)\}_{c=1}^{C}$
        \EndProcedure
    \end{algorithmic}
\end{algorithm}

\subsection{Algorithms to Make an Object Detector Self-Aware}
\label{subsec:makesaod}
In \cref{sec:evaluation}, we summarized how we convert an object detector into a self-aware one.
Specifically, to do so, we use mean(top-3) and obtain an uncertainty threshold $\bar{u}$ through cross-validation using pseudo \gls{OOD} set approach (\cref{sec:ood}), obtain the detections through LRP-optimal thresholding (\cref{subsec:relation}) and calibrate the detection scores using linear regression as discussed in \cref{subsec:calibrationmethods}.
Here, we present the following two algorithms to include the further details on how we incorporate these features into a conventional object detector:
\begin{enumerate}
    \item \textit{The algorithm to make an object detector self-aware in Alg. \ref{alg:train}.} The aim of Alg. \ref{alg:train} is to obtain 
    \begin{itemize}
        \item the image-level uncertainty threshold $\bar{u}$;
        \item the detection confidence score thresholds for each class $\{\bar{v}^c\}_{c=1}^{C}$;
        \item the calibrators for each class $\{\zeta^c(\cdot)\}_{c=1}^{C}$; and
        \item a conventional object detector $f(\cdot)$ on which these features will be incorporated into.
    \end{itemize}
    To do so, after training the conventional object detector $f(\cdot)$ using $\traindata$ (line 2), we first obtain the image-level uncertainty threshold $\bar{u}$ by using our pseudo OOD set approach as described in \cref{sec:ood} (lines 3-7).
    While doing that, we do not apply detection-level thresholding yet; ensuring us to have at least 3 detections from each image on which we compute the image-level uncertainty using mean(top-3) of the uncertainty scores.
    While we enforce this by keeping a maximum of 100 detections following AP-based evaluation, we only use the top-3 scoring detections to compute the image-level uncertainty.
    Then, after cross-validating $\bar{u}$, we cross-validate $\{\bar{v}^c\}_{c=1}^{C}$ for each class (line 8) and using only the thresholded detections we train the calibrators (lines 9-10).
    This procedure allows us to incorporate necessary features into a conventional object detector, making it self-aware.
    \item \textit{The inference algorithm of a \gls{SAODet} in Alg. \ref{alg:infer}.}
    Given a \gls{SAODet}, the inference on a given image $X$ is as follows.
    We first compute the uncertainty of the detector on image $X$, denoted by $\mathcal{G}(X)$, following the same method in Alg. \ref{alg:train}, that is mean(top-3)  (lines 2-3).
    Then, the rejection or acceptance decision is made by comparing $\mathcal{G}(X)$ by the cross-validated threshold $\bar{u}$ (line 5).
    If the image is rejected then, no detection is returned (line 7).
    Otherwise, if the detection is accepted, then its confidence is compared against the cross-validated detection confidence threshold (line 11); enabling us to differentiate between a useful and a low-scoring noisy detection.
    If the confidence of the detection is larger than the detection-level threshold, then it is added into the final set of detections also by a calibrated confidence score (line 12).
    Therefore, Alg. \ref{alg:infer} checks whether the detector is able to make a detection on the given image $X$, if so, it preserves the accurate detections obtained by the object detector by removing the noisy detections as well as calibrates the detection scores; applying all features of a \gls{SAODet} during inference.
\end{enumerate}

\begin{algorithm}
    \caption{The inference algorithm of a \gls{SAODet} given an image $X$  \label{alg:infer} (please also refer to Alg. \ref{alg:train} for the notation)}
    \small
    \begin{algorithmic}[1]
        \Procedure{Inference}{$f(\cdot)$, $\bar{u}$, $\{\bar{v}^c\}_{c=1}^{C}$, $\{\zeta^c(\cdot)\}_{c=1}^{C}$, $X$}
        \State $\hat{D}_{100} = \{\hat{c}_i, \hat{b}_i, \hat{p}_i\}^N = f(X)$ such that $N \leq 100$ 
        \State Estimate $\mathcal{G}(X)$, the image-level uncertainty of $f(X)$ on $X$, using mean(top-3) of the uncertainty scores (\cref{sec:ood})
        \State Initialize the thresholded detection set $\hat{D}_{thr} =$ \O
        \If{$\mathcal{G}(X) > \bar{u}$}
            \State $\hat{a}=0$ 
            \State \textbf{return} $\{\hat{a}, \hat{D}_{thr}\}$ // REJECT $X$ with $\hat{D}_{thr}$ being \O
        \Else
        \State $\hat{a}=1$
        \For{each detection $\{\hat{c}_i, \hat{b}_i, \hat{p}_i\} \in \hat{D}_{100}$}
            \If{$\hat{p}_i \geq \bar{v}^{\hat{c}_i}$}
                \State $\hat{D}_{thr} = \hat{D}_{thr} \cup \{\hat{c}_i, \hat{b}_i, \zeta^{\hat{c}_i}(\hat{p}_i)\}$
            \EndIf
        \EndFor
        \State \textbf{return} $\{\hat{a}, \hat{D}_{thr}\}$ // ACCEPT $X$
        \EndIf
        \EndProcedure
    \end{algorithmic}
\end{algorithm}

\begin{figure}[t]
        \captionsetup[subfigure]{}
        \centering
        \begin{subfigure}[b]{0.24\textwidth}
            \includegraphics[width=\textwidth]{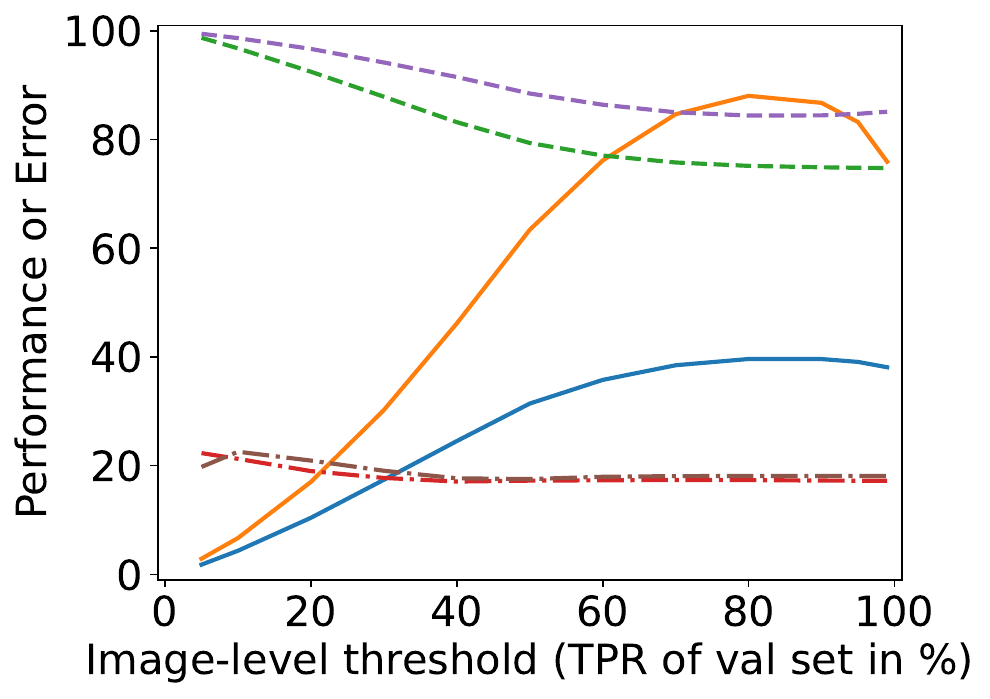}
            \caption{Image-level uncertainty thr.}
        \end{subfigure}
        \begin{subfigure}[b]{0.23\textwidth}
            \includegraphics[width=\textwidth]{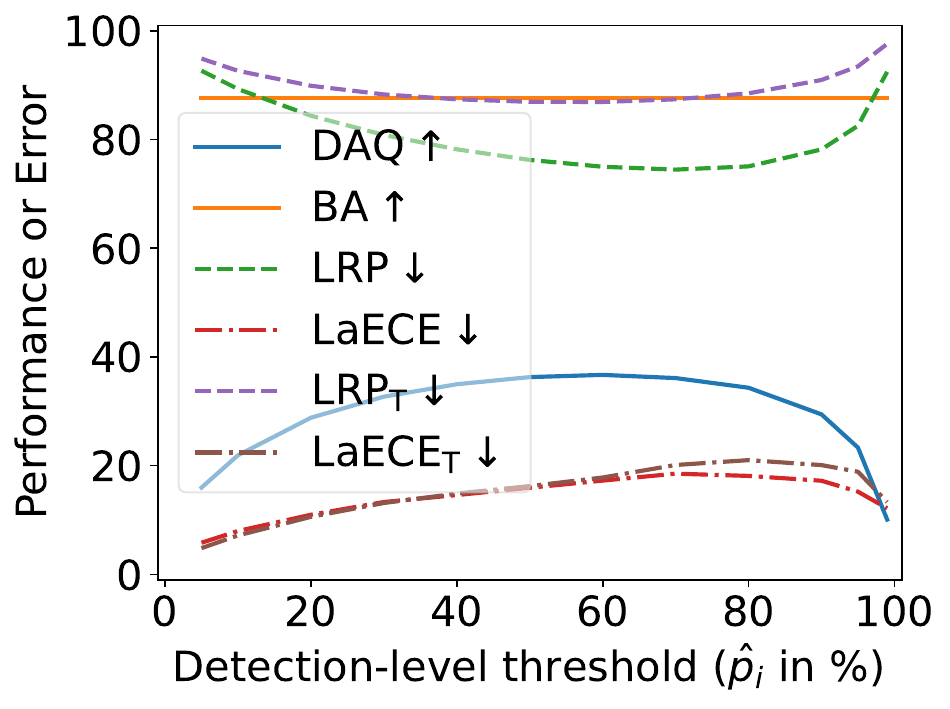}
            \caption{Detection confidence thr.}
        \end{subfigure}
        \caption{The effect of image- and detection-level thresholds.  DAQ (blue curve) decreases significantly for extreme cases such as when all images are rejected or all detections are accepted; implying its robustness to such cases. Here, for the sake of analysis simplicity, we use a single confidence score threshold ($\bar{v}$) obtained on the final detection scores $\hat{p}_i$ in (b) instead of class-wise approach that we used while building \gls{SAODet}s. }
        \label{fig:ablation_threshold}
\end{figure}

\begin{table}
    \centering
    \small
    \setlength{\tabcolsep}{0.05em}
    \caption{Effect of common improvements (epochs (Ep.), Multi-scale (MS) training, stronger backbones) on F-RCNN (SAOD-Gen).
    }
    \label{tab:universalimprovements}
    \begin{tabular}{|c|c|c||c|c|c|c|c|c||c|} \hline
    Ep.&MS&Backbone&DAQ&BA&mECE&LRP&$\mathrm{mECE_T}$&$\mathrm{LRP_T}$&AP\\ \hline
     12&\xmark&R50&$38.5$&$88.0$&$\mathbf{16.4}$&$76.6$&$\mathbf{16.8}$&$85.0$&$24.8$\\ 
     36&\xmark&R50&$38.4$&$87.4$&$18.7$&$75.9$&$20.5$&$85.0$&$25.5$\\ 
    36&\cmark&R50&$39.7$&$87.7$&$17.3$&$74.9$&$18.1$&$84.4$&$27.0$\\
    36 &\cmark&R101&$42.0$&$\mathbf{88.1}$&$17.5$&$73.4$&$19.0$&$82.8$&$28.7$\\
    36 &\cmark&R101-DCN&$\mathbf{45.9}$&$87.4$&$17.3$&$\mathbf{70.8}$&$19.4$&$\mathbf{79.7}$&$\mathbf{31.8}$\\
     \hline 
    \end{tabular}
\end{table}

\begin{figure*}[t]
        \captionsetup[subfigure]{}
        \centering
        \begin{subfigure}[b]{0.70\textwidth}
        \includegraphics[width=\textwidth]{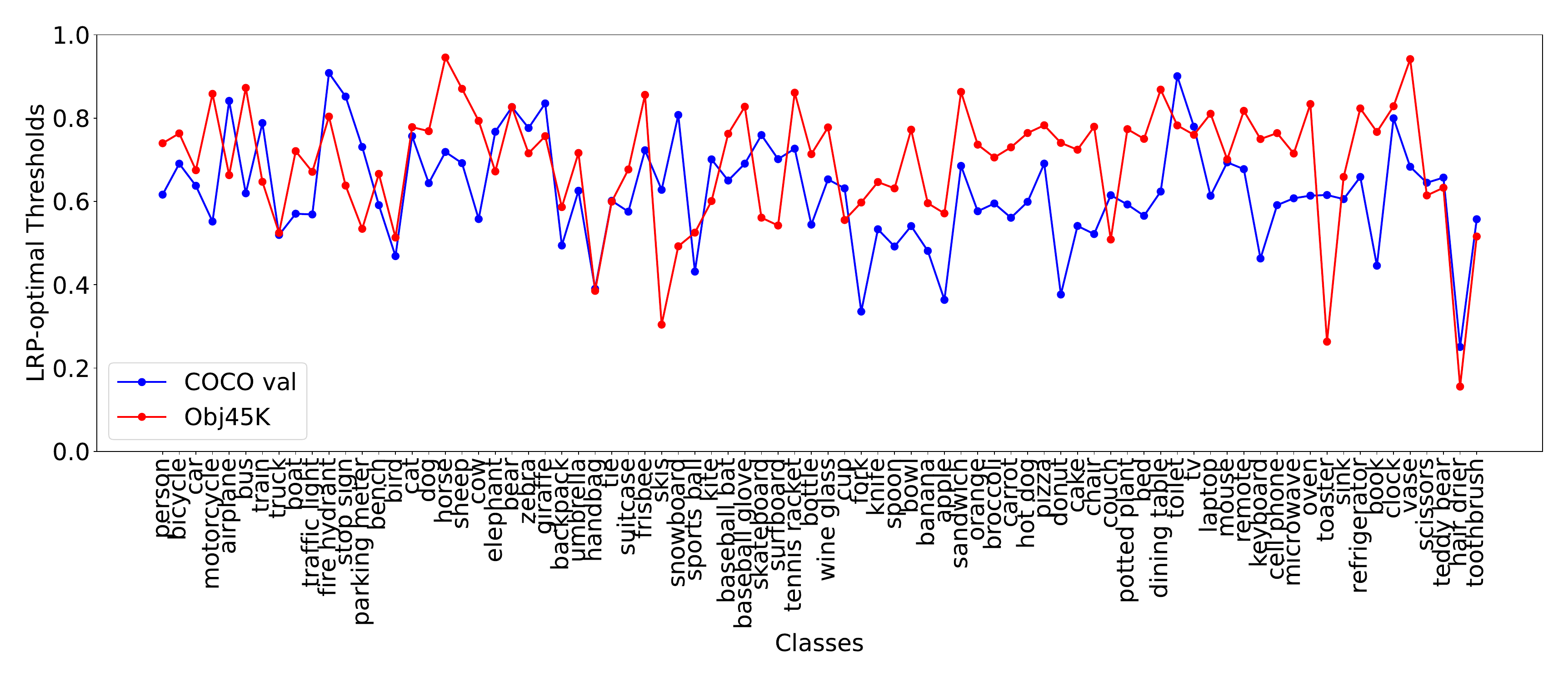}
        \caption{F-RCNN (SAOD-Gen)}
        \end{subfigure}
        \begin{subfigure}[b]{0.28\textwidth}
        \includegraphics[width=\textwidth]{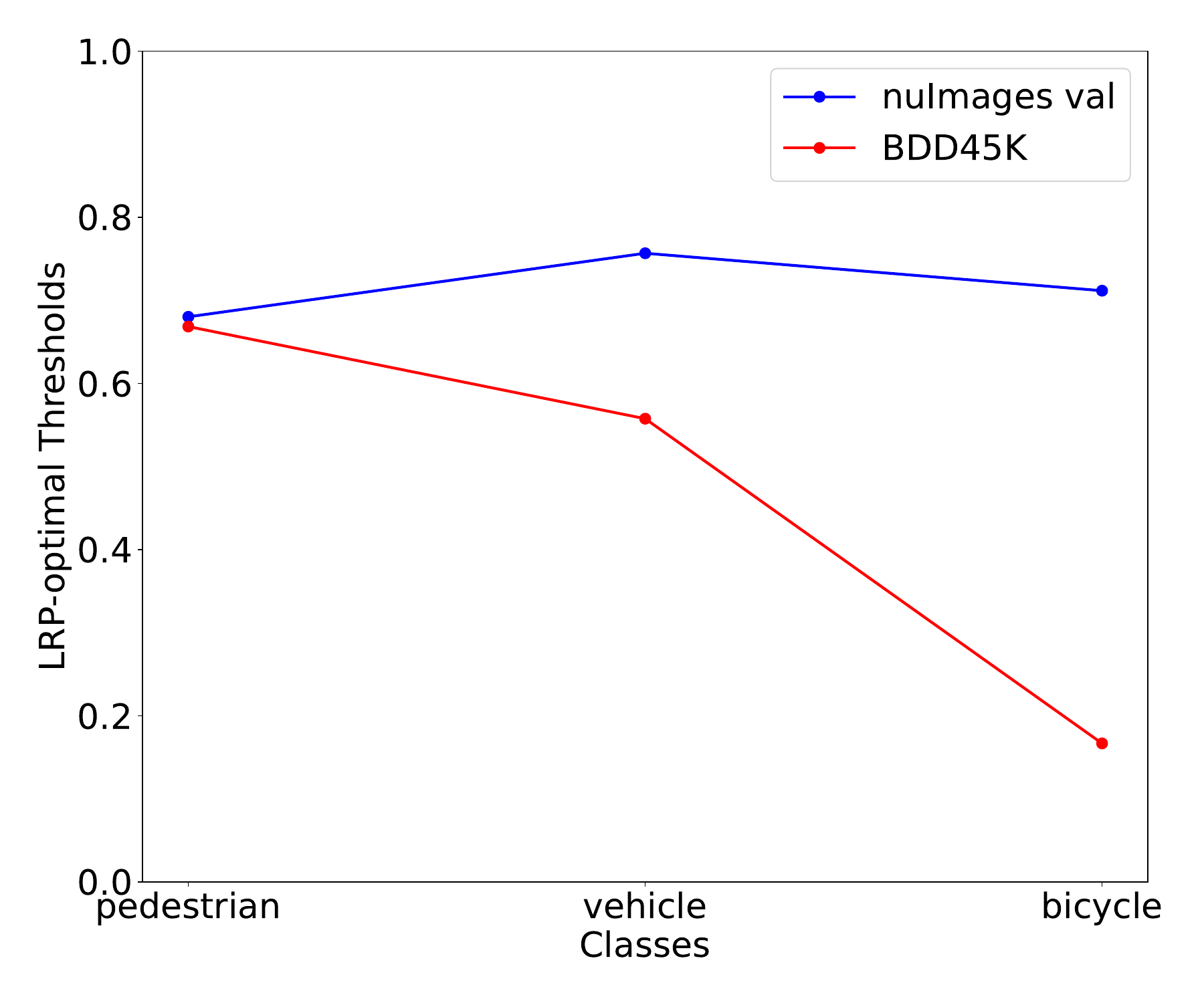}
        \vspace{20pt}
        \caption{F-RCNN (SAOD-AV)}
        \end{subfigure}

        \begin{subfigure}[b]{0.70\textwidth}
        \includegraphics[width=\textwidth]{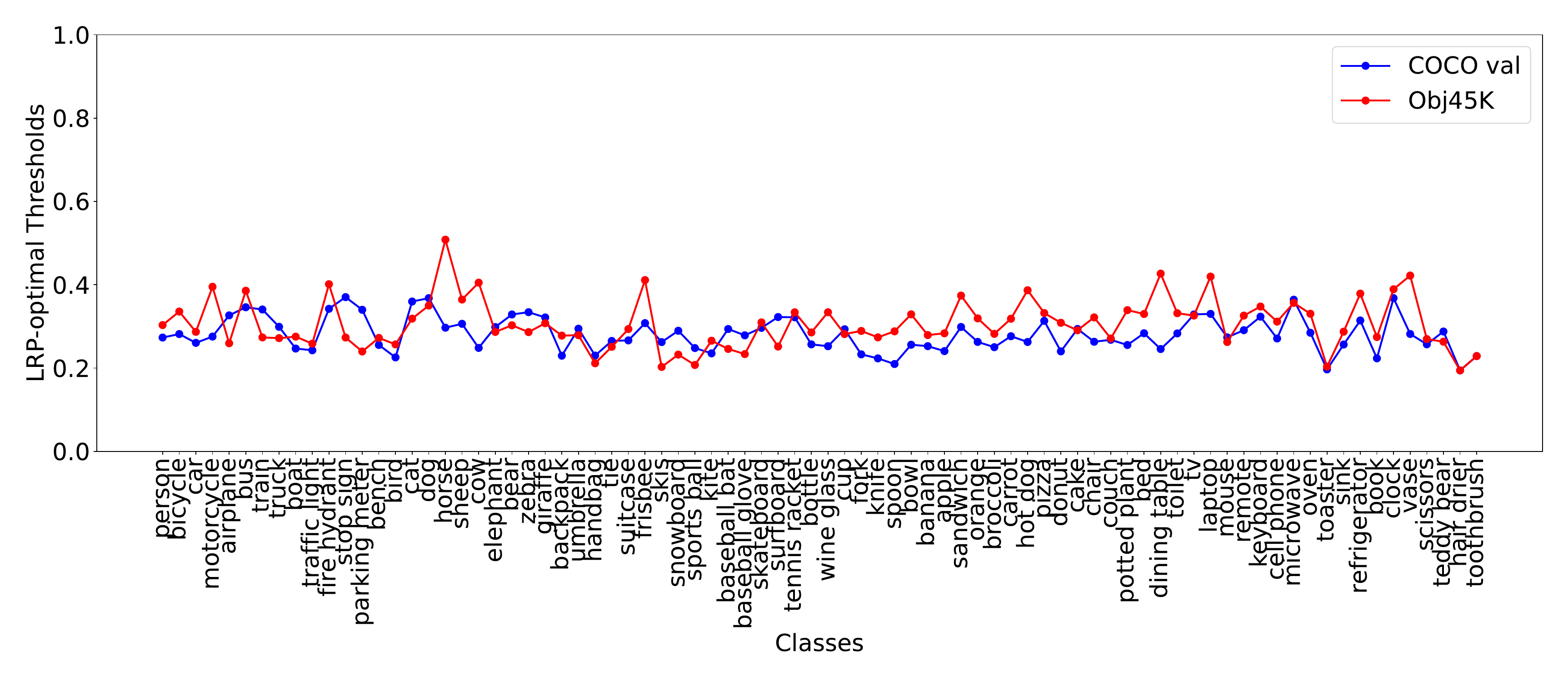}
        \caption{ATSS (SAOD-Gen)}
        \end{subfigure}
        \begin{subfigure}[b]{0.28\textwidth}
        \includegraphics[width=\textwidth]{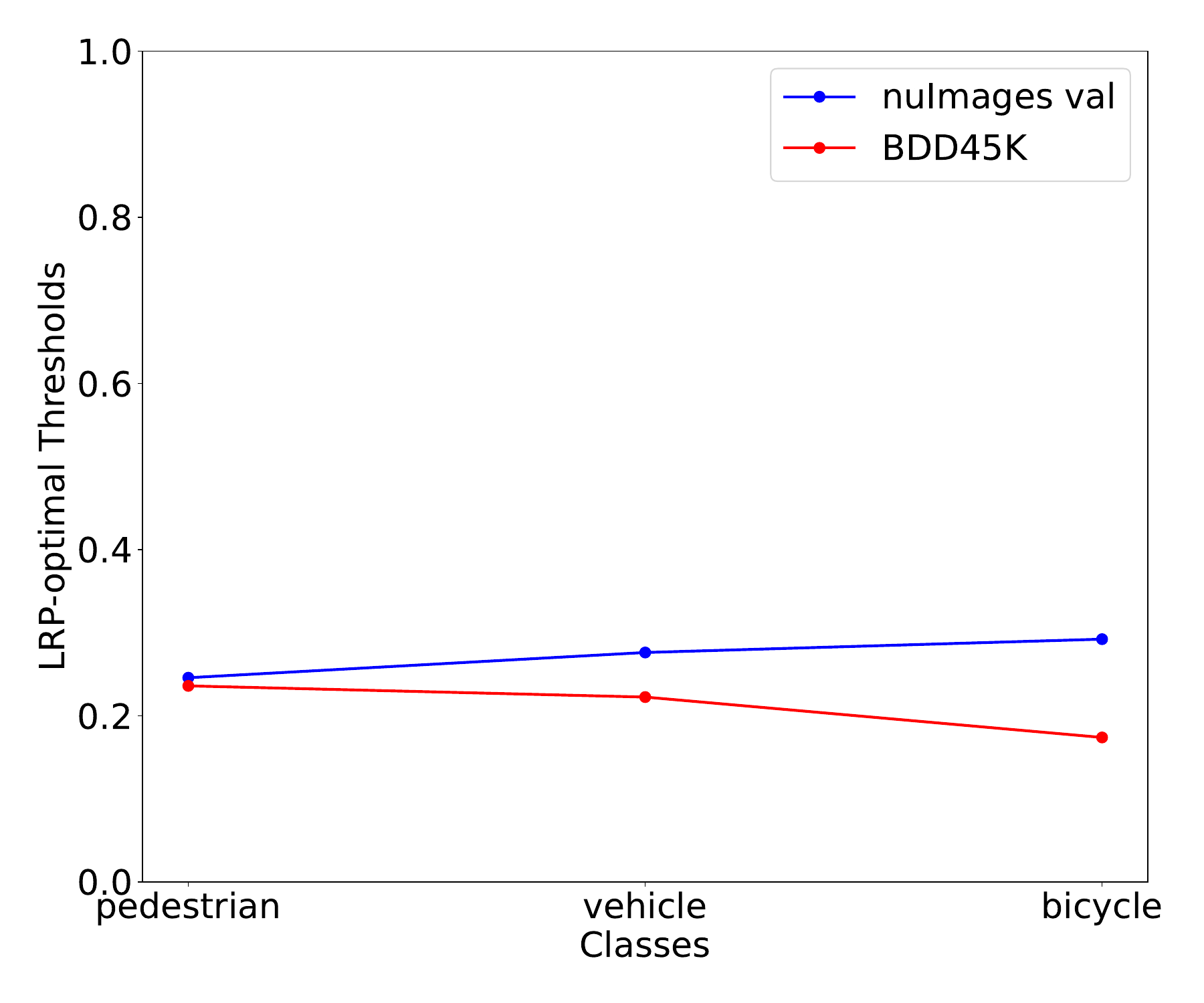}
        \vspace{20pt}
        \caption{ATSS (SAOD-AV)}
        \end{subfigure}
        \caption{Comparison of (i) LRP-optimal thresholds obtained on $\valdata$ as presented and used in the paper (blue lines); and (ii) LRP-optimal thresholds obtained on $\indata$ as oracle thresholds (red lines). Owing to the domain shift between $\valdata$ and $\indata$, the optimal thresholds do not match exactly. The thresholds between $\valdata$ and $\indata$ are relatively more similar for SAOD-Gen compared to SAOD-AV.}
        \label{fig:lrpoptimal}
\end{figure*}

\begin{table*}
    \def\s{\hspace*{0.5ex}}
    \def\r{\hspace*{1ex}}
    \small
    \centering
    \setlength{\tabcolsep}{0.03em}
    \caption{Evaluating self-aware object detectors. In addition to \cref{tab:evaluation}, this table includes the components of the LRP Error for more insight. Particularly, $\mathrm{LRP_{Loc}}$, $\mathrm{LRP_{FP}}$, $\mathrm{LRP_{FN}}$ correspond to the average 1-IoU of TPs, 1-precision and 1-recall respectively.}
    \label{tab:evaluation_}
    \scalebox{0.95}{
    \begin{tabular}{@{}c@{\r}c@{\r}|@{\r}c@{\r}|@{\r}c@{\s}c@{\s}c@{\r}|@{\r}c@{\s}c@{\s}c@{\s}c@{\s}c@{\s}c@{\r}|@{\r}c@{\s}c@{\s}c@{\s}c@{\s}c@{\s}c}
    \toprule
    \midrule
         &Self-aware&\multirow{2}{*}{$\scriptstyle\mathrm{DAQ}\uparrow$}&\multicolumn{3}{c@{\r}|@{\r}}{$\ooddata$ vs. $\indata$}&\multicolumn{6}{c@{\r}|@{\r}}{$\indata$}&\multicolumn{6}{c}{$\shiftdata$} \\
         &Detector& &$\scriptstyle\mathrm{BA}\uparrow$&$\scriptstyle\mathrm{TPR}\uparrow$&$\scriptstyle\mathrm{TNR}\uparrow$&$\scriptstyle\mathrm{IDQ}\uparrow$&$\scriptstyle\mathrm{LaECE}\downarrow$&$\scriptstyle\mathrm{LRP}\downarrow$&$\scriptstyle\mathrm{LRP_{Loc}}\downarrow$&$\scriptstyle\mathrm{LRP_{FP}}\downarrow$&$\scriptstyle\mathrm{LRP_{FN}}\downarrow$&$\scriptstyle\mathrm{IDQ}\uparrow$&$\scriptstyle\mathrm{LaECE}\downarrow$&$\scriptstyle\mathrm{LRP}\downarrow$&$\scriptstyle\mathrm{LRP_{Loc}}\downarrow$&$\scriptstyle\mathrm{LRP_{FP}}\downarrow$&$\scriptstyle\mathrm{LRP_{FN}}\downarrow$\\
    \midrule
    \multirow{4}{*}{\rotatebox[origin=c]{90}{Gen}}
    &SA-F-RCNN&$39.7$&$87.7$&$\mathbf{94.7}$&$81.6$&$38.5$&$17.3$&$74.9$&$20.4$&$48.5$&$52.3$&$26.2$&$18.1$&$84.4$&$21.9$&$52.2$&$72.4$\\
    &SA-RS-RCNN&$41.2$&$\mathbf{88.9}$&$92.8$&$85.3$&$39.7$&$17.1$&$73.9$&$19.3$&$47.8$&$51.9$&$27.5$&$\mathbf{17.8}$&$83.5$&$20.4$&$50.8$&$72.1$\\
    &SA-ATSS&$41.4$&$87.8$&$93.1$&$83.0$&$39.7$&$16.6$&$74.0$&$\mathbf{18.5}$&$47.8$&$52.8$&$27.8$&$18.2$&$83.2$&$\mathbf{20.2}$&$53.2$&$71.1$\\
    &SA-D-DETR&$\mathbf{43.5}$&$\mathbf{88.9}$&$90.0$&$\mathbf{87.8}$&$\mathbf{41.7}$&$\mathbf{16.4}$&$\mathbf{72.3}$&$18.8$&$\mathbf{45.1}$&$\mathbf{50.7}$&$\mathbf{29.6}$&$17.9$&$\mathbf{81.9}$&$20.4$&$\mathbf{49.6}$&$\mathbf{69.4}$\\
    \midrule
    \multirow{2}{*}{\rotatebox[origin=c]{90}{AV}} &SA-F-RCNN&$43.0$&$\mathbf{91.0}$&$94.1$&$\mathbf{88.2}$&$41.5$&$9.5$&$73.1$&$26.3$&$13.2$&$58.1$&$28.8$&$7.2$&$83.0$&$26.7$&$12.2$&$74.7$\\
    &SA-ATSS&$\mathbf{44.7}$&$85.8$&$\mathbf{95.9}$&$77.6$&$\mathbf{43.5}$&$\mathbf{8.8}$&$\mathbf{71.5}$&$\mathbf{25.9}$&$14.2$&$\mathbf{55.7}$&$\mathbf{30.8}$&$\mathbf{6.8}$&$\mathbf{81.5}$&$\mathbf{26.0}$&$14.3$&$\mathbf{72.5}$\\
    \midrule
    \bottomrule
    \end{tabular}
    }
\end{table*}

\subsection{Sensitivity of the \gls{SAOD} Performance Measures to the Image-level Uncertainty Threshold and Detection Confidence Threshold}
Here, we explore the sensitivity of the performance measures used in our \gls{SAOD} framework to the image-level uncertainty threshold $\hat{u}$ and the detection confidence threshold $\bar{v}$.
To do so, we measure DAQ, BA, LRP and LaECE of F-RCNN on $\testdata$ of SAOD-Gen by systematically varying (i) image-level uncertainty threshold $\bar{u} \in [0,1]$ and (ii) detection-level confidence score threshold $\bar{v} \in [0,1]$.
Note that in this analysis, we do not use LRP-optimal threshold for detection-level thresholding, which obtains $\bar{v}$ for each class but instead employ a single threshold for all classes; enabling us to change this threshold easily.
\cref{fig:ablation_threshold} shows how there performance measures change for different image-level and detection-level thresholds.
First, we observe that it is crucial to set both thresholds properly to achieve a high \gls{DAQ}.
More specifically, rejecting all images or accepting all detections $\bar{v}=0$ in \cref{fig:ablation_threshold} results in a very low \gls{DAQ}, highlighting the robustness of \gls{DAQ} in these extreme cases.
Second, setting a proper uncertainty threshold is also important for a high BA (\cref{fig:ablation_threshold}(a)), while it is not affected by detection-level threshold (\cref{fig:ablation_threshold}(b)) since BA indicates the OOD detection performance but not related to the accuracy or calibration performance of the detections.
Third, accepting more images results in better LRP values (green and purple curves in \cref{fig:ablation_threshold}(a)) as otherwise the ID images are rejected with the detection sets being empty.
Similarly, to achieve a high LRP, setting the detection confidence threshold properly is important as well.
This is because, a small confidence score threshold implies more FPs, conversely a large threshold can induce more FNs.
Finally, while we don't observe a significant effect of the uncertainty threshold on LaECE, a large number of detections due to a smaller detection confidence threshold has generally a lower LaECE..
This is also related to our previous analysis in \cref{subsec:relation}, which we show that more detections imply a lower LaECE as depicted in \cref{fig:ablation_threshold}(b) when the threshold approaches $0$.
However, in that case, LRP Error goes to $1$, and as a result, DAQ significantly decreases; thereby preventing setting the threshold to a lower value to superficially boost the overall performance.
%

\subsection{Effect of common improvement strategies on DAQ}
Here, we analyse how common improvement strategies of object detectors affect the \gls{DAQ} in comparison with \gls{AP}.
To do so, we first use a simple but a common baseline model: We use F-RCNN (SAOD-Gen) trained for 12 epochs without multi-scale training.
Then, gradually, we include the following four common improvement strategies commonly used for object detection \cite{RSLoss,ATSS,DETR}:
\begin{enumerate}
    \item increasing number of training epochs,
    \item using multiscale training as described in \cref{app:models},
    \item using ResNet-101 as a stronger backbone \cite{ResNet}, and
    \item using deformable convolutions \cite{DCNv2}.
\end{enumerate}
Table \ref{tab:universalimprovements} shows the effect of these improvement strategies, where we see that stronger backbones increase DAQ, but mainly due to an improvement in LRP Error.
%
%
%
It is also worth highlighting that more training epochs improves AP (e.g. going from 12 to 36 improves AP from $24.8$ to $25.5$), but not DAQ due to a degradation in \gls{LaECE}.
This is somewhat expected, as longer training improves accuracy, but drastically make the models over-confident~\cite{FocalLoss_Calibration}.

\subsection{The Impact of Domain-shift on Detection-level Confidence Score Thresholding}
For detection-level confidence score thresholding, we employ LRP-optimal thresholds by cross-validating a threshold $\bar{v}$ for each class using $\valdata$ against the LRP Error.
%
%
%
While LRP-optimal thresholds are shown to be useful if the test set follows the same distibution of $\valdata$, we note that our $\indata$ is collected from a different dataset, introducing domain shift as discussed in App. \ref{app:datasets}.
As a result, here we investigate whether the detection-level confidence score threshold is affected from domain shift.

To do so, we compute the LRP-optimal thresholds on both $\valdata$ and $\testdata$ of F-RCNN and ATSS, and then, we compare them in  \cref{fig:lrpoptimal} for our both datasets. 
As $\testdata$ is not available during training, the thresholds obtained on $\testdata$ correspond to the oracle detection-level confidence score thresholds.
We observe in  \cref{fig:lrpoptimal} that:
\begin{itemize}
    \item For both of the settings optimal thresholds computed on val and test sets rarely match.
    \item While the thresholds obtained on $\valdata$ (COCO val set) and $\indata$ (Obj45K) for SAOD-Gen dataset is relatively similar, they are more different for our SAOD-AV dataset, which can be especially observed for the \texttt{bicycle} class.
\end{itemize}
Therefore, due to the domain shift in our datasets, the optimal threshold diverges from $\valdata$ to $\indata$ especially for AV-OD dataset. This is not very surprising due to the challenging nature of BDD100K dataset including images at night and under various weather conditions, which also ensues a significant accuracy drop for this setting (\cref{tab:general_od_id_corruption}).

Furthermore, to see how these changes reflect into the performance measures, we include $\mathrm{LRP_{Loc}}$, $\mathrm{LRP_{FP}}$, $\mathrm{LRP_{FN}}$ components of LRP Error that are defined as the average over the localisation errors (1-IoU) of TPs, 1-precision and 1-recall respectively in \cref{tab:evaluation_}.
We would normally expect that the precision and the recall errors to be balanced once the detections are filtered out using the LRP-optimal threshold \cite{LRPPAMI}.
This is, in fact, what we observe in $\mathrm{LRP_{FP}}$ and $\mathrm{LRP_{FN}}$ for the $\mathcal{D}_{\mathrm{ID}}$ of SAOD-Gen setting,.
For example for F-RCNN, $\mathrm{LRP_{FP}}=48.5$ and $\mathrm{LRP_{FN}}=52.3$; indicating a relatively balanced precision and recall errors.
As for SAOD-AV setting, the significant domain shift of BDD45K is also reflected in the difference between $\mathrm{LRP_{FP}}$ and $\mathrm{LRP_{FN}}$ for both F-RCNN and ATSS.
For example for F-RCNN, $\mathrm{LRP_{FP}}=13.2$ and $\mathrm{LRP_{FN}}=58.1$; indicating a significant gap.
Besides, as the domain shift increases with $\mathcal{T}(\mathcal{D}_{\mathrm{ID}})$ on SAOD-AV, the gap between $\mathrm{LRP_{FP}}$ and $\mathrm{LRP_{FN}}$ increases more.
These suggest that more accurate detection-level thresholding methods are required under domain-shifted data.

\begin{figure*}[t]
        \centering
        \includegraphics[width=0.95\textwidth]{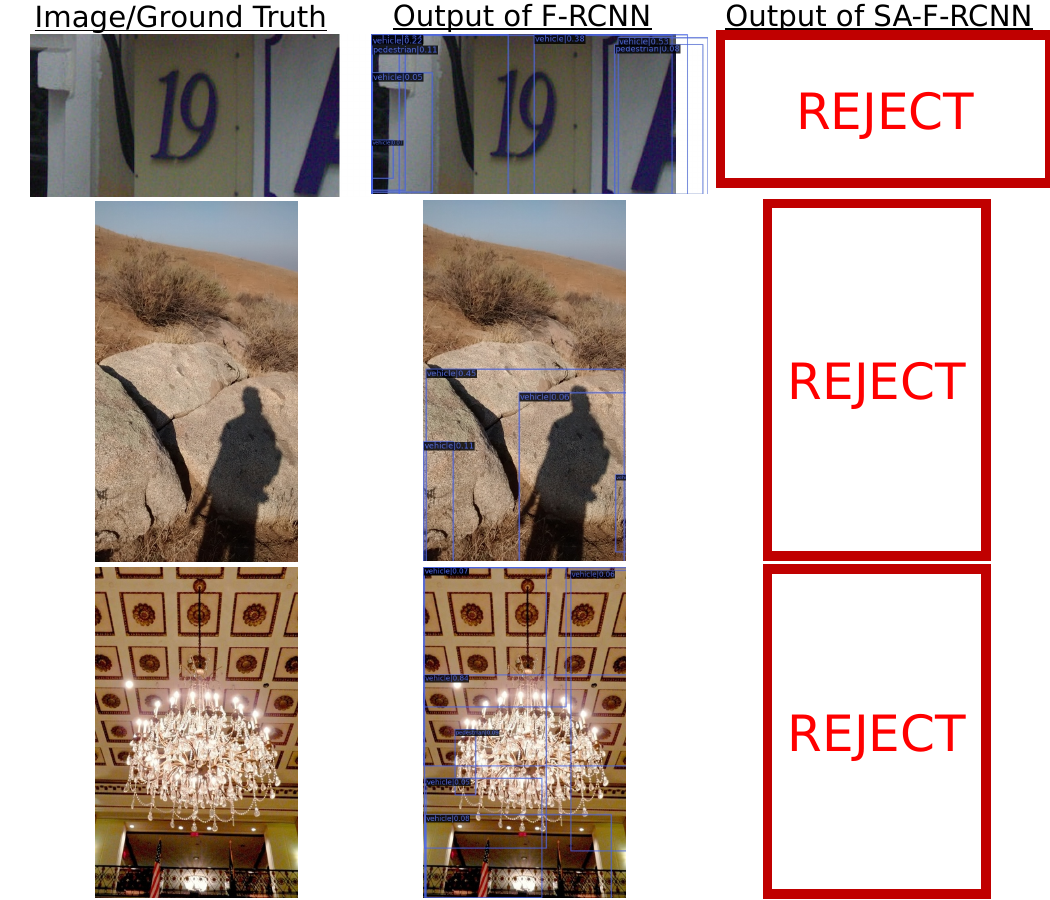}
        \caption{Qualitative Results of F-RCNN vs. SA-F-RCNN on $\ooddata$. The images in the first, second and third rows correspond SVHN, iNaturalist and Objects365 subset of $\ooddata$. While F-RCNN performs inference with non-empty detections sets, SA-F-RCNN rejects all of these images properly.}
        \label{fig:qualitative_FRCNN_OOD}
\end{figure*}

\begin{figure*}[t]
        \centering
        \includegraphics[width=0.95\textwidth]{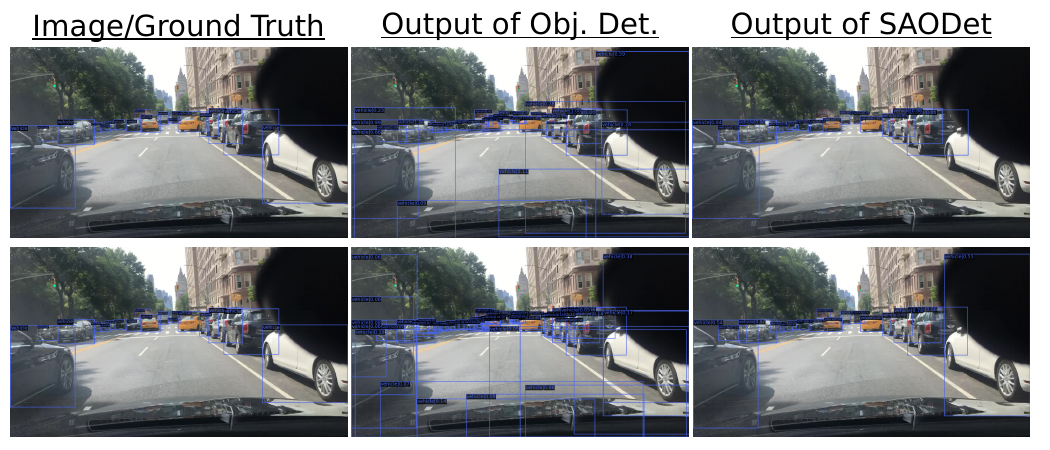}
        \caption{Qualitative Results of Object detectors and \gls{SAODet}s on $\indata$. (First row) F-RCNN vs. SA-F-RCNN. (Second row) ATSS vs. SA-ATSS. See text for discussion. The class labels and confidence scores of the detection boxes are visible once zoomed in.}
        \label{fig:qualitative_FRCNN_ID}
\end{figure*}

\begin{figure*}[t]
        \centering
        \includegraphics[width=0.95\textwidth]{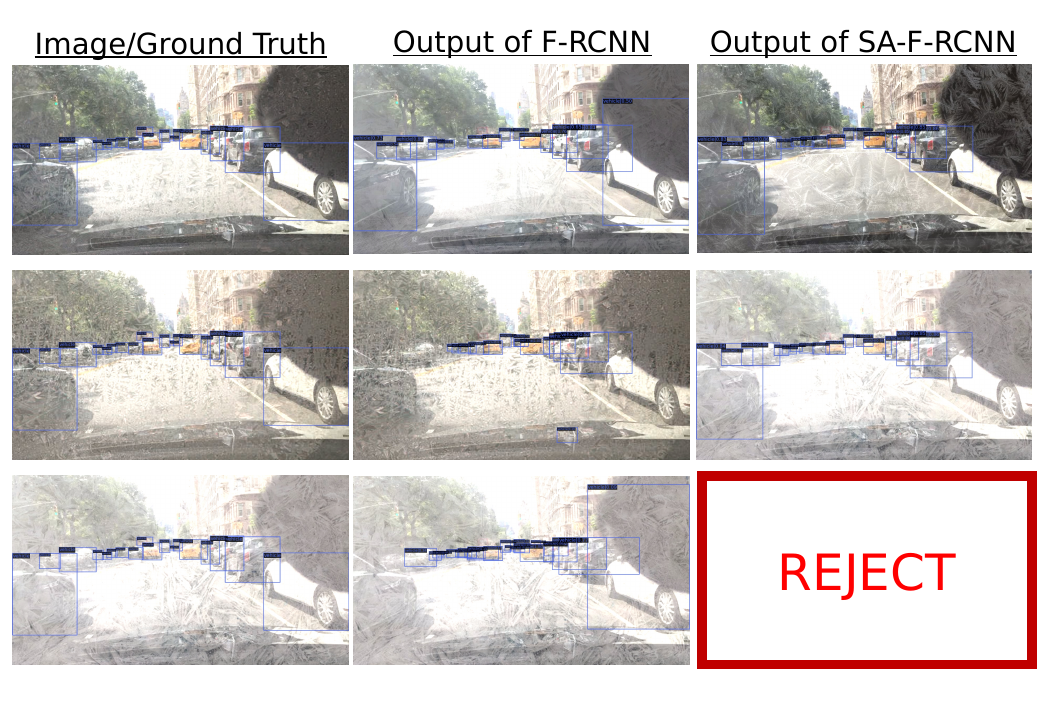}
        \caption{Qualitative Results of F-RCNN vs. SA-F-RCNN on $\shiftdata$ using SAOD-AV dataset. First to third row includes images from $\shiftdata$ in severities 1, 3 and 5 as we used in our experiments. The class labels and confidence scores of the detection boxes are visible once zoomed in. For each detector, we sample a transformation using the `frost' corruption. }
        \label{fig:qualitative_FRCNN_ID_tr}
\end{figure*}

\begin{figure*}[t]
        \centering
        \includegraphics[width=0.95\textwidth]{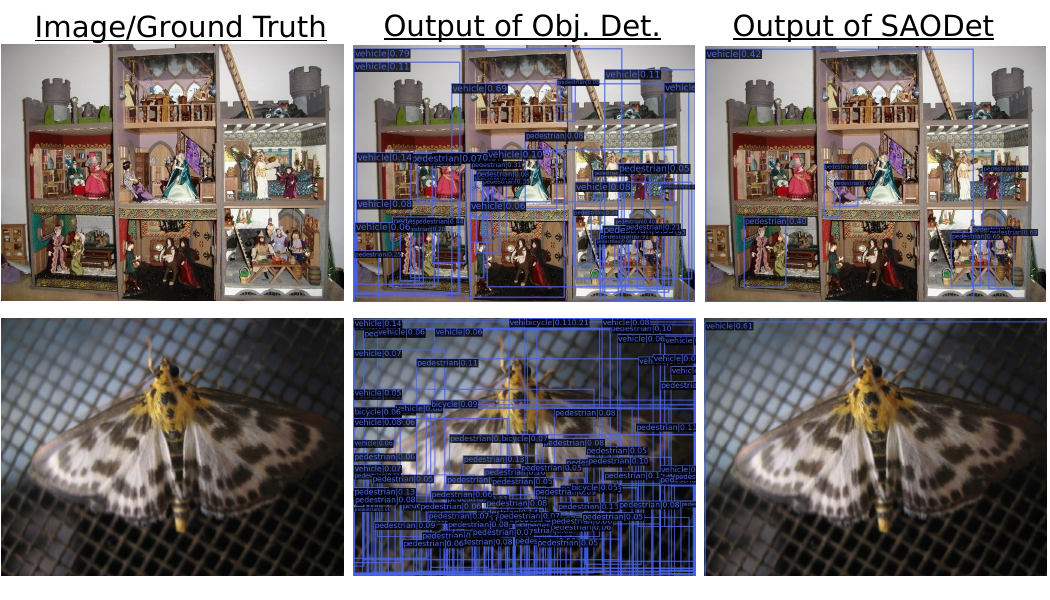}
        \caption{Failure cases of \gls{SAODet}s in comparison to Object detector outputs.  First row includes an image from iNaturalist subset of $\ooddata$ with the detections from ATSS and SA-ATSS trained on nuImages following our SAOD-AV dataset. While SA-ATSS removes most of the low-scoring detections, it still classifies the image as ID and perform inference. Similarly, the second row includes an image from Objects365 subset of $\ooddata$ with the detections from F-RCNN and SA-F-RCNN trained on nuImages again following our SAOD-AV dataset. SA-F-RCNN misclassifies the image as ID and performs inference.}
        \label{fig:qualitative_FRCNN_OOD_failure}
\end{figure*}

\subsection{Qualitative Results of \gls{SAODet}s in comparison to Conventional Object Detectors}
In order provide more insight on the behaviour of \gls{SAODet}s, here we present the inference of \gls{SAODet}s in comparison to conventional object detectors.
Here, we use SA-ATSS and SA-F-RCNN on SAOD-AV dataset.
To be consistent with the evaluation, we plot all the detection boxes as they are evaluated: That is, we use top-$k$ survival for F-RCNN and ATSS and thresholded detections for SA-F-RCNN and SA-ATSS.
In the following, we discuss the main features of the \gls{SAODet}s using \cref{fig:qualitative_FRCNN_OOD}, \cref{fig:qualitative_FRCNN_ID}, \cref{fig:qualitative_FRCNN_ID_tr} and \cref{fig:qualitative_FRCNN_OOD_failure}.

\paragraph{OOD Detection}
\cref{fig:qualitative_FRCNN_OOD} shows on three different input images from different subsets of $\ooddata$ that a conventional F-RCNN performs detection on OOD images and output detections with high confidence.
For example F-RCNN detects a \texttt{vehicle} object with 0.84 confidence on the OOD image from Objects365 (last row).
On the other hand, SA-F-R-CNN can successfully leverage uncertainty estimates to reject these OOD images.

\paragraph{Calibration} 
In \cref{fig:frcnnreliabilityavod}, we presented that ATSS is under-confident and F-RCNN is over-confident.
Now, \cref{fig:qualitative_FRCNN_ID} shows that the calibration performance of these models are improved accordingly.
Specifically, SA-ATSS now has larger confidence scores compared to its conventional version and vice versa for SA-F-RCNN.
This can enable the subsequent systems exploiting these confidence scores for decision making to abstract away the difference of the confidence score distributions of the detectors.

\paragraph{Removing Low-scoring Noisy Detections}
We previously discussed in \cref{subsec:relation} that the detections obtained with top-$k$ survival allows low-scoring noisy detections and that the performance measure AP promotes them (mathematical proof in App. \ref{app:calibration}).
This is also presented in the images of conventional object detectors in  \cref{fig:qualitative_FRCNN_ID}.
For example, the output of ATSS includes several low-scoring detections, which the practical applications might hardly benefit from.
On the other hand, the outputs of the \gls{SAODet}s in the same figure are more similar to the objects presented in ground truth images (first column) and barely contain any detection that may not be useful.

\paragraph{Domain Shift}
\cref{fig:qualitative_FRCNN_ID_tr} includes images with the corruptions with severities 1, 3 and 5.
Following the design of the \gls{SAOD} task, SA-F-RCNN accepts the images with corruptions 1 and 3 and provide detections also by calibrating the detection scores which is similar to $\indata$.
However, for the image with severity 5, different from the conventional detector, SA-F-CNN rejects the image; implying that the detector is uncertain on the scene.

\paragraph{Failure Cases}

Finally in \cref{fig:qualitative_FRCNN_OOD_failure}, we provide images that SA-F-RCNN and SA-ATSS fail to identify the image from $\ooddata$ as OOD, but instead perform inference.

\subsection{Suggestions for Future Work}

Our framework provides insights into the various elements needed to build self-aware object detectors. 
Future research should pay more attention to each elements independently while keeping in mind that these elements are tightly intertwined and greatly impact the ultimate goal.
One could also try to build a self-aware detector by directly optimizing DAQ which accounts for all the elements together, although in its current state it is not differentiable so a proxy loss or a method to differentiate through such non-differentiable functions would need to be employed.

\end{document}